\documentclass[conference]{IEEEtran}
\IEEEoverridecommandlockouts
\usepackage{cite}
\usepackage{amsmath,amssymb,amsfonts}
\usepackage{algorithmic}
\usepackage{textcomp}
\usepackage{xcolor}
\usepackage{multirow}
\usepackage{lscape}
\usepackage{rotating}
\usepackage{subcaption}
\usepackage{graphicx}

\def\BibTeX{{\rm B\kern-.05em{\sc i\kern-.025em b}\kern-.08em
    T\kern-.1667em\lower.7ex\hbox{E}\kern-.125emX}}
\begin{document}

\title{A Threefold Review on Deep Semantic Segmentation: Efficiency-oriented, Temporal and Depth-aware design\\
}

\author{\IEEEauthorblockN{1\textsuperscript{st} Felipe Barbosa}
\IEEEauthorblockA{\textit{Institute of Mathematical and Computer Sciences (ICMC)} \\
\textit{University of S$\tilde{a}$o Paulo}\\
S$\tilde{a}$o Carlos, S$\tilde{a}$o Paulo, Brazil}
\and
\IEEEauthorblockN{2\textsuperscript{nd} Fernando Os\'orio}
\IEEEauthorblockA{\textit{Institute of Mathematical and Computer Sciences (ICMC)} \\
\textit{University of S$\tilde{a}$o Paulo}\\
S$\tilde{a}$o Carlos, S$\tilde{a}$o Paulo, Brazil}
}

\maketitle

\begin{abstract}
Semantic image and video segmentation stand among the most important tasks in computer vision nowadays, since they provide a complete and meaningful representation of the environment by means of a dense classification of the pixels in a given scene.
Recently, Deep Learning, and more precisely Convolutional Neural Networks, have boosted semantic segmentation to a new level in terms of performance and generalization capabilities. However, designing Deep Semantic Segmentation models is a complex task, as it may involve application-dependent aspects. Particularly, when considering autonomous driving applications, the robustness-efficiency trade-off, as well as intrinsic limitations – computational/memory bounds and data-scarcity – and constraints - real-time inference – should be taken into consideration.
In this respect, the use of additional data modalities, such as depth perception for reasoning on the geometry of a scene, and temporal cues from videos to explore redundancy and consistency, are promising directions yet not explored to their full potential in the literature.
In this paper, we conduct a survey on the most relevant and recent advances in Deep Semantic Segmentation in the context of vision for autonomous vehicles, from three different perspectives: efficiency-oriented model development for real-time operation, RGB-Depth data integration (RGB-D semantic segmentation), and the use of temporal information from videos in temporal-aware models.
Our main objective is to provide a comprehensive discussion on the main methods, advantages, limitations, results and challenges faced from each perspective, so that the reader can not only get started, but also be up to date in respect to recent advances in this exciting and challenging research field.

\end{abstract}


\section{Introduction}
\label{section:introduction}
Significant progress has been made in Computer Vision research since the adoption of Deep Learning and, more specifically, Convolutional Neural Networks (CNNs). Particularly, considering the task of semantic segmentation, since the proposition of Fully Convolutional Networks (FCNs) \cite{b0}, and contemporary models \cite{b1} \cite{b2}, many efforts have been made to design highly-accurate Deep Learning-based Semantic Segmentation (Deep Semantic Segmentation) architectures. This led to the development of an accuracy-oriented line of research, where design choices are made so to extract the most meaningful representation out of the available data \cite{b3} \cite{b5} \cite{b6}. Adopting high-resolution inputs and features \cite{b7} \cite{b8}, multi-scale processing \cite{b9} \cite{b10}, spatial and semantic reasoning \cite{b11} \cite{b12} \cite{b13}, local and global context aggregation \cite{b3} \cite{b5} \cite{b14}, as well as different strategies of feature fusion \cite{b1} \cite{b15} \cite{b16} are some of the most common strategies.
\par This increase in accuracy, though, usually means sacrificing performance: accuracy-oriented methods are generally associated with high computational costs and memory requirements, hindering low-latency inference. These prohibitive costs limit the application of such methods in current critical applications that require real-time perception, such as autonomous driving and driver assistance systems \cite{b17}. 
\par As a consequence, efficiency-oriented designs arouse as an alternative towards reduced model complexity and hardware requirements, so to achieve the challenging research objective of real-time inference. The works inside this category can be generally divided into: input-level, architecture-level, and operation-level approaches. Some of the main strategies include reduced input and feature size \cite{b18} \cite{b19} \cite{b20}, adoption of lightweight backbones \cite{b21} \cite{b15} \cite{b22} \cite{b19}, lightweight model design \cite{b23} \cite{b24} \cite{b25} \cite{b26}, weight-sharing and feature reuse \cite{b27} \cite{b11} \cite{b28}\cite{b29}\cite{b30}, optimized operations \cite{b31} Xception: Deep Learning with Depthwise Separable Convolutions] \cite{b32} \cite{b33}, selective feature processing \cite{b34}, and knowledge distillation \cite{b35} \cite{b36} \cite{b37}.
\par However, despite being faster, efficiency-oriented methods struggle in achieving competitive accuracy to their heavy, accuracy-oriented counterparts. In the seek for stablishing a more balanced accuracy-performance trade-off, some promising alternatives have recently been explored, such as multi-modal perception and temporal-aware models.
\par Autonomous vehicles are equipped with a variety of sensors, and cameras are frequently used in perception setups because they provide a large field of view, as well as a colored, textured, and high-resolution representation of the environment. However, relying solely on RGB information limits perception in complex environments. With the advent of 3D sensors, such as stereo camerasZED2, depth maps aroused as a promising new data modality. Because it provides rich geometry and contour information about the environment, 3D data has gained the attention of the computer vision community. Great part of the literature on Deep Learning-based 3D perception, though, is developed considering indoor scenarios, where 3D perception from structured lightKINECT is possible, and where depth ranges are limited to a few meters. The use of depth data from stereo cameras in more challenging scenarios, such as urban scenes, is a relatively under-explored area.
\par Another factor that limits model performance is the wide adoption of single-frame-based architectures in the literature, where frames are treated independently. In spite of that, in real-time applications, such as automated driving, videos are the natural data format captured by cameras. Thus, developing temporal-aware models allows to explore frame correlations as a valuable source of information for temporal continuity and redundancy, which could help improving both accuracy and efficiency. There has been a growing interest in video semantic segmentation, with works leveraging temporal cues mainly for feature and label propagation \cite{b30} \cite{b38} \cite{b39}, and refinement \cite{b28}. 
\par Besides that, leveraging depth and temporal cues can help solving other inherent challenges to Deep Semantic Segmentation, mainly related to data scarcity \cite{b40} \cite{b41} and domain shift issues \cite{b42} \cite{b43} \cite{b44} \cite{b45} \cite{b46}. 
\par Despite the maturity of the field, many new exciting research opportunities have emerged, and recent advancements still lack being gathered in a comprehensive survey. Given the aforementioned scenario, the main goal of this review is to provide a comprehensive study on Deep Semantic Segmentation of urban scenes, considering three main aspects: accuracy-oriented design, RGB-Depth integration, and temporal-awareness. 
\par The remaining of this paper is organized as follows. In section 2, the related works are presented. A general overview on accuracy-oriented strategies is performed in section 3. Efficiency-oriented, RGB-Depth (RGB-D) multi-modal, and temporal-aware Deep Semantic Segmentation are discussed in sections 4, 5, and 6, respectively. In section 7 we discuss the main challenges and recent developments in the field. We conclude our work in section 8.

\section{Related Works}
\label{section:related_works}
Throughout the years, several reviews on semantic segmentation were published. Concerning Deep Learning-based methods, we observed a growth in the availability of surveys since 2015, when the Fully Convolutional Networks (FCNs) \cite{b0} first treated semantic segmentation as a dense per-pixel classification problem by replacing dense layers by corresponding convolutional ones.
\par Initial works compared classical and Deep Learning-based semantic segmentation (Deep Semantic Segmentation) in order to demonstrate the potential of the latter as a more general and powerful approach. Yu et al. \cite{b47} classify existing semantic segmentation models into hand-engineered feature-based methods, learned feature-based methods, and weakly and semi-supervised methods. Atif et al. \cite{b48} covers some of the seminal works in the area, and proposes the division of semantic segmentation methods into accuracy-oriented and efficiency-oriented. Despite providing valuable discussions on general-purpose 2D semantic segmentation, none of these works discuss the use of depth or temporal information. 
\par The adoption of additional data modalities, such as depth, characterize the so-called multi-modal perception architectures, which were addressed by subsequent works. Feng et al. \cite{b49} provide a comprehensive study on Multi-modal Deep Object Detection and Semantic Segmentation in the context of autonomous driving. However, the focus is given to camera and LIDAR perception, just briefly mentioning stereo perception.
\par Although we target the specific context of automated driving, the majority of research on RGB-D Semantic Segmentation covers indoor scenarios. Hence, most of the following related works deal with RGB-D indoor perception. Wang et al. \cite{b50} describe several approaches to RGB-D semantic segmentation, including: multi-modal fusion (input-level, feature-level and output-level); attention-based mechanisms for feature fusion; adaptive convolutions – depth-aware convolution, 3d Neighborhood Convolution (3DN-Conv) filters and receptive fields – and 2.5D convolutions; and RNN-based RGB-D processing. Barchid et al. \cite{b51} propose, in a brief survey, a categorization of RGB-D semantic segmentation methods into depth as input, depth as operation and depth as prediction, according to their use of depth data. Fooladgar et al. \cite{b52} and Hu et al. \cite{b53} review both hand-crafted (traditional) and Deep Learning-based methods, as well as commonly-used RGB-D datasets for indoor Semantic Segmentation. Papadopoulos et al. \cite{b54} cover some of the state-of-the-art Deep Learning models for semantic segmentation, depth estimation, and joint inference of both. 
\par As far as temporal-aware Deep Semantic Segmentation is concerned, the literature is still limited, with few works providing in-depth discussion on the theme. Lateef et al. \cite{b55} propose a categorization of Deep Semantic Segmentation methods into ten classes, according to common concepts underlying their architectures. The authors conduct an overview of spatio-temporal semantic segmentation, with a focus on methods combining CNNs and RNNs. Wang et al. \cite{b56} conduct a detailed survey on Deep Video Segmentation, describing methods, learning paradigms, datasets, performance comparisons and interesting future research directions. 
\par In the context of autonomous driving, the use of temporal information is addressed by Siam et. al \cite{b57}. The authors propose a classification of Deep Semantic Segmentation methods into: classical, fully convolutional networks, structured models, and spatio-temporal models; however, it lacks in-depth discussions on each of the categories, and most of the recent advancements are not covered. In spite of that, a fruitful discussion is conducted on the common characteristics and challenges faced by Deep Semantic Segmentation in the specific context of Autonomous Driving.
\par Simultaneous considerations on leveraging depth and temporal cues can be encountered in \cite{b58} and \cite{b59}. The authors briefly describe a few works on RGB-D segmentation, the integration of contextual knowledge, and video sequences; however, without further exploring the topics, they reserve themselves to describing a few pioneering and example works. 
\par
\begin{figure}[htbp]
\centerline{\includegraphics[width=\columnwidth]{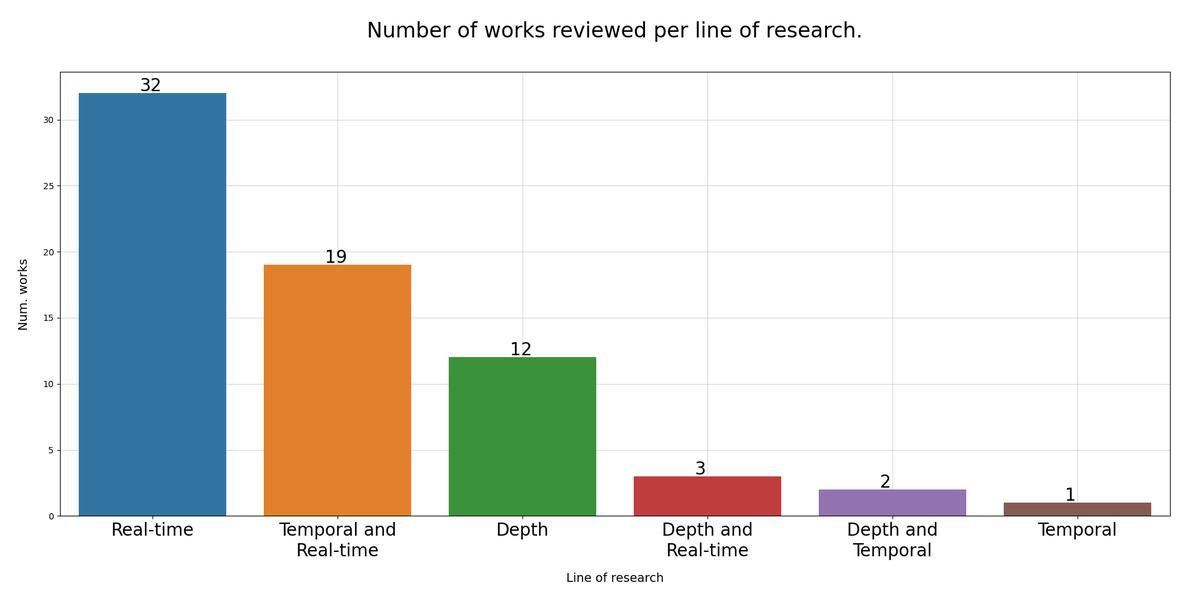}}
\caption{Number of works reviewed per line of research..}
\label{fig:works_per_line}
\end{figure}

 Finally, the number of available reviews on real-time semantic segmentation is the most limited – in spite of being one of the most addressed topics in the recent literature (Fig.~\ref{fig:works_per_line}). Previous works only treated the topic as a challenge and open question for future work \cite{b50} \cite{b49} \cite{b55}, without detailed discussions. Gamal et al. \cite{b60} presents the results from a segmentation benchmarking framework. Within a “meta architecture”, different encoder and decoder combinations are tested in order to evaluate the impact of such structures in real-time performance. Interestingly, the platform used for benchmarking is an embedded device (Jetson TX2), so as to be as close as possible to a real application scenario. Additionally, the authors propose a taxonomy dividing the literature on semantic segmentation into: (1) Fully Convolutional Networks, (2) Context Aware Models, and (3) Temporal Models. The main discussion, though, is on real-time models, where the authors address efficient design and model compression strategies. Nonetheless, besides being a little outdated, the work does not cover alternative data modalities, such as stereo-based depth maps. Mo et al. \cite{b61} address important topics in modern Deep Semantic Segmentation, such as weakly-supervised learning, domain adaptation, multi-modal data fusion, and real-time inference. Nonetheless, it does not cover temporal-aware models and, although being a very recent review, it does not cover the most recent works on multi-modal and real-time semantic segmentation.
\par The previous surveys characterize the area of Deep Semantic Segmentation from particular points of view. Some of them even carry a review were two out of depth perception, temporal awareness and real-time inference are explored jointly; however, when doing so, most of them limit themselves to shallow analysis, where recent advances are not covered. None of them, though, treat all three aspects in conjunction. For instance, RGB-D segmentation and temporal-aware models are presented in \cite{b58} and \cite{b59}; \cite{b60} argues on real-time inference and video segmentation; in \cite{b61}, real-time inference and RGB-D segmentation are addressed. Hence, to the best of our knowledge, our work is the first to address Deep Semantic Segmentation from such a diverse set of perspectives. 
\par In this review, we cover both pioneer and recent literature on urban Deep Semantic Segmentation, focusing on works that provide results on the Cityscapes \cite{b62} dataset, which is widely adopted and accepted as a benchmarking for model performance. As our main contributions, we:
\begin{itemize}
\item Provide a comprehensive overview and discussion on the recent literature about Deep Semantic Segmentation of urban scenes, from three different perspectives: Multi-modal (RGB and depth) perception for geometry reasoning, video processing for temporal awareness, and efficiency-oriented model design for real-time inference;
\item Propose a categorization of depth-aware models based on the different strategies for leveraging depth information in Deep Semantic Segmentation;
\item Propose a categorization of temporal-aware models according to their goal and main architectural characteristics;
\item Propose a categorization of real-time Deep Segmentation models according to the level in which strategies are applied to improve efficiency;
\item Compile the latest results in the literature on depth-aware, temporal-aware, and real-time Deep Semantic Segmentation, with respect to: model design, accuracy, inference time, and operating statistics;
\item Discuss the main and most recent challenges in the field of Deep Semantic Segmentation for autonomous driving.
\end{itemize}

\section{Accuracy-oriented Deep Semantic Segmentation}
\label{section:accuracy}
\begin{figure}[htbp]
\centerline{\includegraphics[width=\columnwidth]{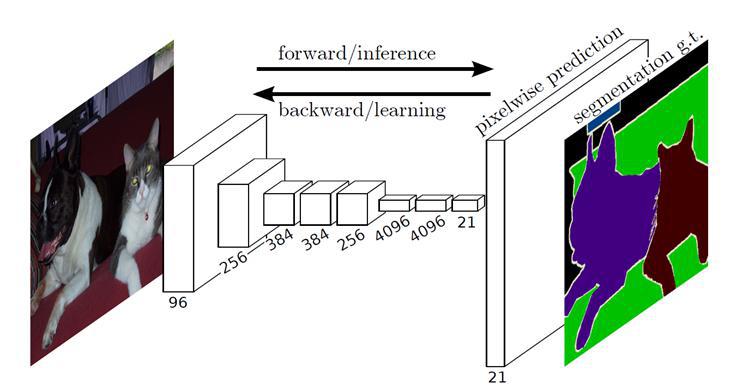}}
\caption{FCN architecture. \cite{b0}.}
\label{fig:fcn_architecture}
\end{figure}

Since the wide adoption of Deep Learning models in Computer Vision research, a superior performance in terms of accuracy and generalization was achieved when compared to the use of hand-crafted features. Probably the most important step towards Deep Semantic Segmentation was given in 2015, with the proposition of the so-called FCNs \cite{b0}, which tackled semantic segmentation as a dense per-pixel classification problem. By replacing the dense layers of a VGG16 classification model \cite{b63} by their convolutional counterparts, the new model was able to process images at any resolution, as well as to deliver a set of heatmaps instead of a vector of probabilities for the entire image, where each position in a certain map corresponds to the pixel’s probability of belonging to the class related to that map. The full-resolution prediction was then generated by an up-sampling layer at the end of the architecture – illustrated in Fig.~\ref{fig:fcn_architecture}.
\par The advancements achieved by this work, as well as subsequent ones, among which stand out SegNet \cite{b2} and U-Net \cite{b1}, paved the road for the development of a rich and diverse literature on Deep Segmentation aimed at improved accuracy.
\par In order to better cover the history of Deep Semantic Segmentation, and justify the importance of efficiency-oriented methods, herein we introduce the line of work which explores the design of accurate architectures. Throughout our discussion, we will mention some of the most common techniques for improving accuracy in Deep Segmentation models.

\subsection{Improve Feature Representativeness}
\label{subsection:accuracy_improve_feature}
\begin{figure}[htbp]
\centerline{\includegraphics[width=\columnwidth]{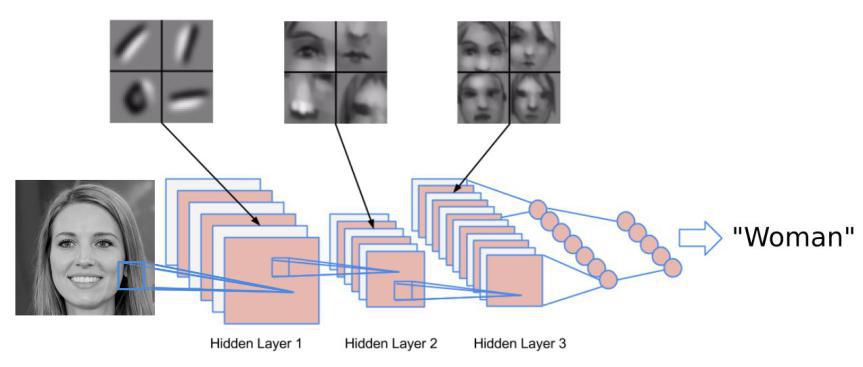}}
\caption{Illustration of a simple CNN architecture. Convolutional layers explore locality through limited receptive fields. The stacking of convolutional layers gradually increases the context coverage and the complexity/semantics of features.}
\label{fig:cnn_architecture}
\end{figure}

Convolutional Neural Networks are hierarchical models, in which features are extracted in increasing levels of complexity (Fig.~\ref{fig:cnn_architecture}). In lower-level layers, patterns such as lines with different orientations are extracted; subsequent levels extract compositions of lines, such as corners; higher-level layers reason on more complex patterns, including textures and colors; finally, in the deepest layers, complex shapes are extracted. This increasing in feature complexity also represents gains in the semantic content of feature maps. In summary, lower-level (high-dimensional) feature maps encode detailed spatial information, while higher-level (low-dimensional) feature maps encode stronger semantic information. 
\par One of the ways by which Deep Semantic Segmentation models try to enhance accuracy is precisely by exploiting the hierarchical behavior of CNNs, so as to improve feature representation through feature fusion techniques. The works lying in this class can be grouped into four main categories: multi-level, multi-scale, multi-modal, and multi-term feature fusion.

\subsubsection{Multi-level feature fusion}
\label{subsubsection:accuracy_improve_feature-multi_level}
\begin{figure*}[htbp]
\centerline{\includegraphics[width=\textwidth]{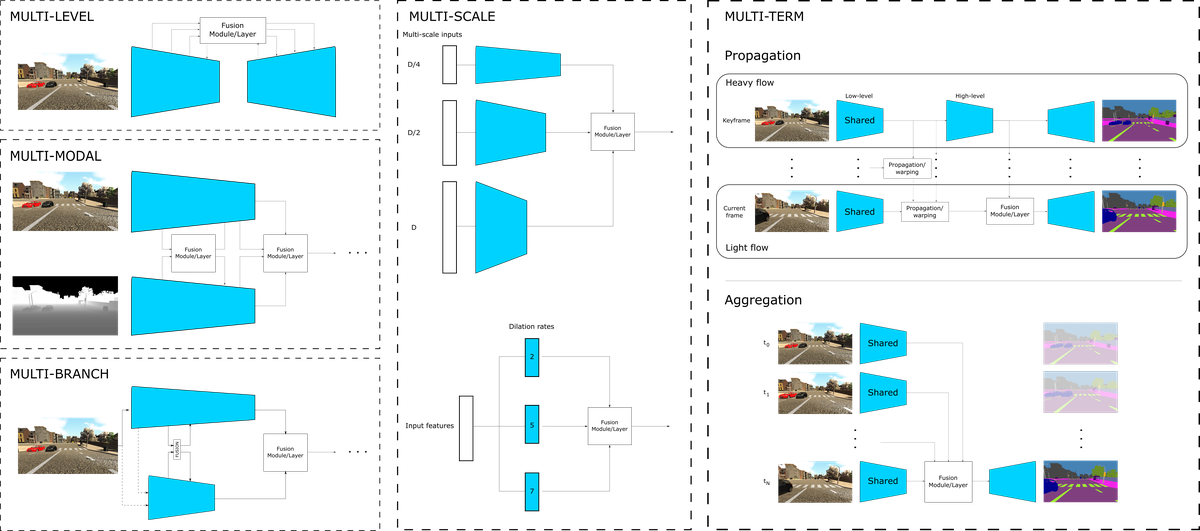}}
\caption{Types of feature fusion.}
\label{fig:fusion_types}
\end{figure*}

Multi-level feature fusion aims at improving feature representation by combining feature maps from different levels of the main network stream – Fig.~\ref{fig:fusion_types}. Great part of such methods tries to fuse low-level (high-resolution) features with high-level (low-resolution) features in order to enhance semantics with more detailed spatial information, which is beneficial to a better delineation of boundaries in the final segmentation.

\par Even in pioneering works \cite{b0} \cite{b1}, authors already applied this strategy. FCNs \cite{b0} gradually combine multi-level features in order to generate higher-dimensional feature maps, consequently reducing the up-sampling factor needed to generate the final segmentations – resulting in finer-grained predictions. U-Net \cite{b1} proposed to combine feature maps from the encoder, and its corresponding decoder feature maps by a “copy and crop” mechanism. The concatenated representation is then processed by convolutional layers and up-sampled to the next stage in the decoder branch, where a similar process takes place. LinkNet \cite{b64} explores a similar mechanism, but using element-wise addition instead of feature concatenation. Si et al. \cite{b15} design a Multi-features Fusion Module (MFM) in order to extract and combine feature from different levels of their encoder.
\par A recent trend in Deep Semantic Segmentation is the extraction of spatial and semantic information through separate encoder branches. This leads to a subdivision of multi-level feature fusion methods that can be termed as multi-branch feature fusion (Fig.~\ref{fig:fusion_types}). The main intuition behind this approach is, similarly to multi-level fusion, to combine spatial details with semantic information in order to get a better delineation of the semantic masks. BiseNet \cite{b22}, BiseNet V2 \cite{b13}, and BiAlignNet \cite{b12} are examples where context and spatial paths are used to extract features from the same input image. Aiming at the promotion of a better detail extraction, a shallower network processes feature maps with a higher number of channels. The context path, on the other hand, has a deeper layer structure, while channel number is kept low. In EACNet \cite{b11}, the spatial branch is implemented with parallel max and average pooling operations followed by convolutions.

\par Simply fusing features, though, may not be optimal since, because of the hierarchical nature of CNNs, there is an inherent semantic gap between features from different levels in the architecture. To tackle this problem, Li et al. \cite{b65} propose the Flow Alignment Module (FAM), in order to learn the Semantic Flow between feature maps of adjacent levels, leading to a more efficient and effective feature fusion. Wu et al. \cite{b12} apply a similar strategy, where not only features from different levels, but also from different branches are fused through a Bidirectional Gated Flow Alignment Module.

\subsubsection{Multi-scale feature fusion}
\label{subsubsection:accuracy_improve_feature-multi_scale}
As mentioned by Atif et al. \cite{b48}, an important step towards accurate models is the ability of dealing with scale variance. 
\par One of the ways to tackle the problem is through multi-scale processing. Multi-scale processing aims to extract features at different input resolutions so to get a richer representation of objects at different scales. This is done for a more powerful feature extraction towards a more robust segmentation of classes with high variation in appearance, as well as small-sized classes. Multi-scale fusion is, thus, the process of fusing features resulting from multi-scale processing (Fig.~\ref{fig:fusion_types}).

\par
\begin{figure*}[htbp]
\centerline{\includegraphics[width=\textwidth]{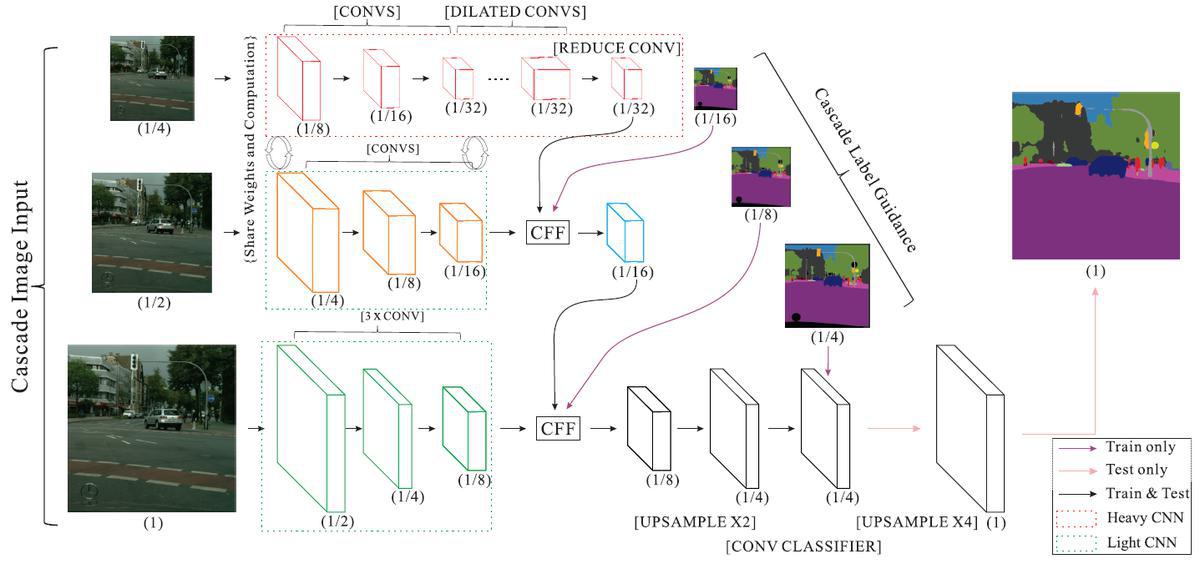}}
\caption{ICNet architecture. \cite{b9}.}
\label{fig:icnet_architecture}
\end{figure*}

 \cite{b9} uses an image cascade input with three different resolutions, each one processed by dedicated branches – Fig.~\ref{fig:icnet_architecture}. Higher-resolution inputs are processed by shallower branches, while lower-resolution inputs are processed by deeper, semantically-stronger, branches. Their main intuition is to extract rich spatial details from high-resolution images, while letting the extraction of semantic cues for low-resolution representations, a mechanism to keep computational burden under control. Oršic and Šegvic \cite{b66} process images at full-resolution, and its ½ and ¼ resized versions, in order to build a multi-scale architecture with shared encoders and multi-level fusion, termed pyramidal fusion. Chen et al. \cite{b10} processes multi-scale features by a shared network scheme, and merges the score maps using an attention mechanism. 

\subsubsection{Multi-modal feature fusion}
\label{subsubsection:accuracy_improve_feature-multi_modal}
Multi-modal feature fusion is generally employed in multi-modal setups, in which besides RGB images, other data modalities are used to enhance feature extraction and scene representation. Works that embed depth information as an additional input in the network generally have some kind of multi-modal feature fusion mechanism in order to aggregate complementary information from modality-specific features – Fig.~\ref{fig:fusion_types}. Methods lying in this category will be discussed in section \ref{section:rgb-d_semantic_segmentation}.

\subsubsection{Multi-term feature fusion}
\label{subsubsection:accuracy_improve_feature-multi_term}
Multi-term feature fusion deals with the aggregation of features extracted from different frames in a given time window. It is usually applied in methods that perform some sort of temporal reasoning. In TDNet \cite{b36}, the authors apply an attention-based module to aggregate features extracted out of different frames in a time window. In \cite{b67}, multi-term features are aggregated using convLSTMs, while in \cite{b30}, a combination of feature concatenation and convolution is applied for feature fusion. Figure~\ref{fig:fusion_types} a illustrates a simple approach for multi-term feature fusion. A more detailed discussion on multi-term fusion strategies will be conducted in section \ref{section:temporal-aware_semantic_segmentation}, where we discuss temporal-aware semantic segmentation.

\subsubsection{Mechanisms for feature fusion}
\label{subsubsection:accuracy_improve_feature-mechanisms}
Generally speaking, simpler fusion mechanisms, such as element-wise addition and channel-wise concatenation, are widely applied in feature fusion. However, this can yield sub-optimal results, since simple fusion or concatenation treats the hugely different information contained in each feature map equally, which may lead to the interference between spatial and semantic information \cite{b68}, as well as among information from different data modalities. Thus, some authors propose more elaborated techniques for selective fusion. In such strategies, the goal is to weight the feature maps according to their importance for the fusion. 
\par Attention and gating mechanisms are a common choice for selective feature fusion, since they allow for a soft weighting of each feature’s contribution to the fused maps. Besides allowing the model to perform better, attention also serves as a filtering stage in order to reduce noise in data – particularly useful in multi-modal models, where noise encountered in a given data modality can negatively influence other modalities under consideration. 
\par In this respect, Xu et al. \cite{b69} and Yu et al. \cite{b13} empirically prove that, for their architectures, gating-based fusion outperforms element-wise addition and channel-wise concatenation. Song et al. \cite{b70} propose the Strip Attention Module and the Attention Fusion Module, to both to maintain contextual consistency along the vertical direction, and improve the effectiveness of information propagation from different levels in the network, respectively. Sun et al. \cite{b71} propose the Attention Feature Complementary module in order to fuse features coming from depth and RGB branches in the architecture. Sun et al. \cite{b72} introduce a refinement module containing two attention units that implement gating mechanisms for prediction and boundary refinement. In \cite{b25} and \cite{b13}, gating-based aggregation layers are responsible for multi-level feature fusion. FPENet \cite{b16} implements channel and spatial attention to fuse multi-level features. Chen et al. \cite{b10} propose an attention mechanism for multi-scale feature fusion, so that they can identify and select the most relevant scales for each pixel location. Their main intuition is that pixels far away will probably receive more information from finer scales (higher resolution). 
\par Besides attention mechanisms, the combination of different fusion operations in specialized modules is also frequently employed in more complex hand-crafted fusion modules. Hong et al. \cite{b24} designs a bilateral fusion module composed by the combination of different operations, such as convolutions and point-wise addition. In the same line, Luo et al. \cite{b32} propose an Encoding-Enhancing module, where multi-level features are fused through a combination of feature concatenation, convolutions, and element-wise multiplication. Yu et al. \cite{b22} design an Attention Refinement Module, and a Feature Fusion Module where feature fusion occurs by a combination of convolutions, element-wise additions and multiplications. As a last example, FASSD-Net \cite{b33} proposes the Muti-resolution Dilated Asymmetric (MDA) module, where the combination of convolutions, concatenation and element-wise addition are used to fuse features from different levels in the architecture.

\subsection{Local and Global Context Modeling}
\label{subsection:accuracy_local_and_global}
\begin{figure*}[htbp]
\centerline{\includegraphics[width=\textwidth]{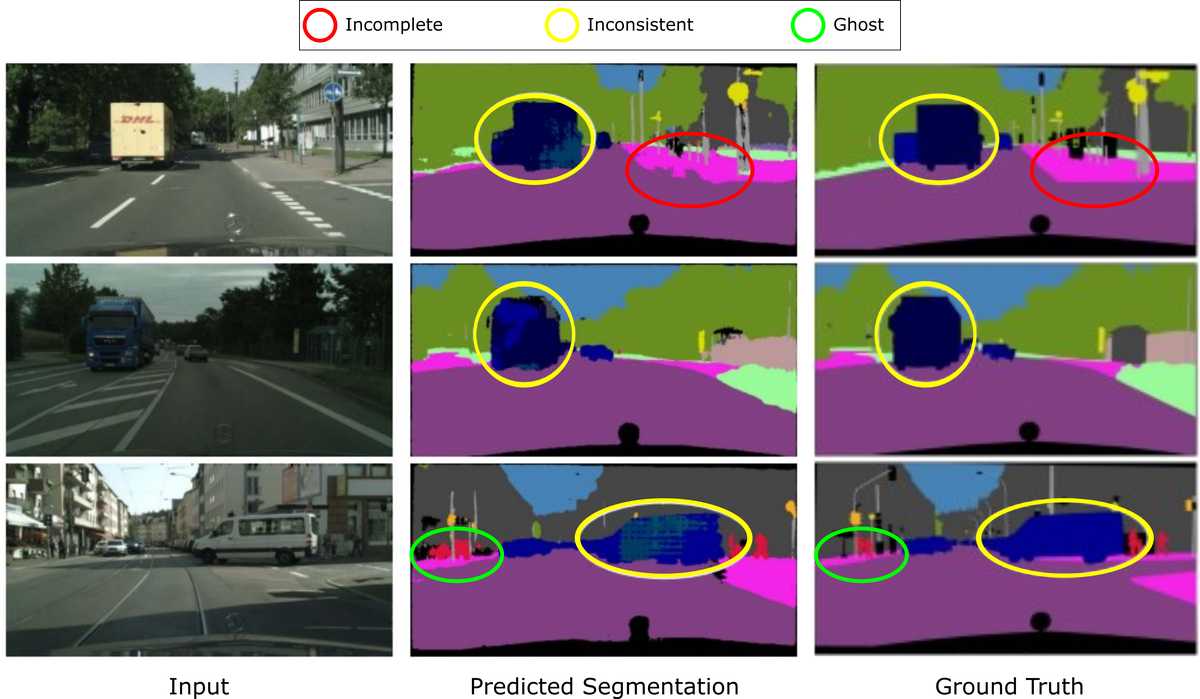}}
\caption{Inconsistent segmentations inside large classes, ghost and incomplete segmentations are some of the common errors in Deep Semantic Segmentation. Adapted from \cite{b8}.}
\label{fig:inconsistent_segmentation}
\end{figure*}

Another line of approach towards accurate Deep Semantic Segmentation deals with the problem of how to properly explore local and global contexts in order to enhance modeling of pixel dependencies. Limited context modeling can lead to inconsistent segmentation inside objects that occupy large areas in the image, such as buses and trucks, and objects that appear close to the camera. Ghost and incomplete segmentations can also be a result of insufficient modeling of contextual information. Covering an excessively wide context, though, can introduce useless information for a given object under consideration, which may negatively affect the segmentation also leading to degraded results. Examples of the aforementioned limitations are illustrated in Fig.~\ref{fig:inconsistent_segmentation}.

\par Local context can be easily extracted when dealing with CNNs, given that the convolution operation naturally explores locality. In summary, its working involves filtering a region of the input – receptive field – at a time by a grid of weights to be learned during optimization – convolutional kernel. The exploration of larger (ultimately global) contexts, in turn, can be performed by the stacking of convolutional layers so that the receptive fields get larger at each level, as illustrated in Fig~\ref{fig:cnn_architecture}. Depending on the input resolution, though, achieving a large receptive field can result in very deep networks, which incur high computational costs, and can lead to problems such as vanishing or exploding gradients, as well as overfitting issues.
\par Given this context, some alternatives are widely adopted in the literature for a good balance of local and global feature exploration.

\subsubsection{Fast Downsampling}
\label{subsubsection:accuracy_local_and_global-fast_downsampling}
The first alternative is to perform a fast down-sampling of feature maps, in order to fully extract contextual information with fewer layers.However, this leads to some of the main challenges to highly-accurate semantic segmentation, as stated by Atif et al. \cite{b48}: reduced localization accuracy introduced by pooling operations to incorporate invariance in the network, and reduced feature map resolution caused by strided convolutions.
\par Loss in localization accuracy was tackled by pioneering works in Deep Semantic Segmentation, such as SegNet \cite{b2}. They proposed to keep spatial details by preserving the location of the values extracted by the max pooling operation in the encoder stages. This information would then be passed to the corresponding decoder layer by means of the so-called Pooling Indices in order to guide the up-sampling stages. \cite{b73} and \cite{b74} try a similar approach.
\par Excessive downsampling of feature maps leads to the loss of information related to small objects. In FCNs, for instance, the output is usually 32 times smaller than the input image \cite{b66}. Throughout the years, various approaches have been proposed in order to reduce loss of spatial details caused by feature downsampling, while guaranteeing the coverage of long-range dependencies. In this context, dilated and deformable convolutions, as well as feature pyramids are some of the most frequently used techniques.

\subsubsection{Dilated and Deformable Convolutions}
\label{subsubsection:accuracy_local_and_global-dilated_deformable_convolutions}
\begin{figure}[htbp]
\centerline{\includegraphics[width=\columnwidth]{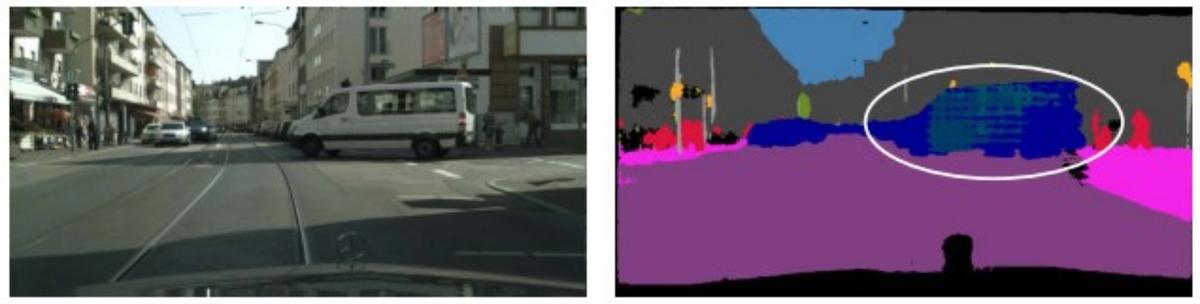}}
\caption{Gridding artifacts. \cite{b8}.}
\label{fig:gridding_artifacts}
\end{figure}

\begin{figure}[htbp]
\centerline{\includegraphics[width=\columnwidth]{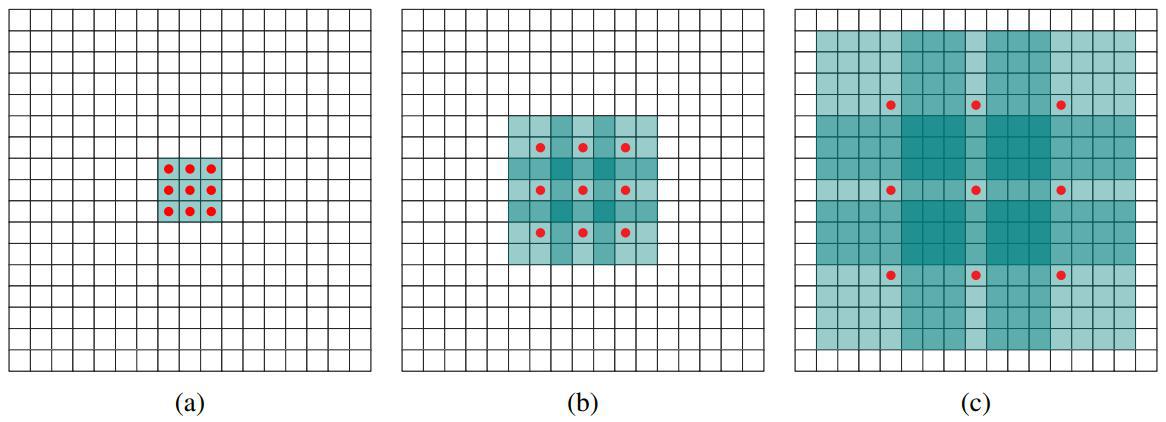}}
\caption{Dilated convolutions. (a) plain 3x3 convolution. (b) 2-dilated convolution, with receptive field increased to 7x7. (c) 4-dilated convolution, with receptive field increased to a 15x15 grid \cite{b179}.}
\label{fig:dilated_convolutions}
\end{figure}

Dilated convolutions are one of the most common practices for enlarging receptive field, while keeping a reasonable feature size. It involves applying a dilated filter, in which intermediate locations are filled with zeros – Fig~\ref{fig:dilated_convolutions}. One drawback of such method, though, is that the model can’t properly backpropagate through these zeroed elements, generating segmentations with gridding artifacts \cite{b8} (Fig.~\ref{fig:gridding_artifacts}). 

\par
\begin{figure}[htbp]
\centerline{\includegraphics[width=\columnwidth]{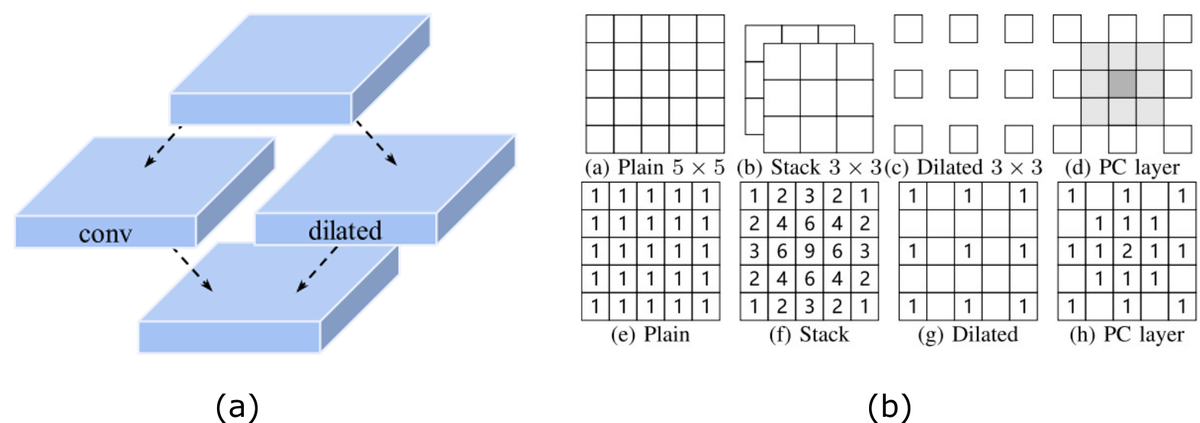}}
\caption{Parallel complementary layer (a), and an illustration of the influence of each pixel in a given receptive field after applying different configurations of stacking of convolutional layers (b). \cite{b8}.}
\label{fig:pc_net}
\end{figure}

 In PCNet \cite{b8}, the authors apply parallel convolutions (plain and dilated) in order to enlarge the receptive field without losing local information – i.e. avoiding inserting gridding artifacts –, but also without relying too much on centering pixels – Fig.~\ref{fig:pc_net}. \cite{b23} proposes a hierarchical dilation block, in which multi-level parallel dilated convolutions are used to enhance feature representation.

\par Although dilated convolutions help spamming longer-range contexts, it does not solve an inherent limitation of CNNs: the fixed kernel structure. Convolutional kernels are generally sets of weights organized in fixed square grids – even when dilated – which are invariant to the spatial location’s characteristics. This leads to limited transformation modeling capabilities in CNNs.
\par
\begin{figure}[htbp]
\centerline{\includegraphics[scale=0.15]{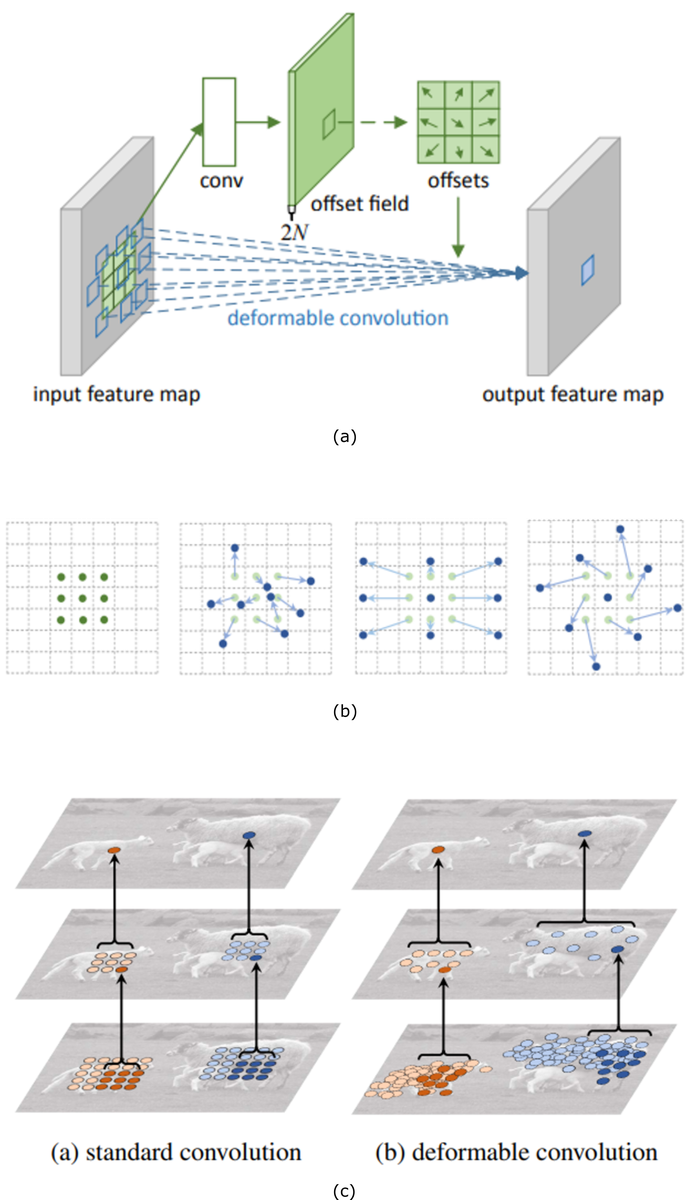}}
\caption{Deformable convolutions. (a) Illustration of a 3x3 deformable convolution. (b) comparison of fixed sampling in plain convolutions \textit{versus} different sampling possibilities in deformable conovlutions. (c) Illustration of fixed receptive field in standard convolution \textit{versus} adaptive receptive field in deformable convolution after the stacking of two layers \cite{b75}.}
\label{fig:deformable_convolutions}
\end{figure}

 Deformable convolutions \cite{b75} enhance the transformation modeling capability of CNNs, by introducing a data-driven mechanism for augmenting the spatial sampling locations in receptive fields – Fig.~\ref{fig:deformable_convolutions}. In summary, the offsets of a given convolutional kernel are learned by a parallel convolutional block. Since the offsets are typically fractional, the final outputs of the deformable convolution are calculated by bilinear interpolation. SGNet \cite{b18} adopts a similar approach, in which depth information is leveraged to guide the sampling offset of the convolutional kernel.

\subsubsection{Feature Pyramids}
\label{subsubsection:accuracy_local_and_global-feature_pyramids}
\begin{figure*}[htbp]
\centerline{\includegraphics[width=\textwidth]{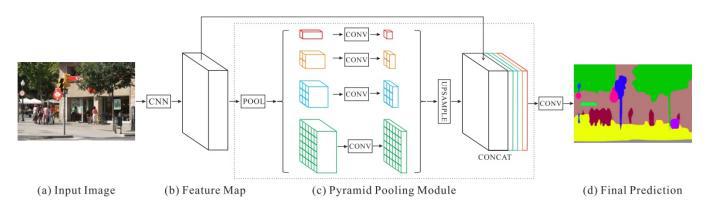}}
\caption{PSPNet architecture. \cite{b3}.}
\label{fig:psp_net}
\end{figure*}

\begin{figure}[htbp]
\centerline{\includegraphics[width=\columnwidth]{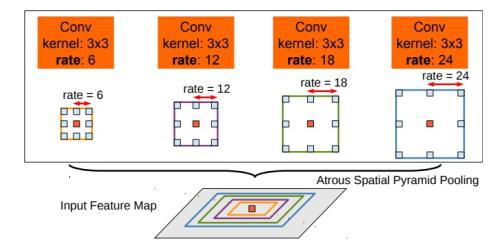}}
\caption{DeepLab architecture. \cite{b5}.}
\label{fig:deeplab}
\end{figure}

Following a similar idea of parallelization of convolutional operations with different sampling ratios, feature pyramids generate several parallel feature maps, extracted using different parameters such as strides and dilation rates, ultimately aggregating them in order to account for both local and global features. PSPNet \cite{b3} applies the so-called Pyramid Parsing Module (PPM) in order to generate feature maps from pooling filters with different sizes – Fig.~\ref{fig:psp_net}. DeepLab \cite{b5} proposes the Atrous Spatial Pyramid Pooling (ASPP), which employs parallel atrous convolutions with different dilation rates – Fig.~\ref{fig:deeplab}. LiteSeg \cite{b76} further develops on the idea of ASPP by increasing its depth with additional convolutions, resulting in the proposed Deeper Atrous Spatial Pyramid Pooling (DASPP) module. In the same line, \cite{b33} proposes the Dilated Asymmetric Pyramidal Fusion (DAPF) where, inspired by ASPP, plain convolutions are replaced by their factorized versions. Yang et al. \cite{b14} propose the DenseASSP module, in which the feature maps obtained with different dilation rates are densely connected. 

\subsubsection{Attention}
\label{subsubsection:accuracy_local_and_global-attention}
Recently, a new trend in research involves the use of attention mechanisms to model long-range dependencies. Non-local Neural Networks \cite{b77} was the first work to bridge self-attention from machine translation to the field of image and video processing in computer vision. 
\par
\begin{figure}[htbp]
\centerline{\includegraphics[width=\columnwidth]{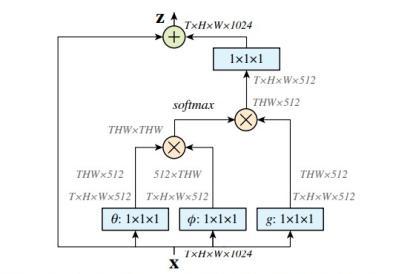}}
\caption{Non-local block. \cite{b77}.}
\label{fig:non_local_block}
\end{figure}

 The basic behavior of attention mechanism involves three components: query, key and value. These are feature embeddings extracted from a given input feature map, usually by a 1x1 convolution, in order to control the channel dimensions of the embeddings – Fig.~\ref{fig:non_local_block}. Query and key are multiplied, and pass through an activation function such as a softmax, so to generate the attention map (sometimes termed affinity matrix). The attention map is then multiplied by the value map, which is, finally, fused with the input feature maps through a mechanism such as addition or concatenation. When this reasoning occurs considering a single feature map, that is, when the attention map models the relationship of every pair of positions – alternatively, channels – in a given feature map, the mechanism is termed self-attention.

\par
\begin{figure}[htbp]
\centerline{\includegraphics[width=\columnwidth]{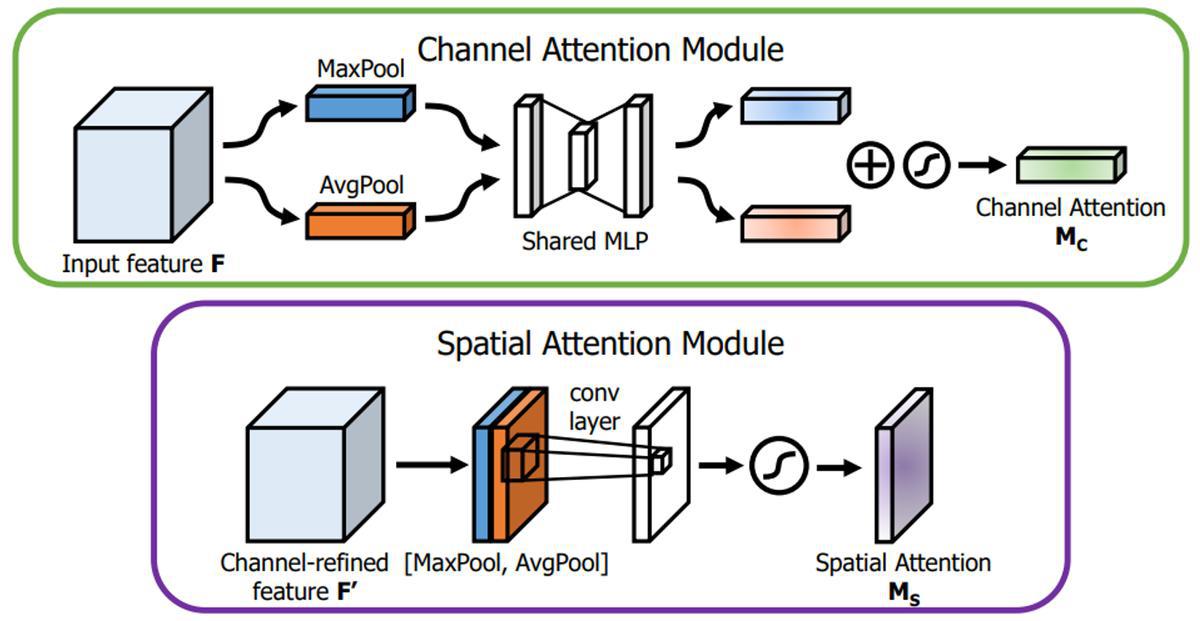}}
\caption{Illustration of channel and spatial attention. (a) In channel attention, features are spatially aggregated to generate an attention map modeling the contribution of channel from the feature given as input. (b) In spatial attention, features are channel-wise aggregated in order to produce a weighting map for each spatial position \cite{b0}. It is worth mentioning that, (I) differently from this example, where attention was built using pooling and MLP, channel and spatial attention can be parameterized by different operations, such as convolutions. (II) This is an example of query-independent attention, where attention maps are computed directly from the input features; its query-dependent version is also possible.}
\label{fig:channel_and_spatial_attention}
\end{figure}

 Attention strategies can implement both channel attention and spatial attention. In channel attention, the contribution of each channel in a given feature map is weighted by aggregating features spatially. Jha et al. \cite{b78}, for instance, apply task-specific channel-attention modules in order to highlight relevant task-specific features. In spatial attention, on the other hand, features are aggregated channel-wise, in order to provide the attention values for each spatial position in a given feature map. An illustration of channel and spatial attention is given in Fig.~\ref{fig:channel_and_spatial_attention}. Query-dependent and query-independent implementations are also possible distinctions \cite{b79}.

\subsection{Alternative Approaches}
\label{subsection:accuracy_alternative}
Applying post-processing steps to the predictions is yet another common practice for accuracy enhancement. Conditional Random Fields (CRFs) are an example of widely-adopted prediction refinement technique. Some examples of works that adopt such a method are the DeepLab family of models \cite{b5}.
\par Feature and label refinement by means of exploring temporal information, as well as depth-aware architectures can also be used for enhanced accuracy – we reserve sections \ref{section:rgb-d_semantic_segmentation} and \ref{section:temporal-aware_semantic_segmentation} to a detailed discussion on these topics.
\par Data-driven behavior, such as spatial-aware operations and optimization compose another line of work which has received attention in recent years. Chen et al. \cite{b18} proposes the use of geometric information to guide both the size of receptive field and the weights of convolutional kernels. Yu et al. \cite{b47} discuss hard-pixel-aware learning techniques. Shi et al. \cite{b80} propose a weighting scheme for the cross-entropy loss based on a discriminatory mask, which gives higher weights to pixels located internally to the instances. Their main intuition is that the border should have more flexibility, since even human annotators can label the data incorrectly, thus inserting noise in the ground truth. Similarly, Zhu et al. \cite{b41} proposed a label relaxation mechanism in order to deal differently with pixels located at boundary regions in the image. A particular case of data-driven behavior refers to dynamic networks. In \cite{b81} an end-to-end dynamic routing framework is proposed to alleviate the scale variance among inputs. In this approach, the forward path is controlled by a data-dependent gating mechanism, which guides the input features towards downsampling, upsampling, or keeping the same spatial resolution. Therefore, small-scale elements tend to be processed by paths where features are upsampled, while large-scale elements are guided through a series of downsampling cells. This mechanism reduces losses caused by excessive or insufficient downsampling.

\par Class-specific segmentation heads have also been used in the literature. \cite{b39} generates prediction heads for each of the classes under consideration, which are then concatenated in the final prediction. \cite{b82} show that using separate heads for each class under consideration boosts the performance at class boundaries and leads to better segmentation of small objects.

\par According to what was previously presented, the search for improved accuracy can result in very complex model architectures, with many hand-engineered pipelines and modules for handling feature extraction and fusion, context aggregation, and scale and appearance issues. This increase in complexity comes with increased computations and memory requirements. These characteristics explain why, in most cases, accuracy-oriented models can’t achieve real-time performance, preventing their application in real scenarios, where computational resources tend to be limited.
\par This practical limitation motivated the current trend into developing efficiency-oriented models, described in the following section.

\section{Efficiency-oriented Semantic Segmentation}
\label{section:efficiency}
Efficiency is a key concern in current research on Deep Learning models applied to realistic scenarios. Many of the contemporary use cases involve embedded devices and edge computing, where hardware has limited capacity in terms of memory and computational power. Additionally, when it comes to safety-critical systems, such as autonomous vehicles, low-latency inference for real-time reasoning is of utmost importance. As a product of such considerations, the research on lightweight models and operations, so that to achieve low computational costs and memory footprint, has gained attention, and can be organized under the class of Efficiency-oriented Deep Semantic Segmentation.
\par In order to provide a structured review, we organize efficiency-oriented approaches according to the level of their design choices into: input-level, architecture-level and operation-level.

\begin{table*}[]
\caption{Summary of the efficiency-oriented works reviewed.}
\label{tab:efficiency_oriented}
\resizebox{\textwidth}{!}{%
\begin{tabular}{cccccccccccccccccc}
\hline
\multirow{2}{*}{Year} &
  \multirow{2}{*}{Model Name} &
  \multirow{2}{*}{Backbone} &
  \multirow{2}{*}{Pre-training} &
  \multirow{2}{*}{Image Resolution} &
  \multirow{2}{*}{Hardware} &
  \multirow{2}{*}{Temporal} &
  \multicolumn{2}{c}{Real-time strategy} &
  \multicolumn{2}{c}{Cityscapes mIoU} &
  FPS &
  Params &
  GFLOPs &
  \multicolumn{3}{c}{Fusion} &
  \multirow{2}{*}{Architecture} \\ \cline{8-11} \cline{15-17}
 &
   &
   &
   &
   &
   &
   &
   &
  Obs &
  Val &
  Test &
   &
   &
   &
  Type &
  Level &
  Mechanism &
   \\ \hline
2017 &
  LinkNet &
   &
   &
  1024x512 &
  Titan X &
   &
  \begin{tabular}[c]{@{}c@{}}Reduced input resolution,\\ Efficient model design\end{tabular} &
  \begin{tabular}[c]{@{}c@{}}Single-stream\\ asymmetric\\ encoder-decoder\end{tabular} &
  76,4 &
   &
  65,8*** &
  11,50 &
  21,20 &
  Multi-level &
  Late &
  Addition &
  Encoder-Decoder \\
2018 &
  ESPNet &
   &
   &
  2048x1024 &
  GTX960M &
   &
  Efficient model design &
  Factorized convolutions &
   &
  60,3 &
  20*** &
  0,40 &
   &
  \begin{tabular}[c]{@{}c@{}}Multi-scale,\\ multi-level\end{tabular} &
  \begin{tabular}[c]{@{}c@{}}Middle,\\ late\end{tabular} &
  Feature pyramid &
  Encoder-Decoder \\
2018 &
  BiseNet &
  Xception39 &
  ImageNet &
  2048x1024 &
  Titan Xp &
   &
  Efficient model design &
  \begin{tabular}[c]{@{}c@{}}Multi-branch\\ asymmetric encoding\end{tabular} &
  69,0 &
  68,4 &
  105,80 &
  5,80 &
  2,9*** &
  \begin{tabular}[c]{@{}c@{}}Multi-level,\\ multi-branch\end{tabular} &
  \begin{tabular}[c]{@{}c@{}}Middle,\\ late\end{tabular} &
  Hand-crafted &
  Multi-branch encoder \\
2018 &
  ICNet &
   &
   &
  2048x1024 &
  Titan X &
   &
  Efficient model design &
  \begin{tabular}[c]{@{}c@{}}Multi-branch\\ asymmetric encoding\end{tabular} &
   &
  70,6***** &
  30,30 &
   &
   &
  Multi-scale &
  Middle &
  Hand-crafted &
  Multi-branch encoder \\
2018 &
   &
  DarkNet &
  ImageNet &
  256x512 &
  1080Ti &
   &
  \begin{tabular}[c]{@{}c@{}}Efficient model design,\\ Reduced input resolution\end{tabular} &
   &
  54,5 &
   &
  61,07*** &
  1,67 &
   &
  Multi-level &
  Late &
  \begin{tabular}[c]{@{}c@{}}Addition,\\ convolution\end{tabular} &
  Encoder-Decoder \\
2018 &
  ERFNet &
   &
  ImageNet &
  1024x512 &
  Titan X &
   &
  Efficient model design &
  Factorized convolutions &
  71,5 &
  69,7 &
  41,67 &
  2,1{[}{]} &
   &
  Multi-level &
  Middle &
  Addition &
  Encoder-Decoder \\
2019 &
  ShelfNet-18-lw &
  ResNet-18 &
   &
  1024x2048 &
  GTX 1080Ti &
   &
  Efficient model design &
  \begin{tabular}[c]{@{}c@{}}Reduced channel depth,\\ shared convolutional weights\end{tabular} &
   &
  74,8 &
  36,90 &
  23,50 &
   &
  Multi-level &
  \begin{tabular}[c]{@{}c@{}}Middle,\\ late\end{tabular} &
  Hand-crafted &
  \begin{tabular}[c]{@{}c@{}}Encoder-Decoder\\ (series of encoding\\ and decoding branches\\  are used, forming a\\  shelf-like structure)\end{tabular} \\
2019 &
  DFANet A' &
  Xception &
  ImageNet &
  512x1024 &
  Titan X &
   &
  Lightweight backbone &
   &
   &
  70,3 &
  160,00 &
  7,80 &
  1,70 &
  Multi-level &
  \begin{tabular}[c]{@{}c@{}}Middle,\\ late\end{tabular} &
  Hand-crafted &
  Encoder-Decoder \\
2019 &
  LiteSeg &
  ShuffleNet &
  \begin{tabular}[c]{@{}c@{}}Cityscapes\\ (coarse)\end{tabular} &
  1024x2048 &
  \begin{tabular}[c]{@{}c@{}}Nvidia GTX \\ 1080Ti\end{tabular} &
   &
  \begin{tabular}[c]{@{}c@{}}Lightweight backbone,\\ Efficient model design\end{tabular} &
   &
  66,1 &
  65,17***** &
  31,00 &
  3,51 &
  2,75*** &
  \begin{tabular}[c]{@{}c@{}}Multi-level,\\ multi-scale\end{tabular} &
  Middle &
  Concatenation &
  Encoder-Decoder \\
2019 &
  FastSCNN &
   &
   &
  1024x2048 &
  Titan X &
   &
  Efficient model design &
  Shared computations &
  69,19 &
  68 &
  62,10 &
  1,01 &
   &
  Multi-branch &
  Middle &
  Addition &
  Encoder-Decoder \\
2019 &
  \begin{tabular}[c]{@{}c@{}}Asymmetric Non-Local\\ Neural Networks\end{tabular} &
  ResNet-101 &
  ImageNet &
  2048×1024 &
  Titan Xp &
   &
  Efficient model design &
  \begin{tabular}[c]{@{}c@{}}Sampling strategy for \\ dimensionality reduction\\ of attention maps\end{tabular} &
   &
  81,3****$^{,}$****** &
  1,63 &
  63,17 &
   &
  \begin{tabular}[c]{@{}c@{}}Multi-level,\\ multi-scale\end{tabular} &
  Middle &
  \begin{tabular}[c]{@{}c@{}}Feature pyramid,\\ attention\end{tabular} &
  Encoder-Decoder \\
2020 &
  SFNet &
  ResNet-18 &
  ImageNet &
  1024x2048 &
  GTX 1080Ti &
   &
  Efficient model design &
   &
   &
  80,4**** &
  26* &
  12,87 &
  123,5*** &
  \begin{tabular}[c]{@{}c@{}}Multi-scale,\\ multi-level\end{tabular} &
  \begin{tabular}[c]{@{}c@{}}Middle,\\ late\end{tabular} &
  \begin{tabular}[c]{@{}c@{}}Feature pyramid,\\ hand-crafted\end{tabular} &
  Encoder-Decoder \\
2020 &
  FANet18+Temp &
  ResNet-18 &
  ImageNet &
  1024x2048 &
  Titan X &
  \begin{tabular}[c]{@{}c@{}}Feature/label\\ refinement,\\ Memory network\end{tabular} &
  \begin{tabular}[c]{@{}c@{}}Lightweight backbone,\\ Efficient model design\end{tabular} &
  Adapted self-attention &
   &
  75,5** &
  72,00 &
   &
  49,00 &
  \begin{tabular}[c]{@{}c@{}}Multi-level,\\ multi-term\end{tabular} &
  Late &
  Attention &
  Encoder-Decoder \\
2020 &
  MSFNet &
  ResNet-18 &
  ImageNet &
  1024x2048 &
  GTX 2080 Ti &
   &
  \begin{tabular}[c]{@{}c@{}}Lightweight backbone,\\ EffIcient model design\end{tabular} &
  Factorized convolutions &
   &
  77,1 &
  41,00 &
   &
  96,80 &
  Multi-level &
  Middle &
  Hand-crafted &
  \begin{tabular}[c]{@{}c@{}}Multi-branch decoder\\ (decoder for class boundary\\  supervision used only during\\ training - auxiliary-task learning)\end{tabular} \\
2020 &
  SwiftNet-18 pyr &
  ResNet-18 &
  \begin{tabular}[c]{@{}c@{}}ImageNet and \\ Mapillary\end{tabular} &
  1024x2048 &
  GTX 1080Ti &
   &
  \begin{tabular}[c]{@{}c@{}}Lightweight backbone,\\ Efficient model design\end{tabular} &
  Asymmetric Encoder-Decoder &
   &
  76,4 &
  34,00 &
  12,00 &
  64,00 &
  \begin{tabular}[c]{@{}c@{}}Multi-scale,\\ multi-level\end{tabular} &
  \begin{tabular}[c]{@{}c@{}}Middle,\\ late\end{tabular} &
  \begin{tabular}[c]{@{}c@{}}Addition,\\ convolution\end{tabular} &
  Multi-branch encoder \\
2020 &
  PBRNet &
  MobileNetV2 &
  ImageNet &
  1024x2048 &
  Titan XP &
   &
  \begin{tabular}[c]{@{}c@{}}Lightweight backbone,\\ Efficient model design\end{tabular} &
   &
   &
  72,4 &
  93,46 &
  2,14 &
  12,80 &
  Multi-level &
  Late &
  Attention &
  Encoder-Decoder \\
2020 &
  FSFNet &
   &
   &
  1024x2048 &
  1080Ti &
   &
  Efficient model design &
  Single-Stream Network &
  69,72 &
  69,1 &
  203* &
  0,82 &
  13,47 &
  Multi-level &
  Middle &
  Addition &
  Encoder-Decoder \\
2020 &
  FasterSeg &
   &
   &
  1024x2048 &
  \begin{tabular}[c]{@{}c@{}}Nvidia Geforce \\ GTX 1080Ti\end{tabular} &
   &
  Efficient model design &
   &
  73,1 &
  71,5 &
  163,9* &
  4,40 &
  28,20 &
  Multi-level &
   &
  \begin{tabular}[c]{@{}c@{}}Concatenation,\\ convolution\end{tabular} &
  Neural Architecture Seach \\
2020 &
  FASSDNetL2 &
  FC-HarDNet-70 &
  ImageNet &
  1024x2048 &
  GTX 1080TI &
   &
  \begin{tabular}[c]{@{}c@{}}Lightweight backbone,\\ Efficient model design\end{tabular} &
  Factorized convolutions &
  77,4 &
  74,1** &
  133,10 &
  2,30 &
  8,70 &
  \begin{tabular}[c]{@{}c@{}}Multi-level,\\ multi-scale\end{tabular} &
  \begin{tabular}[c]{@{}c@{}}Middle,\\ late\end{tabular} &
  Hand-crafted &
  Encoder-Decoder \\
2020 &
  Dynamic* &
   &
   &
  1024x2048 &
   &
   &
  Efficient model design &
  Budget constraint in loss function &
  78,3 &
  79,1**** &
   &
  17,80 &
  111,70 &
  \begin{tabular}[c]{@{}c@{}}Multi-scale,\\ multi-level\end{tabular} &
   &
  Gating, addition &
  Dynamic Networks \\
2021 &
  AttaNet &
   &
   &
  1024x2048** &
  GTX 1080Ti &
   &
  Efficient model design &
  \begin{tabular}[c]{@{}c@{}}Adaped attention mechanism\\ (Strip Attention Module)\end{tabular} &
  80 &
  79,9 &
  71,00 &
  20,08 &
  49,23 &
  Multi-level &
  Late &
  Gating &
  Encoder-Decoder \\
2021 &
  BiAlignNet &
  DFNet2 &
  Mapillary &
  768x1536 &
  1080 Ti &
   &
  Efficient model design &
   &
  79 &
  76,9 &
  50,00 &
  19,20 &
  108,73 &
  \begin{tabular}[c]{@{}c@{}}Multi-scale,\\ multi-branch\end{tabular} &
  Middle &
  \begin{tabular}[c]{@{}c@{}}Feature pyramid,\\ hand-crafted/gating\end{tabular} &
  \begin{tabular}[c]{@{}c@{}}Multi-branch encoder\\ (Context and Detail paths)\end{tabular} \\
2021 &
  DDRNet-23-Slim &
   &
   &
  2048x1024 &
  GTX 2080Ti &
   &
  Efficient model design &
   &
  77,8 &
  77,4**** &
  101,60 &
  36,30 &
  5,70 &
  \begin{tabular}[c]{@{}c@{}}Multi-branch,\\ multi-level,\\ multi-scale\end{tabular} &
  Middle &
  Addition, concatenation &
  \begin{tabular}[c]{@{}c@{}}Multi-branch encoder\\ (Deep Dual-Resolution paths)\end{tabular} \\
2021 &
  EKENet-A &
   &
   &
  1024x2048 &
  Titan X &
   &
  Efficient model design &
  Factorized Convolutions &
   &
  71,8 &
  65,40 &
  0,40 &
  31,70 &
  Multi-level &
  Late &
  \begin{tabular}[c]{@{}c@{}}Convolution,\\ multiplication\end{tabular} &
  Encoder-Decoder \\
2021 &
  FSFNet &
  ResNet-18 &
  ImageNet &
  1024x2048 &
  NVIDIA 1080 TI &
   &
  \begin{tabular}[c]{@{}c@{}}Lightweight backbone,\\ EffIcient model design\end{tabular} &
   &
   &
  77,1 &
  39,00 &
  16,00 &
  45,80 &
  \begin{tabular}[c]{@{}c@{}}Multi-level,\\ multi-scale\end{tabular} &
  \begin{tabular}[c]{@{}c@{}}Middle,\\ late\end{tabular} &
  Hand-crafted/gating &
  Encoder-Decoder \\
2021 &
  SegFormer (MiT-B0) &
  \begin{tabular}[c]{@{}c@{}}Mix Transformer \\ encoders (MiT)\end{tabular} &
  ImageNet &
  1024x1024 &
  (8) Tesla V100 &
   &
  Efficient model design &
  Adapted self-attention &
  71,9 &
   &
  47,60 &
  3,80 &
  17,70 &
  Multi-level &
  Late &
  Hand-crafted &
  Transformer-based Encoder-Decoder \\
2021 &
  STDC1-Seg50 &
  STDC1 &
  ImageNet &
  512x1024 &
  GTX 1080 Ti &
   &
  Efficient model design &
   &
  72,2 &
  71,9 &
  250,4* &
  8,40 &
  0,81 &
  Multi-level &
  Middle &
  Hand-crafted/gating &
  Encoder-Decoder \\
2021 &
  Faster BiseNet &
   &
   &
  1024×512 &
  \begin{tabular}[c]{@{}c@{}}NVIDIA GeForce\\ 1080 Ti\end{tabular} &
   &
  Efficient model design &
   &
  73,7 &
  72,8 &
  187* &
  3,23 &
  20,40 &
  Multi-level &
  Late &
  Gating &
  Encoder-Decoder \\
2021 &
  BiseNet V2 &
   &
   &
  1024x512 &
  \begin{tabular}[c]{@{}c@{}}NVIDIA GeForce\\ 1080Ti\end{tabular} &
   &
  Efficient model design &
   &
  73,4 &
  72,6 &
  156* &
   &
  21,15 &
  Multi-branch &
  Late &
  Gating &
  Multibranch Encoder \\
2021 &
  FPENet &
   &
   &
  2048x1024 &
  Nvidia Titan V100 &
   &
  Efficient model design &
   &
   &
  73,4 &
  116,20 &
  0,40 &
  3,29 &
  \begin{tabular}[c]{@{}c@{}}Multi-scale,\\ multi-level\end{tabular} &
  \begin{tabular}[c]{@{}c@{}}Middle,\\ late\end{tabular} &
  Hand-crafted &
  \begin{tabular}[c]{@{}c@{}}Asymmetric Encoder-Decoder\\ (Neural Architecture Search)\end{tabular} \\
2021 &
  EACNet &
   &
   &
  512x1024 &
  \begin{tabular}[c]{@{}c@{}}NVIDIA RTX \\ 2080Ti\end{tabular} &
   &
  Efficient model design &
  \begin{tabular}[c]{@{}c@{}}Shared computations,\\ asymmetric convolutions\\ and simple fusion\end{tabular} &
   &
  72,4 &
  113,00 &
  1,10 &
   &
  \begin{tabular}[c]{@{}c@{}}Multi-branch,\\ multi-level\end{tabular} &
  \begin{tabular}[c]{@{}c@{}}Middle,\\ late\end{tabular} &
  Hand-crafted &
  Asymmetric Encoder-Decoder \\
2022 &
  SGCPNet1 &
   &
   &
  1024x2048 &
  \begin{tabular}[c]{@{}c@{}}NVIDIA GeForce\\ 1080Ti\end{tabular} &
   &
  Efficient model design &
   &
   &
  70,9 &
  103,70 &
  0,61 &
  4,50 &
  Multi-level &
  Late &
  Hand-crafted &
  Encoder-Decoder \\
2022 &
  PCNet* &
   &
   &
  1024x2048 &
  GTX 2080 Ti &
   &
  Efficient model design &
  \begin{tabular}[c]{@{}c@{}}Fast downsampling and\\ factorized convolutions\end{tabular} &
   &
  72,9 &
  79,10 &
  1,49 &
  11,50 &
  \begin{tabular}[c]{@{}c@{}}Multi-scale,\\ multi-level\end{tabular} &
  \begin{tabular}[c]{@{}c@{}}Middle,\\ late\end{tabular} &
  Hand-crafted &
  Encoder-Decoder \\ \hline
\end{tabular}%
}
\end{table*}

\subsection{Input-level Strategies}
\label{subsection:efficiency_input_level}
Input-level strategies are straightforward, and involve reducing the input resolution. This leads to reduced feature maps and, consequently, fewer computations and memory consumption. This can be achieved either by resizing the entire image \cite{b18}, or by randomly selecting crops of the input image \cite{b19} \cite{b41} \cite{b20}. A particularly interesting strategy is proposed in ICNet \cite{b9}, where a cascade image input, composed of the original image resolution, as well as its two down-sampled copies, are fed into resolution-specific encoders. Efficiency is achieved by controlling the depth of each encoder according to the specific input resolution. Lower-resolution inputs are processed by deeper encoders in order to extract semantic information, since operations will naturally involve lower-resolution feature maps. On the other hand, the full-resolution input is used to extract spatial detailed information by a shallower encoder to keep computational costs low, since the feature maps generated are larger. 
\par Nonetheless, simply reducing the input resolution may not be suitable to critical applications, such as vision for autonomous vehicles and driver assistance systems, since it causes loss of information related to small classes; in fact, there are evidences that semantic segmentation drastically improves with high-resolution images \cite{b7} \cite{b57}.

\subsection{Architecture-level strategies}
\label{subsection:efficiency_architecture_level}
Architecture-level strategies generally involve modifying the network architecture in order to reduce the number of computations and parameters to be optimized/learned. 

\subsubsection{Single-stream networks}
\label{subsubsection:efficiency_architecture_level-single-stream}
Adopting single-stream architectures is the most straightforward way to keep computations under control \cite{b83}. For instance, Li et al. \cite{b74} adopt a single-stream encoder-decoder network by employing a simplified version of AlexNet \cite{b84}. However, relying on simpler architectures may limit the model ability of extracting more meaningful features out of the available data.

\subsubsection{Asymmetric architectures}
\label{subsubsection:efficiency_architecture_level-asymmetric_architectures}
The notion that encoders and decoders do not necessarily need to be symmetric – in encoder-decoder architectures – raises another possible approach to high-speed models. Since the decoder’s main role is to simply up-sample the embedded representation to match the input resolution, it can be kept simple; the encoder, on the other hand, should be able to extract meaningful features out of the inputs, hence heavier computation is allowed. This reasoning leads to asymmetric encoder-decoder architectures, where the decoder is shallower – with less layers and learnable parameters – than the encoder \cite{b64} \cite{b66} \cite{b83} \cite{b76}.
\par Following a similar idea, multi-branch asymmetric methods \cite{b9} \cite{b22} try to find a good balance between extraction capabilities and inference time, usually by applying separate asymmetric encoder branches for context and detail extraction. In ICNet \cite{b9}, different inputs are parallelly processed by different encoder branches, with structures directly related to their specific input resolutions. In BiseNet \cite{b22} and BiseNet V2 \cite{b13} the same input is processed by different branches, termed spatial and context branches. While the first has shallower architecture and wider channels for spatial detail extraction, the second has narrow channels and deeper layer structure for contextual/semantic extraction.

\subsubsection{Lightweight backbones}
\label{subsubsection:efficiency_architecture_level-lightweight_backbones}
Another common technique is to construct models on top of lightweight pre-trained backbones, such as the MobileNet family of models \cite{b85} \cite{b86} \cite{b87}. Although providing the advantage of leverage pre-trained weights, inserting new elements to their architectures can be challenging, and fine-tuning is required for minimizing issues related to domain shift. Hu et al. \cite{b21}, Si et al. \cite{b15}, and Pei et al. \cite{b68} adopt ResNet-18 \cite{b88} as a lightweight backbone. BiseNet \cite{b22} and DFANet \cite{b19} adopt modified versions of Xception \cite{b89} as their backbones. Sun et al. \cite{b72} adopted MobileNet V2, while Hao et al \cite{b90} uses MobileNet V3 as their lightweight backbone. The authors of SwiftNet \cite{b66} adopt ResNet-18 and MobileNet V2 as lightweight backbones in their experiments. LiteSeg run experiments using MobileNet, ShuffleNet \cite{b91} and DarkNet19 \cite{b92} as lightweight backbones.

\subsubsection{Hand-crafted modules and models}
\label{subsubsection:efficiency_architecture_level-hand_crafted}
Other approaches, also related to the model backbone, rely on designing hand-crafted modules or even entire backbones from scratch, according to the application’s specific needs. Hong et al. \cite{b24} propose a family of networks termed DDRNet, where dual-resolution branches with multi-level channel fusion, and pyramid pooling context aggregation are used in order to keep computations low while attaining high accuracy. Faster BiSeNet \cite{b25} and STDC Network \cite{b26} proposed efficient architectures based on the previous BiSeNet model. In FSFNet \cite{b83}, the authors provide a detailed study on indirect and direct measures of convolution operations – and their combination – in order to define their architecture in the most efficient way possible. Although very flexible, the main disadvantage of such approach is the need for training the model from scratch, missing a huge regularization opportunity offered by knowledge transfer from larger and more diverse recognition datasets \cite{b66}; besides that, hand-crafted architectures are more prone to issues like underfitting or overfitting, depending on the amount and representativeness of the data available for training. Contrarily, though, Poudel et al. \cite{b27} show that ImageNet pre-training, or even training with additional weakly-labeled data (e.g. coarse labels on Cityscapes) provides minimal benefits for low-capacity models implemented by lightweight designs.
\par Neural Architecture Search (NAS) has also been used for efficient model design. In FasterSeg \cite{b93}, the authors explore multi-resolution branches, as well as a latency-based regularization term and a co-searching strategy for simultaneously searching for a teacher and a student network. In FPENet \cite{b16}, NAS is applied to find the optimal structure for each Feature Pyramid Encoding (FPE) block used for feature aggregation. One drawback of NAS is that is does not allow leveraging pre-training.

\subsubsection{Reduced feature dimensions}
\label{subsubsection:efficiency_architecture_level-feature_dimensions}
Keeping computations low by reducing feature size is also an alternative, generally achieved by fast down-sampling \cite{b8} or by reducing the depth or spatial size of feature maps \cite{b21} \cite{b94}\cite{b95}. Shi et al. \cite{b96} uses a 1×1 convolution layer before a ConvLSTM module to reduce the features’ depth. In FANet \cite{b21}, the authors explore inserting an additional down-sampling stage to their backbone in order to reduce computations. Yu et al. \cite{b13} conduct experiments investigating channel capacity, block design, and expansion ratio of their semantics branch. In a recent work, Fan et al. \cite{b26} argue that, on the contrary of what is widely adopted in the literature, having deeper layers with a high number of channels solely increases the redundancy in the network. Conversely, a better approach would be to extract as much information as possible in lower-level layers (therefore, with more feature maps/channels), and just tune this information while keeping a reduced number of channels as the depth of the network increases. 
\par The main drawback of such approaches is reduced learning capability. In the case of reducing feature size, spatial and detailed information is lost, while in the second (reducing the depth of feature maps) modeling ability is harmed. In order to overcome such problems and provide a good trade-off between accuracy and efficiency, feature reuse, fusion and aggregation schemes are leveraged to improve feature representativeness. Hao et al. \cite{b90} study the influence of maintaining high-resolution feature maps on metrics such as FLOPs and runtime. They find that an increase by a factor of 8 in feature resolution results in up to 20 times more FLOPs and runtime latency on a ResNet-50 model – the results are even worse for deeper architectures, such as ResNet-101. To deal with the problem, the authors adopt a lightweight encoder, progressively down-sampling features to 1/32 of input resolution and, in order to improve modeling capabilities, propose a complex scheme for feature aggregation, composed by top-down and bottom-up feature refinement.

\subsubsection{Layer and weight sharing}
\label{subsubsection:efficiency_architecture_level-weight_sharing}
Layer and weight-sharing architectures generally share computation in order to reduce complexity. Most of the models lying in this category have a common encoder, and multiple decoder heads, characterizing multi-task learning frameworks. There are also approaches that share subsets of layers from the encoder. Faster BiseNet \cite{b25} and STDC Net \cite{b26} share the first N out of the M total layers of the encoder. While the output extracted from the Nth layer is used as input to edge head \cite{b25} and detail head \cite{b26}, the following M-N layers are used to extract contextual features. Finally, features corresponding to both levels in the architecture (spatial and contextual) are fused by specific feature fusion modules. Similarly, in EACNet \cite{b11} a shared initial block is used for feature extraction in order to feed parallel branches for detail and context reasoning. In Fast-SCNN \cite{b27}, a “learning to downsample” module is proposed, in which shared low-level feature extraction is performed in order to feed multi-resolution branches simultaneously.

\subsubsection{Feature reuse}
\label{subsubsection:efficiency_architecture_level-feature_reuse}
Feature reuse is usually leveraged in video-based semantic segmentation frameworks \cite{b28}\cite{b29}\cite{b30} (section \ref{section:temporal-aware_semantic_segmentation}). The main intuition behind this technique comes from the fact that between adjacent frames there is no considerable change in appearance and, hence, in semantic information. Therefore, propagation strategies can be applied to reuse both features and labels, warping them from key frames to non-key or adjacent frames. In \cite{b28} a hybrid CPU-GPU approach is used to leverage optical flow in label propagation and refinement. \cite{b38} uses optical flow to propagate features from key frames, extracted by a heavy architecture, to non-key frames; computational costs are diminished by adopting a lightweight update network, which combined with the warped version of the previously computed keyframe features, delivers the final segmentation. \cite{b30} propagates high-level features from keyframes to neighboring frames by using spatially-adaptive convolutional weights.

\subsubsection{Knowledge distillation}
\label{subsubsection:efficiency_architecture_level-knowledge_distillation}
\begin{figure}[htbp]
\centerline{\includegraphics[width=\columnwidth]{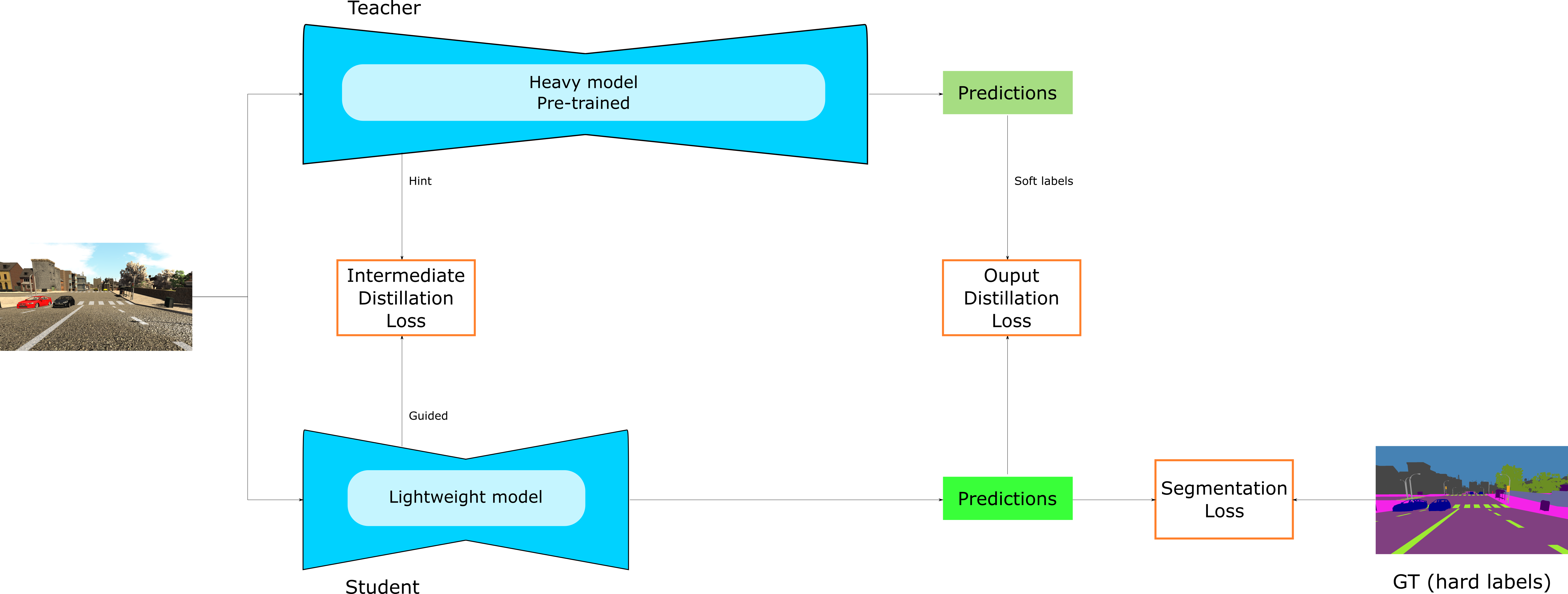}}
\caption{Knowledge distillation setup.}
\label{fig:knowledge_distillation}
\end{figure}

Knowledge distillation is also frequently used to reduce model complexity. It involves a teacher and a student network. The teacher is generally a heavy and complex model, while the student is a reduced, more efficient version of it – the common practice is to use separate networks as teacher and student. The main goal is to translate the behavior of the teacher network into the student network in order to obtain a lightweight and efficient network, with minimal loss in accuracy (Fig.~\ref{fig:knowledge_distillation}). It can be done by aligning the models’ outputs \cite{b97} \cite{b98} or by aligning both outputs as well as some of their intermediate stages \cite{b36} \cite{b37}.

\subsubsection{Dynamic networks}
\label{subsubsection:efficiency_architecture_level-dynamic_netowkrs}
Finally, an interesting direction is to apply dynamic networks. In such models, the processing flow is controlled by gates, so that the model chooses the best path to follow, and doesn’t get stuck in a fixed structure. In DVSNet \cite{b99}, a decision network is used to guide the processing of frame regions through heavier or lighter CNNs, according to a confidence score that represents the amount of change in each region. In \cite{b81} an end-to-end dynamic routing framework is proposed to alleviate the scale variance among inputs. Efficiency is explored through the definition of a budget constraint in the loss function, based on the expected cost of the nodes in a given route.

\subsection{Operation-level strategies}
\label{subsection:efficiency_operation_level}
Operation-level strategies remodel or build operations from scratch in order to reduce computation costs and memory requirements, while maintaining acceptable accuracy levels.

\subsubsection{Convolution factorization}
\label{subsubsection:efficiency_architecture_level-conv_factorization}
\begin{figure*}[htbp]
\centerline{\includegraphics[width=\textwidth]{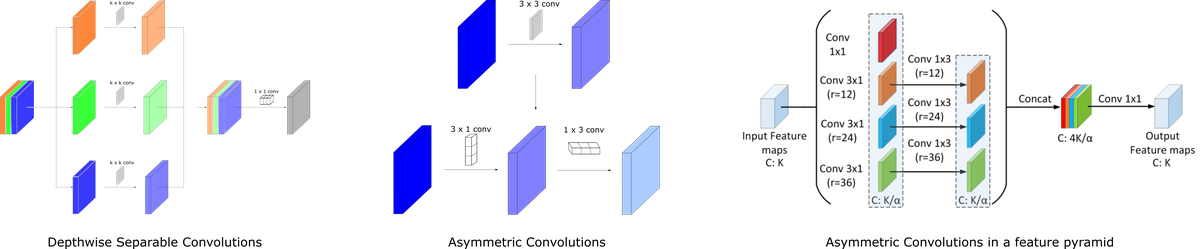}}
\caption{Factorized convolutions. From left to right: depthwise separable convolutions, asymmetric convolutions, and an interesting application of asymmetric convolutions to an ASPP module. \cite{b33}.}
\label{fig:factorized_convolutions}
\end{figure*}

Depthwise Separable Convolutions \cite{b89} and Asymmetric Convolutions \cite{b31}\cite{b32} \cite{b33} are common techniques in the literature that, by factorizing and rearranging convolution operations, result in considerably less computations (Fig.~\ref{fig:factorized_convolutions}). ERFNet \cite{b31} proposes a non-bottleneck-1D residual layer in which plain convolutions are replaced by a sequence of 1D convolutions – for 3x3 convolutions, this decomposition results in 33\% reduction in parameters. FASSD-Net \cite{b33}, factorizes 3x3 convolutions into two consecutive 1D convolutions - 1x3 followed by 3x1 convolution; this reduces computational complexity and also helps in feature learning, because of the non-linear activations in between the convolutions. In EACNet \cite{b11}, Enhanced Asymmetric Convolution modules apply depth-wise asymmetric convolution and dilated convolution to enhance feature extraction, while keeping computations under control. In ESPNet \cite{b100}, the Efficient Spatial Pyramid module is proposed for feature aggregation over multiple scales. The authors apply convolution factorization – point-wise convolution followed by a spatial pyramid of dilated convolutions – and hierarchical feature fusion in order to reduce computational costs and gridding artifacts.

\subsubsection{Simplified fusion}
\label{subsubsection:efficiency_architecture_level-simplified_fusion}
As discussed in section \ref{subsection:accuracy_improve_feature}, it is a common practice in the literature on Deep Semantic Segmentation to use some type of feature fusion method. In order to keep computational costs under control, several works adopt simpler fusion operations, such as element-wise addition \cite{b64} \cite{b66} and channel-wise concatenation \cite{b19} \cite{b11}. More sophisticated gating and attention mechanisms for feature fusion, despite yielding better results, incur more memory consumption and computations \cite{b20}, and might fail in recovering local and multi-scale contextual information \cite{b68}. To tackle this problem, Hu et al. \cite{b21} proposed a modification to the self-attention mechanism in order to capture the same spatial context, but with a fraction of the computational cost, by changing the order of the operations. In TDNet \cite{b36}, the size of attention maps is reduced to save computations. LMANet \cite{b101} proposes an attention mechanism with limited search region in order to save memory and computation. Zhu et al. \cite{b102} propose the Asymmetric Non-local Neural Network, in which a sampling strategy is applied to key and value generation to reduce computational complexity of the attention mechanism. In order to account for long-range relationships, the authors further explore pyramidal pooling for feature aggregation, since pooling operations do not add learnable parameters.

\subsubsection{Reduced learnable parameters}
\label{subsubsection:efficiency_architecture_level-reduced_learnable_parameters}
Reducing the number of learnable parameters is also suitable in operation-level approaches. An example of such a method involves setting pre-defined filter weights, so to capture certain correspondences among feature maps \cite{b29}. This approach frees the model of the burden of learning the convolutional weights by inserting priors in it; nonetheless, this technique harms model’s flexibility and learning ability.

\subsubsection{Selective feature processing}
\label{subsubsection:efficiency_architecture_level-selective_feature_processing}
\begin{figure}[htbp]
\centerline{\includegraphics[width=\columnwidth]{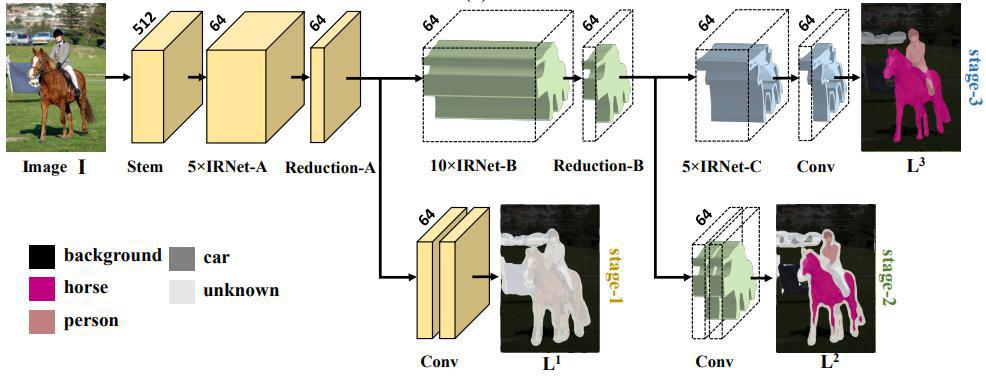}}
\caption{Illustration of selective feature processing, where only regions with high uncertainty are propagated to the next processing stage of the network. \cite{b34}.}
\label{fig:not_all_pixels_are_equal}
\end{figure}

Another interesting approach deals with selective feature processing. The main intuition behind this technique is that not all image regions must be processed in every stage of the network. \cite{b103} performs adaptive key and query selection in order to give more importance to difficult regions in the image – regions with diverse number of classes. \cite{b34} proposes a mechanism where only regions with high uncertainty are propagated for further processing in subsequent layers of the network – Fig.~\ref{fig:not_all_pixels_are_equal}.

\subsection{Auxiliary Training Strategies}
\label{subsection:efficiency_auxiliary_training}
There is also a line of work in efficient model design that explores alternatives for improving model accuracy during training, while keeping model efficiency during inference. The main intuition here is to explore computational power and memory availability during training by leveraging training techniques and additional structures for accuracy improvement, and then keep only the necessary components during inference.
\par Pre-training and auxiliary training strategies – sometimes termed deep supervision – are some examples. Besides providing regularization, such strategies help in model optimization, especially when the amount of labeled data is limited in the target domain. Siam et al. \cite{b60} discuss pre-training with coarse segmentations – such as the ones provided in Cityscapes – as a way to improve model accuracy. \cite{b65}, \cite{b66}, \cite{b104} and \cite{b41} show that pre-training on datasets with domains similar to the target domain – e.g. pre-training on Mapillary Vistas \cite{b105} and then testing on Cityscapes \cite{b62} – can further improve segmentation by providing better initialization. 
\par
\begin{figure}[htbp]
\centerline{\includegraphics[width=\columnwidth]{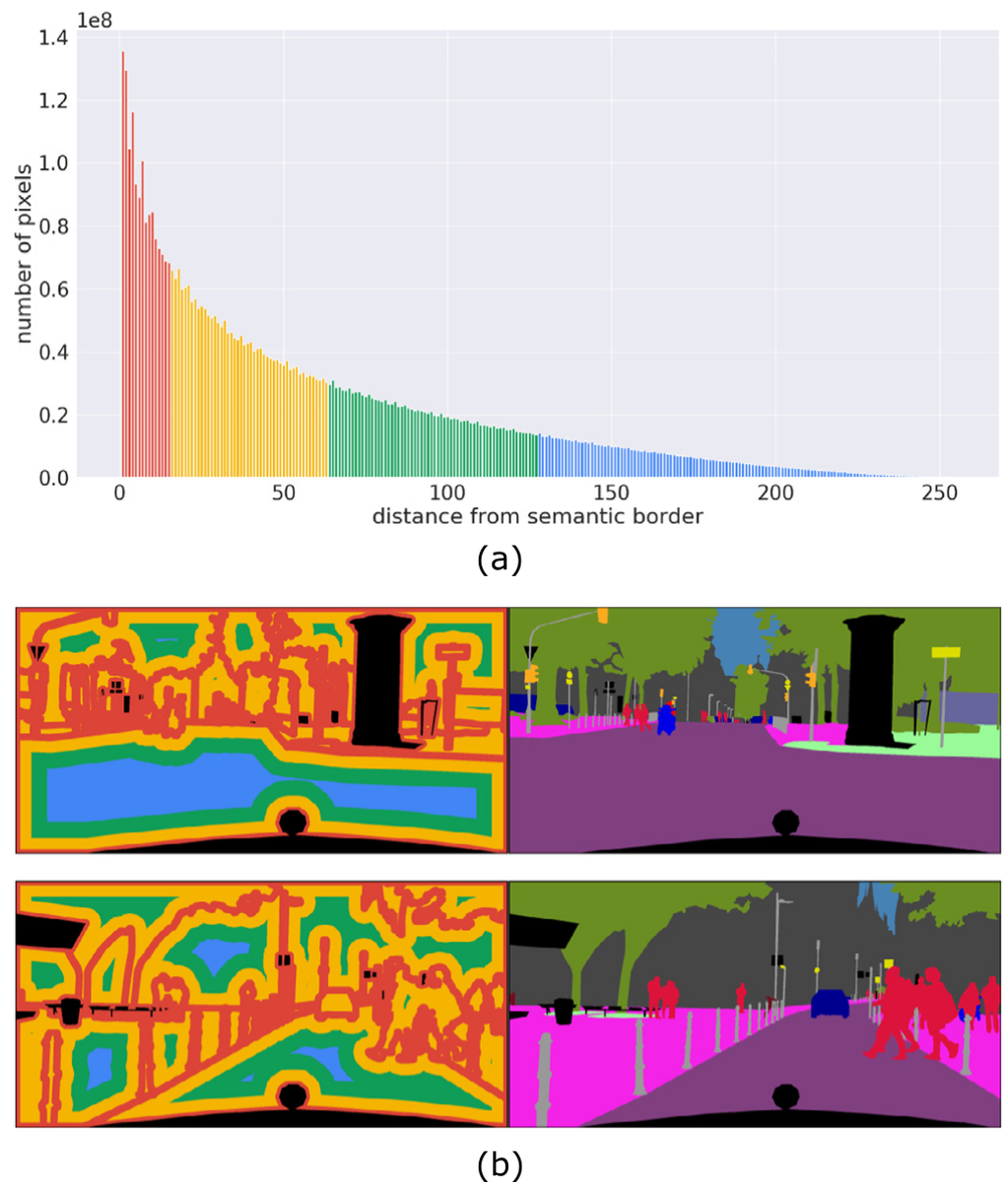}}
\caption{Illustration of the weights attributed to pixels in a boundary-aware loss \cite{b66}. (a) Histogram of distances to the closest semantic border on Cityscapes train at
full resolution; each color designate a loss weight. (b) Pixels closer to the semantic boundaries receive more weight when compared to pixels lying inside scene elements.}
\label{fig:boundary_aware_loss}
\end{figure}

 Wu et al. \cite{b12}, besides proving the benefit of pre-training using similar domains, apply deep supervision by an edge detection head in order to improve feature extraction in their Spatial Path. Si et al. \cite{b15} apply Class Boundary Supervision to avoid the loss of edge-related spatial information. Hong et al. \cite{b24} and Sun et al. \cite{b72} design intermediate segmentation heads as auxiliary supervision signals during training. In BiseNet V2 \cite{b13} and Faster BiSeNet \cite{b25}, the authors propose a booster training strategy where auxiliary segmentation heads are coupled to different levels of their semantic branch, in order to further improve the segmentation performance without increasing inference costs. In \cite{b25} there is also an edge head for edge detection, in order to improve the extraction capabilities of the spatial branch. In the same line of thought, Zhao et al. \cite{b9} propose a Cascade Label Guidance mechanism in order to enhance the learning ability in each of the resolution-specific branches of their model ICNet. In SwiftNet \cite{b66}, motivated by the finding that a huge number of pixels lie in positions close to class borders, a boundary-aware loss is proposed to give more weight to pixels closer to the boundaries (Fig.~\ref{fig:boundary_aware_loss}). In \cite{b106}, the authors propose a loss function composed by four terms: warp loss, semantic loss, edge loss, and edge-semantics loss, in order to better optimize the model.

\hfill \break
\par It is worth noting that the techniques discussed in the previous sections are not applied in isolation; instead, they are used in conjunction in great part of the works previously described. For instance, in \cite{b15}, besides using a lightweight backbone, the authors design efficient modules for feature fusion and context aggregation based on factorized convolutions. As another example, Chaurasia et al. \cite{b64} propose a single-stream network composed by an asymmetric encoder-decoder with skip connections for addition-based feature fusion.

\section{RGB-D Semantic Segmentation}
\label{section:rgb-d_semantic_segmentation}
When it comes to the navigation in complex urban environments, relying on the use of 2D RGB data as the only input modality for semantic segmentation can limit the network’s accuracy, since in some scenarios texture and color cannot fully describe the environment. In this respect, depth information has been increasingly used as an auxiliary source of geometric information towards a better understanding of the scene structure.
\par However, directly applying depth information into existing RGB frameworks may lead to inferior performance. Wang et al. \cite{b50} argue that the considerable variations between RGB and depth modalities and the uncertainty of depth measurements are the two main challenges for RGB-D semantic segmentation. Additionally, variations in scene illumination, and incomplete representation of objects due to complex occlusions are other challenging factors \cite{b58}. In this regard, Wang et al. \cite{b50} highlight as main research points for RGBD semantic segmentation: how to fuse RGB and depth information, and how to integrate depth information into the convolution operation - so that the convolution has depth perception capabilities. Besides that, how to achieve real-time RGBD semantic segmentation is also an important research objective, especially in autonomous systems.
\par Therefore, how to properly identify, represent, and fuse the complementarity of both modalities, as well as how to integrate depth into the network structure while guaranteeing efficiency, are the main points addressed in current research on RGB-D semantic segmentation.
\par Throughout the years, several categorizations were proposed for Deep RGB-D Semantic Segmentation literature, according to the strategy used for embedding depth into the network. Hu et al. \cite{b53} divide Deep RGB-D semantic segmentation into channel concatenation-based approaches - RGB-Depth fusion before feature extraction - and multi-branch networks - modality-specific branches fused at the end of the network. Other works, such as \cite{b50} and \cite{b52} divide RGB-Depth fusion methods into three categories, according to the level of the network in which the fusion occurs: early, middle and late fusion – image layer fusion, feature layer fusion and output layer fusion are also possible classifications. In addition to that, several works employ multi-level feature fusion \cite{b82} \cite{b19}. 
\par
\begin{figure*}[htbp]
\centerline{\includegraphics[width=\textwidth]{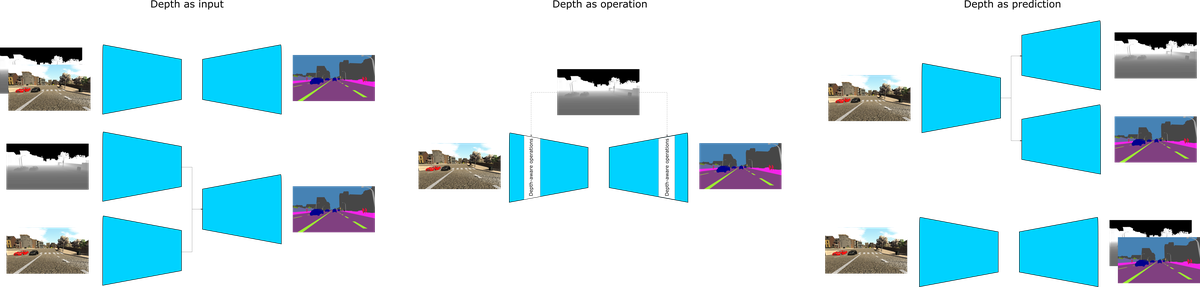}}
\caption{Auxiliary (a) and multi-task (b) learning setups. \cite{b0}.}
\label{fig:depth_as_input_op_pred}
\end{figure*}

 The taxonomy we adopt in this work is closer to the one proposed by Barchid et al. \cite{b51}. The authors classify RGB-D methods according to how depth data is used in the network, into: depth as input, depth as operation and depth as prediction - Fig.~\ref{fig:depth_as_input_op_pred}.  In addition to that, and considering recent developments, we propose three new categories on the use of depth information in RGB-D semantic segmentation: depth as pre-training, depth as augmentation and depth as loss regularizer.

\begin{table*}[]
\caption{Summary of the depth-aware works reviewed.}
\label{tab:depth_aware}
\resizebox{\textwidth}{!}{%
\begin{tabular}{ccccccccccccccccccc}
\hline
\multirow{2}{*}{Year} &
  \multirow{2}{*}{Model Name} &
  \multirow{2}{*}{Backbone} &
  \multirow{2}{*}{Pre-training} &
  \multirow{2}{*}{Image Resolution} &
  \multirow{2}{*}{Hardware} &
  \multicolumn{2}{c}{Depth} &
  \multirow{2}{*}{Temporal} &
  \multirow{2}{*}{Real-time strategy} &
  \multicolumn{2}{c}{Cityscapes} &
  \multirow{2}{*}{FPS} &
  \multirow{2}{*}{Params} &
  \multirow{2}{*}{GFLOPs} &
  \multicolumn{3}{c}{Fusion} &
  \multirow{2}{*}{Architecture} \\ \cline{7-8} \cline{11-12} \cline{16-18}
 &
   &
   &
   &
   &
   &
   &
  Obs &
   &
   &
  Val &
  Test &
   &
   &
   &
  Type &
  Level &
  Mechanism &
   \\ \hline
2018 &
  PADNet &
  ResNet101 &
  ImageNet &
  640x640 &
  Titan X &
  Depth as Prediction &
   &
   &
   &
   &
  80,30 &
   &
   &
   &
  Multi-modal &
  Middle &
  Gating &
  Multi-Branch decoder \\
2018 &
   &
  ResNet101 &
  ImageNet &
  800x800 &
  Titan X &
  \begin{tabular}[c]{@{}c@{}}Depth as Prediction,\\ Depth as operation\end{tabular} &
  \begin{tabular}[c]{@{}c@{}}Depth-guided\\ pooling size\\ selection\end{tabular} &
   &
   &
  79,1***** &
  78,2**** &
   &
   &
   &
  Multi-scale &
  Late &
  \begin{tabular}[c]{@{}c@{}}Feature pyramid,\\ concatenation\end{tabular} &
  Encoder-decoder \\
2018 &
   &
  FCN &
  \begin{tabular}[c]{@{}c@{}}CityDriving\\ (self-supervised\\ depth estimation)\end{tabular} &
  512x512 &
   &
  Depth as Pretraining &
   &
   &
   &
  60,50 &
   &
   &
   &
   &
   &
   &
   &
  Encoder-decoder \\
2018 &
  ScaleNet &
  ResNet101 &
  ImageNet &
   &
   &
  Depth as Operation &
  \begin{tabular}[c]{@{}c@{}}Depth-guided\\ pooling size\\ selection\end{tabular} &
   &
   &
  75,10 &
   &
   &
   &
   &
  Multi-scale &
  Late &
  Feature pyramid &
  Encoder-decoder \\
2019 &
  LDFNet &
  ERFNet &
   &
  512x1024 &
  Titan X Maxwell &
  Depth as Input &
   &
   &
   &
  68,48 &
  71,30 &
  18,40 &
  2,31 &
   &
  Multi-branch &
  Middle &
  Addition &
  Multi-branch dncoder \\
2019 &
  RPPNet/ITNet &
  ResNet &
  ImageNet &
   &
  \begin{tabular}[c]{@{}c@{}}NVIDIAGeForce\\ GTX1080\end{tabular} &
  Depth as Input &
  \begin{tabular}[c]{@{}c@{}}Concatenated\\ stereo pair\end{tabular} &
  \begin{tabular}[c]{@{}c@{}}Feature/label\\ refinement\end{tabular} &
   &
   &
  67,08 &
  3,03 &
  140,00 &
   &
  Multi-term (RNN) &
  Late &
  RNN, concatenation &
  Encoder-LSTM-decoder \\
2019 &
   &
  VGG &
  ImageNet &
  512x1024 &
   &
  Depth as Input &
   &
   &
   &
  63,13 &
   &
   &
   &
   &
  \begin{tabular}[c]{@{}c@{}}Multi-branch,\\ multi-level\end{tabular} &
  Middle &
  Addition &
  Multi-branch encoder \\
2020 &
  RFNet &
  ResNet18 &
  ImageNet &
  2048x1024 &
  RTX 2080Ti &
  Depth as Input &
   &
   &
  Efficient model design &
  72,50 &
   &
  22,20 &
  23,69 &
   &
  \begin{tabular}[c]{@{}c@{}}Multi-branch,\\ multi-level\end{tabular} &
  Middle &
  Attention, addition &
  Multi-branch encoder \\
2020 &
   &
  ResNet50 &
   &
   &
   &
  Depth as Input &
  HHA &
   &
   &
  81,7 &
  82,8**** &
   &
  63,40 &
  204,9*** &
  \begin{tabular}[c]{@{}c@{}}Multi-branch,\\ multi-level\end{tabular} &
  Middle &
  Hand-crafted/gating &
  Multi-branch encoder \\
2020 &
  MTUNET &
  UNet &
   &
   &
   &
  Depth as Prediction &
   &
   &
   &
  73,19cls &
   &
   &
   &
   &
  \begin{tabular}[c]{@{}c@{}}Multi-branch,\\ multi-level\end{tabular} &
  Middle &
  Concatenation &
  Multi-branch decoder \\
2021 &
  QuadroNet &
  RetinaNet &
   &
  1920x1200** &
  TeslaV100 &
  Depth as Prediction &
   &
   &
  Efficient model design &
   &
  80,73** &
  49,75* &
   &
   &
  Multi-branch &
  Late &
  Hand-crafted &
  Multi-branch decoder \\
2021 &
  SGNet &
  ResNet101 &
  ImageNet &
  1024x2048 &
  1080 Ti &
  Depth as Operation &
   &
   &
   &
  80,6**** &
  81,2**** &
  28*** &
  58.3 &
   &
   &
   &
   &
  Encoder-decoder \\
2021 &
  GRB &
  ERFNet &
   &
  1024x512 &
  1080 Ti &
  Depth as Input &
  HHA &
   &
   &
  72,00 &
  69,70 &
  22,50 &
   &
   &
  \begin{tabular}[c]{@{}c@{}}Multi-branch,\\ multi-level\end{tabular} &
  Middle &
  Hand-crafted/gating &
  Multi-branch encoder \\
2021 &
  MENet &
  VGG &
  ImageNet &
  1024x2048 &
  \begin{tabular}[c]{@{}c@{}}GeForceGTX\\ 1080Ti\end{tabular} &
  Depth as Prediction &
   &
   &
   &
  61,50 &
   &
  0,50 &
   &
   &
  \begin{tabular}[c]{@{}c@{}}Multi-branch,\\ multi-level,\\ multi-scale\end{tabular} &
  Middle &
  Hand-crafted &
  \begin{tabular}[c]{@{}c@{}}Multi-branch encoder/\\ Multi-branch decoder\end{tabular} \\
2021 &
   &
  ResNet101 &
  ImageNet &
  1024x512 &
   &
  \begin{tabular}[c]{@{}c@{}}Depth as Pretraining,\\ Depth as Prediction,\\ Depth as Augmentation\end{tabular} &
   &
  \begin{tabular}[c]{@{}c@{}}Consecutive frame\\ processing\\ for pose and\\ motion estimation\end{tabular} &
   &
  71,16 &
   &
   &
   &
   &
  Multi-branch &
  Late &
  Attention &
  Multi-branch decoder \\
2021 &
  OSDNet &
  MobileNet V2 &
  ImageNet &
  1024x512 &
  \begin{tabular}[c]{@{}c@{}}Nvidia GTX\\ 1080Ti\end{tabular} &
  Depth as Prediction &
   &
   &
  \begin{tabular}[c]{@{}c@{}}Lightweight backbone,\\ Efficient model design\end{tabular} &
  54,67 &
   &
   &
   &
   &
  Multi-level &
  Middle &
  Addition &
  Encoder-decoder \\
2021 &
  S$^{3}$DMTNet &
  SegNet &
   &
  128x256 &
   &
  Depth as Prediction &
   &
   &
   &
  55,63cls &
   &
   &
   &
   &
  Multi-branch &
  Middle &
  Concatenation &
  \begin{tabular}[c]{@{}c@{}}Multi-network\\ (encoder-decoder)\end{tabular} \\ \hline
\end{tabular}%
}
\end{table*}

\subsection{Depth as Input}
\label{subsection:rgb-d_semantic_segmentation_depth_as_input}
Depth as input treats depth as an additional input to the network, which can either be fed as an additional channel to the network, or processed as a separate input by a dedicated branch. A common approach in pioneering works was to treat depth information as an additional channel to RGB images, and then perform the semantic segmentation of this concatenated representation through Encoder-Decoder networks \cite{b73} \cite{b74}.
\par However, throughout the years, and until nowadays, multi-branch encoders have been widely adopted. In \cite{b107}, a dual-branch encoder is proposed to separately process RGB data and its corresponding depth map, which are then processed by a shared decoder to get the final segmentation. \cite{b108} proposes, besides the RGB and Depth streams, a third Interaction stream, which, by using residual fusion blocks with gating mechanisms, performs feature fusion in order to model interdependencies and aggregate complementary features from both modalities.
\par Other works use the stereo pair as input to the network \cite{b109} \cite{b110}. In \cite{b109}, the concatenated stereo pair is fed to a dual-branch architecture for the processing of semantics and disparity. In \cite{b111}, sequences of concatenated stereo pairs are embedded into a LSTM-based model. \cite{b110} applies a multi-view strategy, in which the stereo pair is independently processed through a Siamese encoder.
\par Besides Depth and RGB, other image-like data can be considered as additional modality for feature extraction. In \cite{b112}, depth, optical flow and RGB are considered as inputs, and a mid-level multi-modal fusion strategy is implemented. \cite{b82} uses intensity and depth maps as inputs to a two-branch encoder, and generates free-path and road objects segmentation through separate decoder heads. In \cite{b113} RGB images and the concatenated Depth and Luminance are fed to separate encoders.
\par Simple input fusion, although being faster, has been proved to yield limited performance. Multi-branch approaches, on the other hand, while performing better and being widely used in the literature, can get expensive in terms of computational costs and memory requirements, due to the use of separate encoders for each modality.

\subsection{Depth as Operation}
\label{subsection:rgb-d_semantic_segmentation_depth_as_operation}
Depth as operation aims to embed depth into typical DCNN operations, such as convolution and pooling, giving place to depth-aware operations – that is, operations whose behavior is influenced by depth information. One of the pioneering works in this direction was \cite{b114}, in which depth-aware convolution and average pooling are introduced in order to embed geometry information into the network, without the need of additional parameters. Their main intuition is that neighboring pixels with the same class also share similar depths. Hence, by introducing a depth similarity term into the convolution operation, the authors enforce pixels with similar depths to the center pixel to have more contribution to the output than others. In \cite{b115} an auxiliar depth prediction network is used to guide, pixel-wise, the choice of the dilation rate for subsequent atrous convolutions, allowing the model to cope with the ambiguity in appearance between identical distant and close-by objects. Chen et al. \cite{b18} propose a Spatial Information guided Convolution (S-Conv) in which geometric information is used to both adapt the receptive field and incorporate geometry information into the feature learning by generating spatially-adaptive convolutional weights. Kong and Fowlkes \cite{b116} propose a depth-aware gating mechanism that, based on the object scale, adaptively selects the pooling sizes from a pyramid of pooling operations. Therefore, small details are preserved for distant objects while larger receptive fields are used for those nearby. An interesting finding from this work is that the use of depth predicted by the model yields better results than the use of raw depth data acquired from sensors. This can be explained by the fact that raw depth data carries noise and, sometimes, inaccurate readings, factors that can harm performance.
\par In summary, depth as operation has as its main advantage the possibility of directly embedding depth into the network for further enhancement of feature extraction capabilities. Compared to methods that apply depth as input, it incurs a smaller computational burden, since there is no need for duplicated parameters and computation, as it is the case when using modality-specific branches.
\par Although being an interesting concept, using depth as operation still has limited application. One of the main reasons to that is the non-fixed, and possibly deformed, receptive field generated when depth is used as a modifier of its sampling locations. This leads to sub-optimal suitability for current accelerators, such as GPUs, when compared to plain convolutions.

\subsection{Depth as Prediction}
\label{subsection:rgb-d_semantic_segmentation_depth_as_prediction}
Generally speaking, depth-as-prediction is frequently used in multi-task learning frameworks. One of the first works to study the structural correlation among visual tasks, in a transfer-learning scenario, was \cite{b117}. In such scenarios, from a single input, different visual perception tasks are solved at once. This allows the model to implicitly extract the complementary geometric information with the auxiliary depth prediction task, since the models generally have a shared encoder that is optimized considering the different tasks. 
\par Some of the main advantages of multi-task learning are improved computational efficiency, regularization and scalability \cite{b118}. Additionally, multi-task learning methods can achieve better generalization than task-specific models, when tested on datasets different from the one used for training, but from a related domain \cite{b119}. Wang et al. \cite{b46} enforces this idea by stating that the correlation between tasks in more invariant across domains than the individual modalities.  There are also evidences that leveraging multi-task learning allows to attain good performance for smaller models \cite{b7}.
\par
\begin{figure*}[htbp]
\centerline{\includegraphics[width=\textwidth]{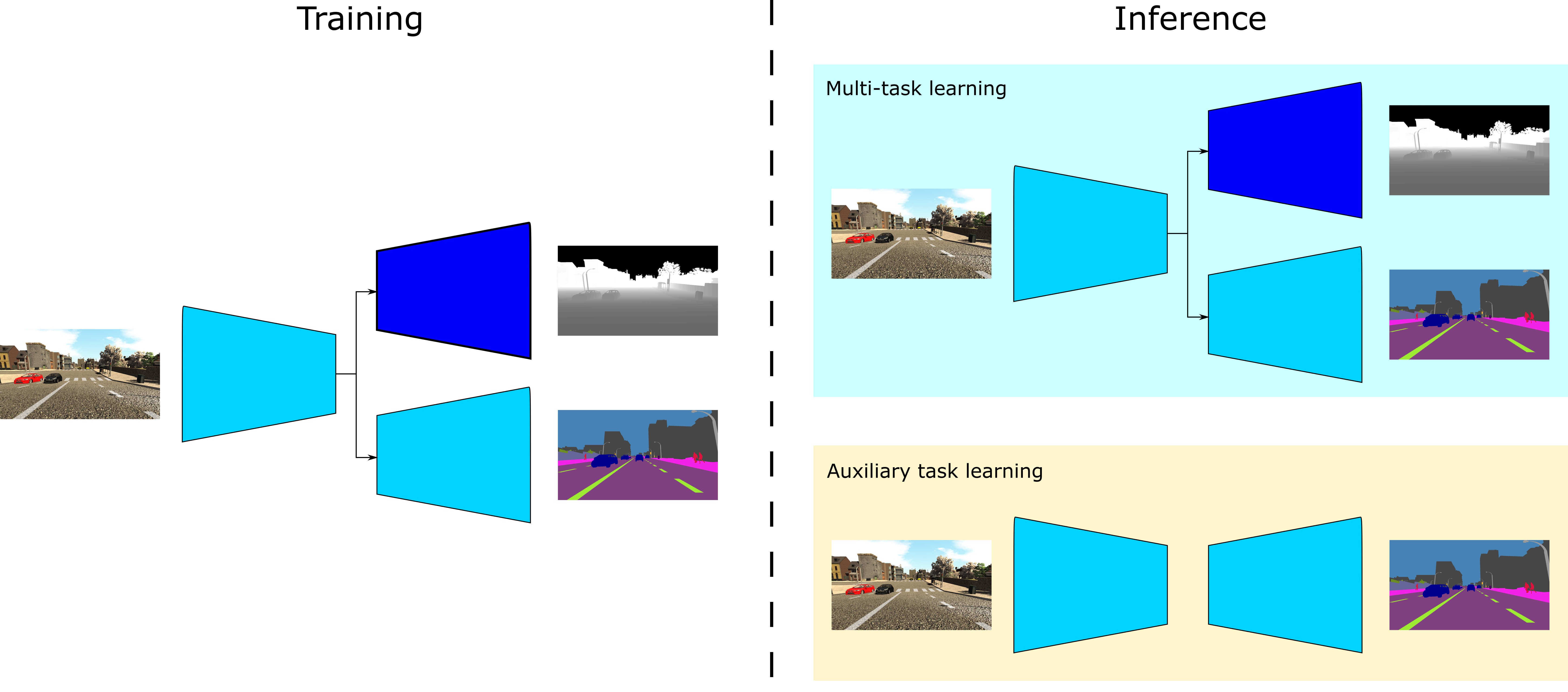}}
\caption{Auxiliary (a) and multi-task (b) learning setups.}
\label{fig:aux_task_vs_multi_task}
\end{figure*}

 \cite{b118} goes further and subdivides multi-task learning into auxiliary-task learning and multi-task learning – Fig~\ref{fig:aux_task_vs_multi_task}. In the auxiliary-task learning setup, the auxiliary task learned during training serves the only purpose of enhancing the main task’s performance, and is nonoperational during inference. In a multi-task learning setup, the auxiliary task is operational during both optimization and inference.

\par It is also possible to make a distinction according to the level of parameter sharing in a multi-task framework. Particularly, Jha et al. \cite{b120} distinguishes between soft and hard-sharing. In the first case, models have a separate network for each task under consideration, resulting in a disjoint set of parameters. In the case of hard-sharing - the most adopted in the literature - the tasks share the feature encoder, and task-specific decoders are added on top of it.

\par A common approach to multi-task learning for autonomous navigation involves bringing depth estimation together with semantic segmentation. Hoyer et al. \cite{b121} support that depth estimation and semantic segmentation are correlated in terms of sample difficulty.  According to Wang et al. \cite{b46} domain-robust correlations between semantics and depth – e.g., sky is always faraway, while roads and sideways are always flat – have the potential to largely improve the target semantic segmentation performance in the presence of a domain shift. In the same line, Cardace et al. \cite{b45} argue that depth reasoning deals with appearance, shape, relative sizes and spatial relationships of the stuff and things in the 3D images, and tend to extract accurate information for regions characterized by repeatable and simple geometries, such as roads and buildings, which feature strong spatial and geometric priors - e.g., the road is typically a plane in the bottom part of the image.
\par
\begin{figure}[htbp]
\centerline{\includegraphics[width=\columnwidth]{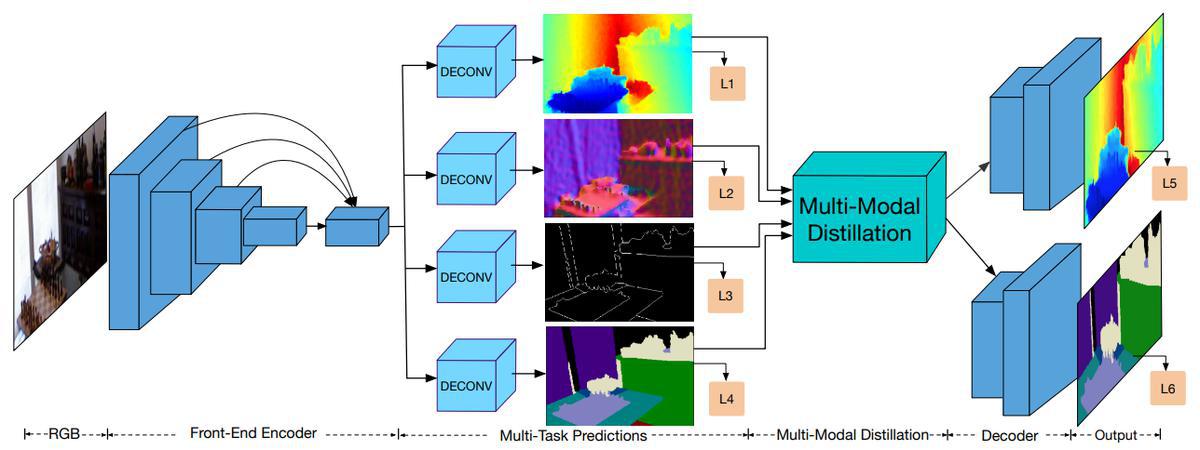}}
\caption{PadNet architecture. \cite{b69}.}
\label{fig:pad_net}
\end{figure}

 Goel et al. \cite{b122} propose a network architecture that jointly produces outputs for four tasks through four task-specific decoders: 2D detection, instance segmentation, semantic segmentation and depth estimation. In PAD-Net \cite{b69}, intermediate auxiliary tasks – monocular depth estimation, surface normal prediction, semantic segmentation and contour prediction – are used for both supervising the training and providing more meaningful representations to the downstream tasks – semantic segmentation and depth estimation (Fig.~\ref{fig:pad_net}). Aladem et al. \cite{b119}, on the other hand, propose to perform multi-task learning by predicting depth as an additional channel to the segmentation output, which allows an entirely shared network architecture, and a loss function composed solely by regression terms.

\par Chennupati et al. \cite{b123} process consecutive frames through a Siamese-like encoder, and outputs segmentation, depth and motion maps through task-specific decoders. In \cite{b118}, a multi-task learning framework composed by depth estimation (auxiliary task) and semantic segmentation (main task) is proposed. S3DMT-Net \cite{b120} is a particular case of depth as prediction, in which not only decoders, but also encoders for each task are independent. That is, it proposes a framework composed of task-specific networks connected through multi-level intermediate feature fusion through concatenation.
\par Other works leverage semantic segmentation as a helper task to refine disparity map generation \cite{b110}, stereo matching \cite{b124}, and monocular depth estimation and completion \cite{b125} \cite{b7} \cite{b126}.
\par One of the main advantages of leveraging depth as prediction is that it does not require depth sensors during inference, since depth is not provided as input to the network, what allows the use of cheaper camera sensors – nonetheless, in most cases, training requires full supervision, making pre-processing for ground-truth depth map generation a necessary step.
\par However, when compared to RGB-only and operation-level RGB-D segmentation, multi-task learning can lead to very complex and computationally heavy setups, with many hand-crafted feature extraction and fusion strategies. This reduces network flexibility, due to rigid architectures and many priors introduced in the model \cite{b109}.

\subsection{Depth as Augmentation}
\label{subsection:rgb-d_semantic_segmentation_depth_as_augmentation}
\begin{figure*}[htbp]
\centerline{\includegraphics[width=\textwidth]{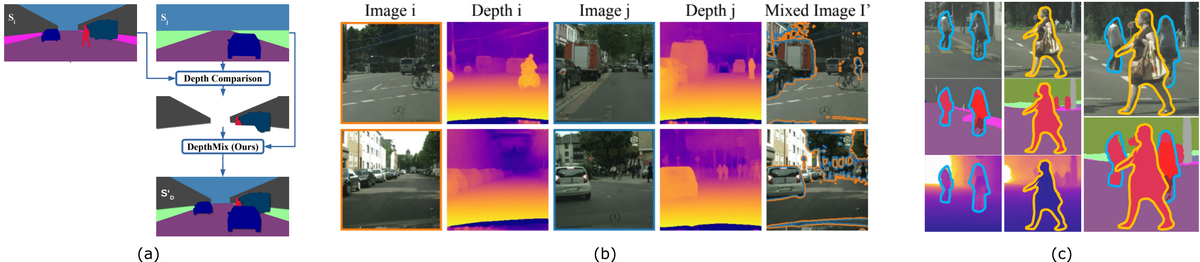}}
\caption{Depth used as auxiliary source of information in pseudo-label generation for data augmentation.}
\label{fig:depth_as_augmentation}
\end{figure*}

Depth as augmentation aims to leverage depth to generate new pseudo samples and labels out of existing labeled data, in order to augment the labeled data available for supervision. In this line of research, \cite{b121} and \cite{b45} explore depth to generate geometry-consistent new samples and labels through blending mechanisms – Fig~\ref{fig:depth_as_augmentation}.

\subsection{Depth as pre-training}
\label{subsection:rgb-d_semantic_segmentation_depth_as_pre-training}
\begin{figure}[htbp]
\centerline{\includegraphics[width=\columnwidth]{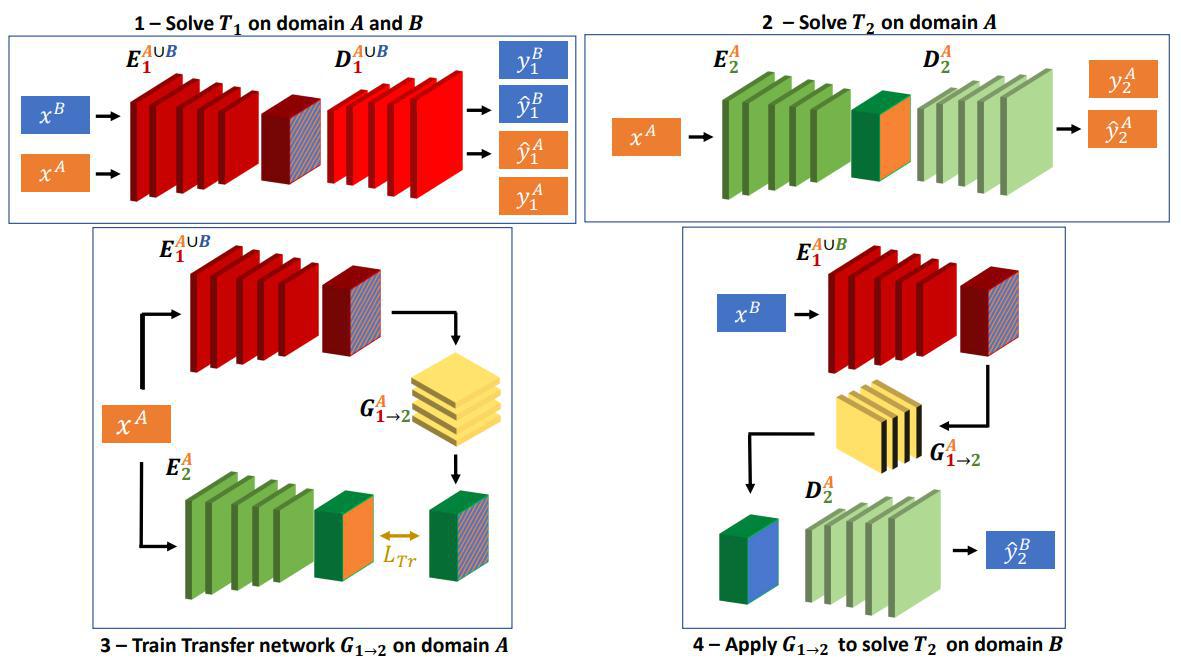}}
\caption{In depth as pre-training setups, depth estimation is usually leveraged as a proxy task for semantic segmentation (target task) . An example of use of depth as pre-training occurs in domain adaptation and cross-task/cross-domain learning scenarios \cite{b42}. In this example, T$_{1}$ corresponds to depth estimation, and T$_{2}$ corresponds to semantic segmentation.}
\label{fig:depth_as_pretraining}
\end{figure}

Depth as pre-training involves a pre-training step on depth estimation, followed by a fine-tuning on semantic segmentation \cite{b127}. That is, depth estimation can be viewed as a proxy task to improve the performance of the semantic segmentation, the target task. Some approaches dealing with domain adaptation and cross-domain learning propose a similar method \cite{b45} \cite{b42} \cite{b44}; however, in these works, separate networks are used for depth estimation and semantic segmentation, leaving to a third network the task of learning the relationship between both representations extracted – Fig.~\ref{fig:depth_as_pretraining}.
\par One of the main advantages of this approach is that it can be performed without additional side structures, since the same networks are used during pre-training and fine-tuning. In addition to that, the pre-training task can be used so that it belongs to the same domain as the target task, in order to reduce the effect of domain shift encountered in similar ImageNet pre-training approaches – even with fewer amounts of data, the pre-training on a similar domain could yield competitive results. 

\par Although interesting from the point of view of domain adaptation and transfer-learning, this approach still provides limited results in terms of accuracy, leaving a lot of room for improvements.

\subsection{Depth as Regularization}
\label{subsection:rgb-d_semantic_segmentation_depth_as_regularization}
Depth as a regularization uses depth information, either in the form of depth or disparity, as a component of the loss functions, such as a regularization term or a weighting factor.
\par Papadopoulos et al. \cite{b128} explore depth as an additional term that penalizes for inconsistent boundaries between semantic predictions and their corresponding depth maps. The intuition behind this idea is that semantic objects tend to stand out in depth maps, leading to co-occurrence of image gradients in the two tasks. Loukkal et al. \cite{b129} propose to use disparity as a weighting factor of the per-pixel cross-entropy loss, in order to give more importance to closer objects during optimization. It shows improvements over its no-weighted and frequency-weighted counterparts.
\par The main advantages of this technique are: the possibility of explicitly take depth into consideration to focus on certain ranges, according to the requirements of the task at hand; and the increase in accuracy without any additional computation burden, since the network architecture does not need to be redesigned with additional modules. Its main disadvantage is the explicit addition of inductive bias in the form of prior knowledge – in \cite{b129}, for instance, more weight is given to closer elements; hence, flexibility is reduced.

\section{Temporal-aware Semantic Segmentation}
\label{section:temporal-aware_semantic_segmentation}
Video sequences are the natural source of information from the world when dealing with visual perception for autonomous navigation systems, such as self-driving vehicles. The main advantage of video processing is the possibility of exploiting temporal correlations among frames in order to leverage redundancy, coherence and motion as additional cues to improve accuracy and reduce computations. In the case of automated driving, for instance, the difference between a static and a moving vehicle is not easily perceivable in single frames but easily captured in a video sequence \cite{b130}.
\par Throughout the years, several works proposed different alternatives in order to better explore video sequences in Deep Segmentation. The simplest approach is to independently process every frame of a video sequence through single-frame architectures. Although very straightforward, this approach totally ignores the inherent temporal information in videos. To properly leverage temporal information in videos, several temporal-aware video segmentation architectures were proposed in recent years. 
\par Generally speaking, DCNN models can explore both short-term and long-term temporal information. Short-term methods consider short frame intervals. The simplest way to doing so is by applying channel-wise frame concatenation. In this technique, the frames of a video sequence are concatenated before being fed to the network. The model is then expected to implicitly learn temporal and spatial features from this grouped representation. 3D convolutions \cite{b131} and 3D FCNs \cite{b132} are some examples of works in this line.  Another common line of work applies optical flow as additional inputs, so to model frame-by-frame temporal correlations. In \cite{b133}, \cite{b134} and \cite{b135}, for instance, the authors propose dual-branch architectures composed by appearance and motion branches where spatial and temporal reasoning are performed, respectively. One drawback of such technique is that it is prone to errors derived from the process of optical flow estimation – caused by fast motions, occlusions and disocclusions. 
\par Long-term relationships involve a more complete aggregation of features over time. In this respect, a well-known and widely-applied method for sequence modeling – not only in computer vision tasks – refers to recurrent neural networks, in which LSTMs \cite{b136} and GRUs \cite{b137} stand out in the literature. One of the main disadvantages of such approaches is that they work with flattened representations of the input data, which prevents spatial reasoning. In computer vision tasks, where spatial relationships play main role, this is highly undesirable. Consequently, ConvLSTMs \cite{b138} and ConvGRUs \cite{b139} \cite{b140} were proposed as alternatives to their flattened versions so that to allow spatial reasoning. Zhou et al. \cite{b96} use the ConvLSTM module to capture relevant information between video frames, and show improvements in accuracy especially for dynamic and small objects, such as truck, pedestrian and pole. Wang et al. \cite{b67} model temporal coherence using a ConvLSTM-based model. Lup and Nedevschi \cite{b141} explore dense optical flow and Spatio-Temporal Transformer Gated Recurrent Unit (STGRU) \cite{b142} for prediction refinement in video semantic segmentation. Siam et al. \cite{b130} adopt a Convolutional GRU unit as intermediate stage in a FCN model, in order to model temporal information. Nevertheless, one of the main disadvantages of RNN-based approaches is their sequential nature, which hinders leveraging parallelism to its full potential, such as in GPUs.
\par In spite of the specific pros and cons of each technique previously discussed, there are common challenges when developing temporal-aware Deep Semantic Segmentation methods. Firstly, since there is no natural (not synthetic) dataset where consecutive annotations are provided, due to the costs inherent to the labeling process, fully supervised video semantic segmentation is not possible. Besides that, balancing accuracy and inference time in video semantic segmentation methods is a challenge even more critical than in single-frame approaches. 
\par The current literature has tried to deal with these problems by adopting different strategies for feature and label reuse and aggregation. Herein, we will divide such approaches according to their main goal: reduction of computational burden, increase of label efficiency, or increase of accuracy.

\begin{table*}[]
\caption{Summary of the temporal-aware works reviewed.}
\label{tab:temporal_aware}
\resizebox{\textwidth}{!}{%
\begin{tabular}{ccccccccccccccccccc}
\hline
\multirow{2}{*}{Year} &
  \multirow{2}{*}{Model Name} &
  \multirow{2}{*}{Backbone} &
  \multirow{2}{*}{Pre-training} &
  \multirow{2}{*}{Image Resolution} &
  \multirow{2}{*}{Hardware} &
  \multicolumn{2}{c}{Temporal} &
  \multicolumn{2}{c}{Real-time strategy} &
  \multicolumn{2}{c}{Cityscapes mIoU} &
  \multirow{2}{*}{FPS} &
  \multirow{2}{*}{Params} &
  \multirow{2}{*}{GFLOPs} &
  \multicolumn{3}{c}{Fusion} &
  \multirow{2}{*}{Architecture} \\ \cline{7-12} \cline{16-18}
 &
   &
   &
   &
   &
   &
   &
  Obs &
   &
  Obs &
  Val &
  Test &
   &
   &
   &
  Type &
  Level &
  Mechanism &
   \\ \hline
2018 &
   &
  ResNet-101 &
  ImageNet &
   &
   &
  Feature/label propagation &
  \begin{tabular}[c]{@{}c@{}}Adaptive keyframe \\ selection\end{tabular} &
  Asymmetric frame processing &
  Feature/label reuse &
  75,89 &
   &
  8,40 &
   &
   &
  Multi-term &
  Late &
  \begin{tabular}[c]{@{}c@{}}Convolution,\\ multiplication\end{tabular} &
  Encoder-Decoder \\
2018 &
  DVSNet &
  DeepLab-Fast &
   &
   &
  NVIDIA GTX 1080 Ti &
  \begin{tabular}[c]{@{}c@{}}Feature/label propagation \\ and refinement\end{tabular} &
   &
  Asymmetric frame processing &
   &
  63,20 &
   &
  30,40 &
   &
   &
  Multi-term &
  Late &
  Hand-crafted &
  Multi-network \\
2019 &
   &
  WideResNet-38 &
  Mapillary &
  800x800 &
  V100 &
  Feature/label propagation &
  \begin{tabular}[c]{@{}c@{}}Joint image-label\\ propagation\\ (pseudo-label generation)\end{tabular} &
  Input cropping &
   &
   &
  83,5****$^{,}$***** &
   &
   &
   &
   &
   &
   &
  Encoder-Decoder \\
2019 &
  Accel18 &
  ResNet-18 &
  ImageNet &
  1024x2048 &
  Tesla K80 &
  \begin{tabular}[c]{@{}c@{}}Feature/label propagation \\ and refinement\end{tabular} &
   &
  Asymmetric frame processing &
   &
  72,1** &
   &
  2,20 &
   &
   &
  Multi-term &
  Late &
  Convolution &
  Encoder-Decoder \\
2020 &
  EVS 01 &
  ICNet &
  Cityscapes &
  2048x1024 &
  Titan Xp &
  \begin{tabular}[c]{@{}c@{}}Feature/label propagation\\ and refinement\end{tabular} &
  Optical flow &
  \begin{tabular}[c]{@{}c@{}}Efficient model design,\\ Assymetric frame processing\end{tabular} &
  \begin{tabular}[c]{@{}c@{}}Hybrid CPU-GPU processing,\\ Feature/label reuse\end{tabular} &
   &
  67,6** &
  37,00 &
   &
   &
  Multi-term &
  Late &
  Hand-crafted &
  Encoder-Decoder \\
2020 &
   &
  ERFNet &
  ImageNet &
  1024x512 &
   &
  Feature/label refinement &
  Optical flow and GRU &
  Lightweight backbone &
   &
  72,48 &
   &
   &
   &
   &
  Multi-term &
  Late &
  RNN (STGRU) &
  Encoder-Decoder + RNN(STGRU) \\
2020 &
  TD4PSP18 &
  BiseNet &
  ImageNet &
   &
  Titan Xp &
  \begin{tabular}[c]{@{}c@{}}Distributed/parallel\\ frame processing\end{tabular} &
   &
  \begin{tabular}[c]{@{}c@{}}Lightweight backbone,\\ Efficient model design\end{tabular} &
  Subsampled attention maps &
  75,00 &
  74,9(multi-term average) &
  47,60 &
   &
   &
  Multi-term &
  Late &
  Attention &
  \begin{tabular}[c]{@{}c@{}}Encoder-Decoder\\ (Knowledge Distillation setup)\end{tabular} \\
2020 &
  BDNet18 &
  DeepLabV2 &
   &
  512x1024 &
  Tesla K80 &
  \begin{tabular}[c]{@{}c@{}}Feature/label propagation\\ and refinement\end{tabular} &
   &
  Asymmetric frame processing &
   &
  65,40 &
   &
  7,14 &
   &
   &
  Multi-term &
  Middle &
  Warping/distillation &
  Encoder-Decoder \\
2020 &
  DAVSS &
  DeepLabv3+ &
  ImegeNet &
  1024x2048 &
  NVIDIA GTX 1080Ti &
  \begin{tabular}[c]{@{}c@{}}Feature/frame propagation \\ and correction\end{tabular} &
   &
  Efficient model design &
   &
  77,33 &
   &
   &
   &
  \begin{tabular}[c]{@{}c@{}}826,378\\ (213 for \\ non-key \\ frames)\end{tabular} &
  \begin{tabular}[c]{@{}c@{}}Multi-term,\\ multi-level\end{tabular} &
  \begin{tabular}[c]{@{}c@{}}Middle,\\ late\end{tabular} &
  Hand-crafted &
  Encoder-Decoder \\
2020 &
  LERNet &
   &
   &
  512x1024 &
  GTX 1080 Ti &
  Feature/label refinement &
  Attention &
  Efficient model design &
  Factorized convolutions &
  69,50 &
  66,50 &
  100,00 &
  0,65 &
  12,70 &
  \begin{tabular}[c]{@{}c@{}}Multi-level,\\ multi-term\end{tabular} &
  Middle &
  Attention &
  Encoder-Decoder \\
2020 &
   &
  MobileNetV2 &
   &
   &
  NVIDIA GTX 1080 Ti &
  \begin{tabular}[c]{@{}c@{}}Feature/label propagation,\\ temporal consistency-based\\  loss/knowledge distillation\end{tabular} &
   &
  Knowledge distillation &
   &
  73,90 &
   &
  20,80 &
  3,20 &
   &
  Multi-term &
  Late &
  LSTM &
  \begin{tabular}[c]{@{}c@{}}Encoder-Decoder\\ (Knowledge Distillation setup)\end{tabular} \\
2021 &
   &
  DenseASPP &
   &
   &
   &
  Feature/label refinement &
  ConvLSTM &
  Efficient model design &
  Channel compression &
  77,60 &
   &
   &
   &
   &
  Multi-term &
  Middle &
  ConvLSTM &
  Encoder-ConvLSTM-Decoder \\
2021 &
  NoisyLSTM &
  ICNet &
   &
  512x1024 &
  Tesla V100 &
  Feature/label refinement &
  ConvLSTM &
  Lighweight backbone &
   &
  62,50 &
  61,60 &
   &
   &
   &
  \begin{tabular}[c]{@{}c@{}}Multi-scale,\\ multi-term\end{tabular} &
  Middle &
  ConvLSTM &
  Encoder-ConvLSTM-Decoder \\
2021 &
   &
  CSRNet &
  Cityscapes &
  1024x2048 &
  GTX 1080 Ti &
  \begin{tabular}[c]{@{}c@{}}Feature/label propagation\\ and refinement\end{tabular} &
   &
  Lightweight backbone &
   &
  73,71 &
   &
  25,84 &
   &
  241,72 &
  \begin{tabular}[c]{@{}c@{}}Multi-level,\\ multi-term\end{tabular} &
  Late &
  Hand-crafted &
  Encoder-Decoder \\
2021 &
  GSVNetSNR18 &
  SwiftNet-R18 &
  ImageNet &
  1536x768 &
  GTX 1080Ti &
  \begin{tabular}[c]{@{}c@{}}Feature/label propagation\\ and refinement\end{tabular} &
   &
  Lghtweight backbone &
   &
  71,8** &
   &
  142,00 &
  \begin{tabular}[c]{@{}c@{}}47,2\\ (1,6 for \\ non-key \\ frames)\end{tabular} &
  16,70 &
  \begin{tabular}[c]{@{}c@{}}Multi-branch,\\ multi-level\end{tabular} &
  \begin{tabular}[c]{@{}c@{}}Middle,\\ late\end{tabular} &
  Hand-crafted &
  Multi-branch/Multi-network \\
2021 &
  LMANet &
  ERFNet &
   &
  1024x512 &
  RTX 2080Ti &
  Memory &
   &
  Lightweight backbone &
   &
  73,72** &
   &
  86,21 &
   &
   &
  Multi-term &
  Middle &
  Attention &
  Encoder-Decoder \\
2021 &
  CSANet &
  ResNet-101 &
  ImageNet &
  480x480 &
  (8) TITAN RTX &
  Pseudo-labels &
  \begin{tabular}[c]{@{}c@{}}Pseudo-labels \\ from unlabeled\\ frames in mutual \\ learning technique\end{tabular} &
   &
   &
  80,76 &
  79,60 &
  2,00 &
   &
   &
  \begin{tabular}[c]{@{}c@{}}Multi-level,\\ multi-term\end{tabular} &
  Middle &
  Attention &
  Encoder-Decoder \\
2021 &
  TMANet &
  ResNet-50 &
   &
  768x768 &
  Tesla V100 &
  Memory &
   &
  Efficient model design &
  Shared backbone &
  80,30 &
   &
   &
   &
  754,00 &
  Multi-term &
  Middle &
  Attention &
  Encoder-Decoder \\
2021 &
  STTBise18 &
  ResNet-18(BiseNet18) &
   &
   &
   &
  Memory &
   &
  Efficient model design &
  \begin{tabular}[c]{@{}c@{}}Reduced attention map\\ (key and query selection)\end{tabular} &
  75,80 &
   &
  44,20 &
   &
   &
  Multi-term &
  Middle &
  \begin{tabular}[c]{@{}c@{}}Attention/\\ Transformer\end{tabular} &
  Encoder-Transformer-Decoder \\ \hline
\end{tabular}%
}
\end{table*}

\subsection{Reduction of Computational Burden}
\label{subsection:temporal-aware_semantic_segmentation-reduce_comput_burden}
\begin{figure}[htbp]
\centerline{\includegraphics[width=\columnwidth]{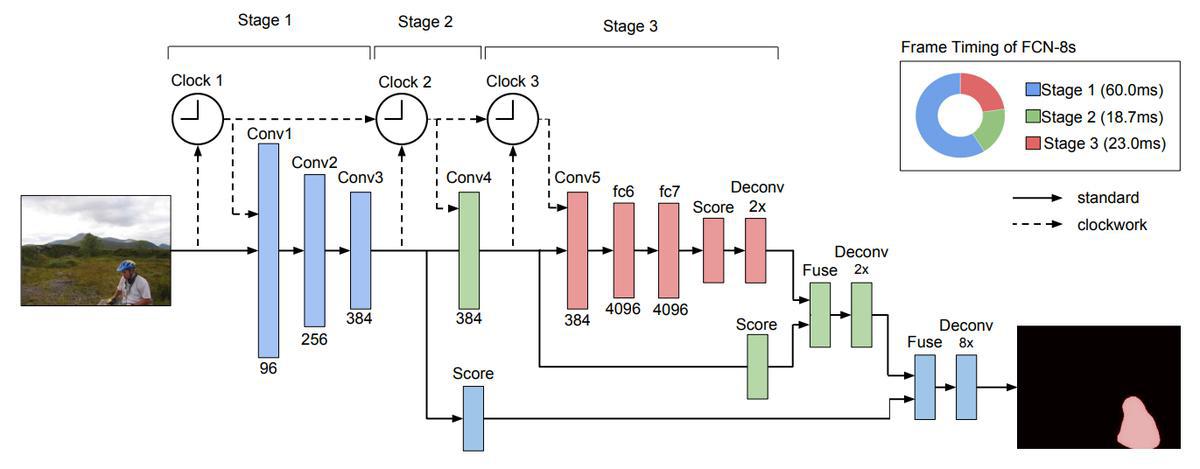}}
\caption{Clockwork architecture \cite{b143}.}
\label{fig:clockwork}
\end{figure}

Works aiming at the reduction of computational burden are generally characterized by adaptive processing schedules. Clockwork \cite{b143} proposes a clock-based network update mechanism from the intuition that semantic features change at a slower rate than features with rich spatial details; hence, deeper layers in the architecture do not need to be updated as frequently as shallower ones - Fig.~\ref{fig:clockwork}. The intuition that semantic information changes at a slower rate is what allows us to explore redundancy between nearby frames for reducing the computational burden in video semantic segmentation.

\par
\begin{figure}[htbp]
\centerline{\includegraphics[width=\columnwidth]{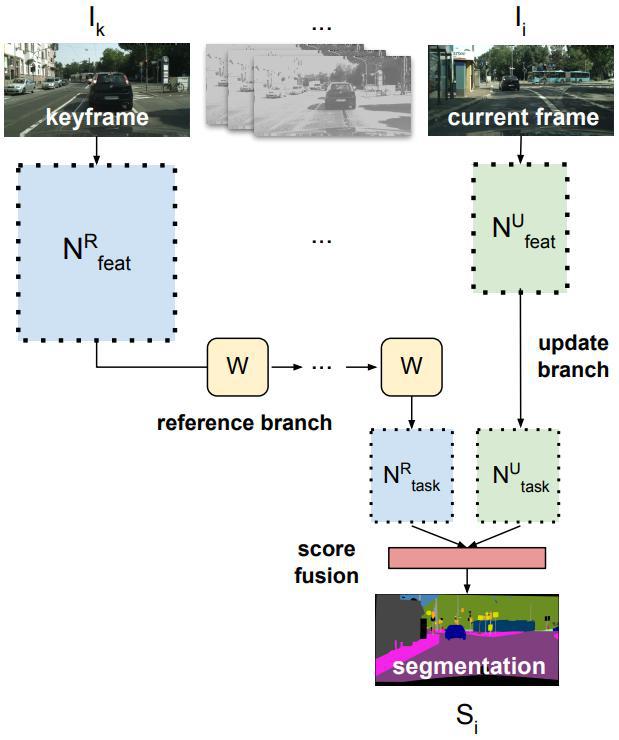}}
\caption{Accel architecture \cite{b38}.}
\label{fig:accel}
\end{figure}

 Asymmetric frame processing is also a very common approach in the literature \cite{b28} \cite{b30} \cite{b144} \cite{b145}. In this strategy, a given video sequence is divided into key and non-key frames. While keyframes are processed by the entire architecture, in order to extract both low-level and high-level features, non-key frames usually pass through lighter architectures, where usually only low-level features are extracted; high-level features are then calculated by propagating and aggregating high-level features extracted from previous keyframes - usually by means of the optical flow between adjacent frames. In \cite{b144}, computation cost for non-key frames is reduced four times when compared to key frames. Paul et al. \cite{b28} propose an optical flow-based feature and label warping. Li et al. \cite{b30} design an adaptive feature propagation scheme, where spatially-variant convolutions are used. The kernel weights are computed based on low-level feature maps, hence adapting to not only the locations but also the frame contents, obtaining great expressive power. Wu et al. \cite{b145} divide their backbone into low-level network and high-level network. Keyframes pass through the entire pipeline, generating both low and high-level features, while for non-key frames only low-level features are calculated, and high-level features are propagated using an attention mechanism. A similar approach is adopted by Wu et al. \cite{b145}. Lu and Deng \cite{b37} use motion vectors as additional inputs to the network in order to perform feature warping in non-key frames. In Accel \cite{b38}, key frames are processed by a reference branch, while non-key frames are processed by a smaller update branch (Fig.~\ref{fig:accel}). Finer granularity is studied in DVSNet \cite{b99}, where, instead of asymmetric frame processing, the authors propose asymmetric region processing. Regions with low confidence score are processed by a segmentation network in the segmentation path, while regions with high confidence are processed by a flow network in the spatial warping path, in order to warp features from previous frames, since they changed at a slower pace.

\par In the previous works, both adaptive and fixed schedules for key frame selection are possible. Adaptive selection is more adaptable to sudden changes in the frames, but incurs more computation. Li et al. \cite{b30} proposes an adaptive scheduler based on the deviation of current frame in relation to the keyframe. Since defining the deviation based on the segmentation masks would incur heavy computation, the authors use low-level features for calculating its value. Fixed strategies, on the other hand, can miss relevant changes in the video, but are computationally cheaper. In EVS \cite{b28} a fixed keyframe selection is applied. 
\par It’s worth noting that some authors argue that such an unbalanced processing scheme is harmful to model performance \cite{b56}. An alternative for that is presented by Li et al. \cite{b30}. Since the maximum latency is not decreased due to the inherent heavy computation at key frames, given a new key frame definition, the authors propose to store a temporary version of the high-level features of the key frame, computed by feature propagation, called “fast track”. Meanwhile, a background process is launched to properly compute the high-level features via the “slow track” – the high-level network –, without blocking the main procedure. When the computation is done, this version will replace the cached features.
\par Distributed/shared processing is also a strategy for reducing computations. TDNet \cite{b36} leverages the temporal continuity in videos, and distributes the computation of high-level features to several low-level feature computations by shallower networks. These are then recomposed into high-level features for the desired frame using an attention-based mechanism. This allows for reducing the amount of computation, since at each time step just a lightweight computation is required to extract sub-features (low-level features). TMANet \cite{b94} processes a memory sequence and the current frame by a shared backbone, subsequently aggregating features using an attention mechanism. Wu et al. \cite{b145} adopt a similar strategy.
\par Moreover, temporal consistency can be used to guide knowledge distillation mechanisms, so that to obtain smaller models \cite{b35}. Hu et al. \cite{b36} apply grouped knowledge distillation to explore the complementarity of intermediate representations extracted from sequential frames. Liu et al. \cite{b35}, besides exploring temporal consistency in the loss function of both teacher and student networks, defines pair-wise and multi-frame dependency distillation terms in order to propagate temporal awareness from the teacher network to the student network. Long and short-range dependencies are explored through a random sampling policy, in which previous and future frames are randomly sampled – instead of using frames right next to the labeled ones.

\subsection{Accuracy improvement}
\label{subsection:temporal-aware_semantic_segmentation-accuracy_improvement}
Some of the mechanisms used for accuracy improvement include: feature refinement, feature/label propagation and refinement, and memory networks.

\subsubsection{Feature refinement}
\label{subsubsection:temporal-aware_semantic_segmentation-accuracy_improvement-feature_refinement}
Feature refinement involves enhancing the features of a given frame by aggregating features extracted from previous frames. Chen et al. \cite{b29} propose the spatio-temporal continuity (STC) module to explore temporal region continuity in videos for feature refinement. 

\subsubsection{Feature/label propagation and refinement}
\label{subsubsection:temporal-aware_semantic_segmentation-accuracy_improvement-feature_label propagation_and_refinement}
Feature/label propagation and refinement refers to the process of propagating features and/or labels to neighboring frames, and then correcting discrepancies in the warped representations. Similar to what is done in asymmetric frame processing (section~\ref{subsection:temporal-aware_semantic_segmentation-reduce_comput_burden}), the propagation is frequently done by optical flow-based methods \cite{b146} – optical flow is usually calculated by a dedicated network \cite{b147}\cite{b148}\cite{b149}\cite{b150}. In addition to that, Wang et al. \cite{b94} further categorizes the literature covering non-optical-flow-based methods, in which feature propagation is usually achieved through some sort of attention mechanism \cite{b20} \cite{b36} \cite{b145}.
\par Whatever the propagation method is, the correction can be done based on either motion cues \cite{b28}, mismatching and uncertainty \cite{b106}, or on previous features \cite{b38}.
\par In \cite{b106}, after optical flow-based feature warping, feature rectification is performed by means of a distortion map-guided mechanism, which serves as a weighting factor for warped and updated features of a given frame. The supervision of distorted regions is made by an additional edge-semantics loss term in the objective function. Lu and Deng \cite{b37} apply feature warping based on motion vectors, and further enhance feature representations by adopting a multi-task learning framework where edge detection is used as auxiliary task. Knowledge distillation is used for feature and label refinement. Lee et al. \cite{b39} performs optical flow-based label propagation, while label refinement is performed through a series of steps involving shifted representations of the warped labels, as well as spatially and context-variant convolutions, calculated using an auxiliar network. Zhuang et al. \cite{b144} propose an optical flow-based feature and frame propagation, with distortion-aware feature correction. Hence, distorted regions are corrected, while the remaining regions preserve the warped features. Accel \cite{b38} adopts optical flow to propagate keyframe features, computed by a heavy model, to non-key frames, and refines these warped representations using the current frame’s features, extracted by an additional lightweight update network.
\par CSANet \cite{b20} uses attention to correlate and refine features from different frames in order to promote mutual learning. TDNet \cite{b36} aggregates multi-term sub-features using an attention-based mechanism. By discarding the optical flow, computations are saved, and error propagation due to incorrect flow estimation can be reduced. Output and feature-level refinement are performed using a knowledge distillation strategy. Wu et al. \cite{b145} design the Temporal Holistic Attention module to perform propagation of high-level features from key frames to the following non-key frames. Feature refinement is performed through a residual mechanism in which the predicted high-level features are adaptively fused with the original low-level features of the current non-key frame.

\subsubsection{Memory networks}
\label{subsubsection:temporal-aware_semantic_segmentation-accuracy_improvement-memory_networks}
In Memory Networks \cite{b101} \cite{b94} \cite{b21}, features computed from previous frames are stored using a memory or backup mechanism, and subsequently accessed to enhance the segmentation of the current frame - used as a query to retrieve highly-correlated information from the memory. 
\par Hu et al. \cite{b21} stores previous key and value features maps to improve key and value maps for the current frame so to enhance the final segmentation. Paul et al. \cite{b101} aggregate information from past frames into a memory module, which is accessed through an attention mechanism that only covers the spatial neighborhood of a given pixel in past frames. The main intuition behind this approach is that the content of the current frame at a given position is more likely to be found at a similar position in the previous few frames. Their goal is, besides reducing memory and computation consumption, make the model recover consistent behavior from past frames, while learning sudden changes from current features. TMANet \cite{b94} encodes keys and values of a memory sequence, and use the current frame as the query for recovering related information from previous frames, thus adaptively integrating long-term relations and enhancing the representation of the current frame. Li et al. \cite{b103} proposes a Sparse Temporal Transformer in order to correlate features from previous and current frame. Features extracted from previous frames are stored as keys, while features related to the current frame are treated as query. In order to reduce the known high computation cost of attention and, specifically, multi-head self-attention in transformers, the authors propose key and query selection strategies. For query selection, the most complex regions are selected based on the variety of classes encountered in a given region; key selection is made by enlarging the search region in key features as we move back in time – that is, frames far from the current frame have a larger search space. Frames are then correlated by a temporal transformer.

\subsection{Labeling Efficiency}
\label{subsection:temporal-aware_semantic_segmentation-labeling_efficiency}
Finally, methods for increasing labeling efficiency address the problem of label scarcity in semantic segmentation scenarios. Due to the costs involved in data labeling, the most frequently used video datasets for autonomous driving, such as Cityscapes and KITTI \cite{b151}, do not have labels for all their frames. Instead, they provide labels for particular frames in a video sequence – e.g., Cityscapes only provides labels for the 20th frame of its 30-frame video snippets. 
\par In this scenario, various works try to leverage unlabeled frames and previous knowledge to overcome label scarcity. In this respect, Lup and Nedevschi \cite{b141} show that, by leveraging optical flow and STGRUs for modeling temporal information, the proposed architecture was able to achieve the same segmentation quality as the static network employed, while using only 25
\par Pseudo-label generation \cite{b40} \cite{b41} is probably the most common approach to the problem in the recent literature. Zhu et al. \cite{b41} demonstrate that training segmentation models on datasets augmented by synthesized samples leads to significant improvements in accuracy. The authors exploit the model ability to predict future frames in order to also predict future labels, in a joint image-label propagation strategy. They also introduce a boundary label relaxation technique that allows for predicting multiple classes for a given boundary pixel during training. In their experiments, they show that by relaxing the boundary labels, training becomes more robust to annotation noise and accumulated propagation artifacts, allowing the model to benefit from longer-range training data propagation. Liu et al. \cite{b35} explores temporal consistency with respect to neighboring unlabeled frames in order to improve accuracy, while reducing the need for a large amount of dense labels.  Zhuang et al. \cite{b144} propose a dual deep supervision (DDS), where pseudo-labels of an intermediate frame from the interval under consideration is used to calculate an additional supervisory signal in order to improve model optimization in the absence of sufficient data. Varghese et al. \cite{b152} explore unlabeled video sequences in order to improve temporal consistency of single-frame methods. Pseudo-labels and optical flow-based label propagation are leveraged in a parallel stage of model training. Yuan et al. \cite{b20} leverage unlabeled frames during training by applying an inter-frame mutual learning strategy.

\section{Challenges and Future Directions}
\label{section:challenges}

\subsection{Data Scarcity}
\label{subsection:challenges-data_scarcity}
The majority of the works in Deep Semantic Segmentation literature still assumes the availability of labeled datasets for full supervision. However, dense annotation of images, such as semantic masks, is time-consuming and impracticable for large datasets (the “curse of data labeling”, as per \cite{b153}). This scenario is even worse when considering video datasets; besides containing a large number of frames per video snippet, they must have many of these snippets in order to guarantee enough representativeness for the task at hand. Therefore, a common approach is to provide sparse labels; that is, for a given sequence, a small number of frames has corresponding annotations, while the remaining is left unlabeled \cite{b62}. 
\par Given this context, several workarounds have been proposed in the literature – besides the ones already presented in section \ref{subsection:temporal-aware_semantic_segmentation-labeling_efficiency}. Noisy-LSTM \cite{b67} proposes a training strategy where, given a video sequence composed of context frames and the target frame, one of the context frames is replaced by noise, which can be either a noisy image or an image from a different domain. By spoiling the temporal coherence, the authors try to make the model robust to occasional and rare changes in frames, improving its ability of feature extraction and serving as a regularizer, without requiring extra annotations or computational costs.  Gurram et al. \cite{b154} use different datasets for each target output – depth estimation and semantic segmentation. 
\par Besides that, great part of the works tackling labelling efficiency either adopt alternative learning procedures or exploit synthetic datasets as a source of labeled data.

\subsection{Alternative learning procedures}
\label{subsection:challenges-alternative_learning}
Alternative learning procedures try to deal with the lack of labeled data by gradually reducing either the complexity or the need for ground truth data. The first method, weakly-supervised learning, tries to minimize the dependency on dense labels by adopting simpler ground-truth representations. Besides that, semi and self-supervised learning leverage the availability of large amounts of unlabeled images, such as in video datasets, by gradually reducing the need for ground truth.
\par More on weakly, semi and self-supervised learning will be presented in the following sections.

\subsubsection{Weakly-Supervised Learning}
\label{subsubsection:challenges-alternative_learning-weakly_supervised}
The idea behind weakly-supervised learning is to train a model using simpler ground-truth representations (weak labels), compared to the proper labels needed in common training setups; that is, considering a semantic segmentation task, where dense labels are necessary for fully-supervised learning, weakly-supervised setups would try to achieve similar results by using coarse masks, bounding boxes, or even image-level labels as ground truth. 
\par In WeClick \cite{b155}, a weakly-supervised training scheme is proposed, where click annotations are leveraged in order to reduce the need for dense labels. Temporal information is explored during training, in the form of optical flow for label warping. Saleh et al. \cite{b156} propose a weakly-supervised method which leverages video-level tags and motion extracted with optical flow.

\subsubsection{Semi-Supervised Learning}
\label{subsubsection:challenges-alternative_learning-semi_supervised}
Semi-Supervised Learning deals with the absence of labels by using unlabeled data additionally to the available ground-truth data. The main objective is to obtain a better performance than that get with full supervision on the limited set of labeled data. A common approach to Semi-Supervised Semantic Segmentation is to leverage unlabeled data in pseudo-label generation, so that to augment the available labeled data. \cite{b121} leverages depth estimates to generate geometry-consistent additional training data – new images and labels are generated by a blending method termed DepthMix.
\par Besides that, Cardace et al. \cite{b45} further fine-tune the trained network by its own predictions, which are used in a geometry-aware pseudo-label generation process. 

\subsubsection{Self-Supervised Learning}
\label{subsubsection:challenges-alternative_learning-self_supervised}
Self-supervised learning involves exploiting solely unlabeled data in order to extract useful representations from it. Works in this category usually apply some transformation on the unlabeled data, and use this modified version as the ground truth for contrastive learning \cite{b157} \cite{b158} \cite{b159} or proxy (also surrogate) task. However, this strategy is usually used only as a pre-training step, still being necessary to fine-tune the model for the specific downstream task.
\par Yang et al. \cite{b160} propose a subdivision of proxy task-based learning into context prediction and image generation. 
\par Context prediction exploit the spatial context of data, and can be used to predict relative location \cite{b161} or to solve jigsaw puzzles. For instance, Yang et al. \cite{b160} propose to solve a jigsaw puzzle comprising 25 patches, then transferring the learned features to a semantic segmentation task. Supervised fine-tuning is performed using only a 1/6 of the training images in Cityscapes, so that to mimic a scenario of data scarcity. The strategy achieves 5.8
\par Image generation methods usually operate by removing part of an image, and recovering the missing part afterwards. Image colorization \cite{b162} \cite{b163} and inpainting \cite{b164} \cite{b165} are common strategies.
\par In addition to the previous methods, image rotation prediction \cite{b166} and real-versus-synthetic image prediction \cite{b167} are also present in the literature.
\par For videos, common approaches to self-supervised learning are temporal order prediction and verification \cite{b168} \cite{b169} \cite{b170}, and future frame prediction \cite{b171} \cite{b111}.
\par Interestingly, the work from Zhang and Leonard \cite{b172}, instead of applying self-supervised learning as a pre-training step, adopts unlabeled data in a fine-tuning step (i.e. after supervised pre-training) to spot temporal inconsistencies in the prediction and create a training signal to correct deficiencies in long-term learning. Major improvements are observed in classes characterized by thin elements, such as fence, pole and traffic sign.

\subsection{Synthetic Data}
\label{subsection:challenges-synthetic_data}
\begin{figure*}[htbp]
\centerline{\includegraphics[width=\textwidth]{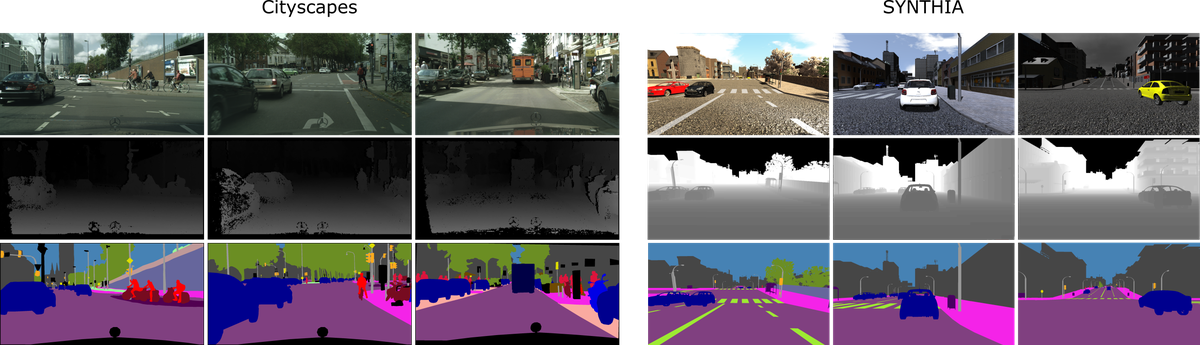}}
\caption{Natural versus synthetic image domains. We can observe that, although SYNTHIA dataset provides images with considerable scene complexity - buildings, vehicles, different textures and illuminations -, it is still far from reaching the variety of elements, textures and colors from natural scenes provided in Cityscapes. Besides that, depth measurements in Cityscapes are clearly influenced by noise and wrong measurements, factor that leads to performance degradation when adapting models pretrained in synthetic datasets to natural ones.}
\label{fig:natural_vs_synthetic}
\end{figure*}

The use of synthetic datasets \cite{b173}\cite{b174}\cite{b175} brings as main advantage the prompt availability of labeled data along with RGB – and even other sensing modalities, such as Depth maps. However, models trained using synthetic data may not be directly applicable to real scenarios, since they will present performance degradation caused by what is called domain gap (Fig.~\ref{fig:natural_vs_synthetic}) – it is worth mentioning that domain gap can also exist among real (not synthetic) datasets. Therefore, several works proposed domain adaptation setups in order to reduce this domain gap, and allow the use of synthetic data for performance boost in the lack of enough labeled data.

\subsection{Domain adaptation}
\label{subsection:challenges-domain_adaptation}
Deep Domain adaptation \cite{b176} has recently gained attention in the literature as an alternative to alleviate the need for large amounts of labeled data. Its main objective is to extract domain-invariant features which could then be transferred from a source to a target domain without considerable loss in performance. Generally, the preferred features for such methods are those from deep layers in the network, since they present semantic information in a higher abstraction level, which, presumably, are less domain-dependent than lower-level features.
\par Unsupervised domain adaptation (UDA) is a variant of domain adaptation, where no label data is available in the target domain. Because of this characteristic, many UDA methods rely on self-supervised learning approaches. Nonetheless, some authors support UDA as a variant of semi-supervised learning, where the labeled data come from a different (source) dataset \cite{b177}. 
\par The authors in \cite{b177} adopt the semi-supervised approach of self-training, with end-to-end co-evolving pseudo-labels generated by a momentum network – a slowly advancing copy of the original model (pre-trained on the synthetic dataset). Additionally, they enforce consistency of the semantic maps produced by the model across image perturbations – photometric jitter, flipping and multi-scale cropping. 
\par Other works further add the problem of task shift to domain shift/adaptation setups. Ramirez et al. \cite{b42} propose a framework to operate both across tasks and domains, so as to use the relationships between tasks to reduce the strong need for labeled data on new target domains. First, in a fully-supervised domain, such as synthetic data, the framework learns to transfer knowledge across tasks through an auxiliar network that learns the translation between target and source task embeddings; then, in a second step, this knowledge is used on a different domain where supervision is available only for one of two tasks - labels are available only for the auxiliary task, such as depth estimation. Chen et al. \cite{b43} propose a Generative Adversarial Learning strategy where, through min-max optimization, they explore depth and semantics in order to both generate realistic-style images from synthetic ones (input-level raw adaptation) and to discriminate between segmentations and depth estimations from either the generated or the real images in the target domain (output-level adaptation). Chavhan et al. \cite{b44} deal with domain discrepancy through generative models so that to generate a domain-invariant deep feature representation (embeddings) in order to reduce the domain-gap. The authors use min-max optimization to extract maximally task-descriptive and domain-invariant features.
\par Wang et al. \cite{b46} explore depth-semantics correlation in a setup where no labels are provided for both main and auxiliary tasks in the target domain. More specifically, the authors exploit task correlation in two ways: first, by modeling the task correlation in the source domain; second, by calculating the adaptation difficulty as the discrepancy between depth predictions in both domain-specific depth decoders, and then using it to refine segmentation pseudo-labels in the target domain.
\par Cardace et al. \cite{b45} propose a plug-in module to add depth information into existing UDA methods. Their main intuition is that semantic labels obtained from depth are smoother and more precise on objects with large and regular shapes (strong geometric priors), such as road, sidewalk, wall and building, while semantics from UDA performs better in regions where semantic from depth is imprecise, like thin objects partially overlapped or fine-grained structures in the background. 

\subsection{Data unbalance}
\label{subsection:challenges-data_unbalance}
\begin{figure}[htbp]
\centerline{\includegraphics[width=\columnwidth]{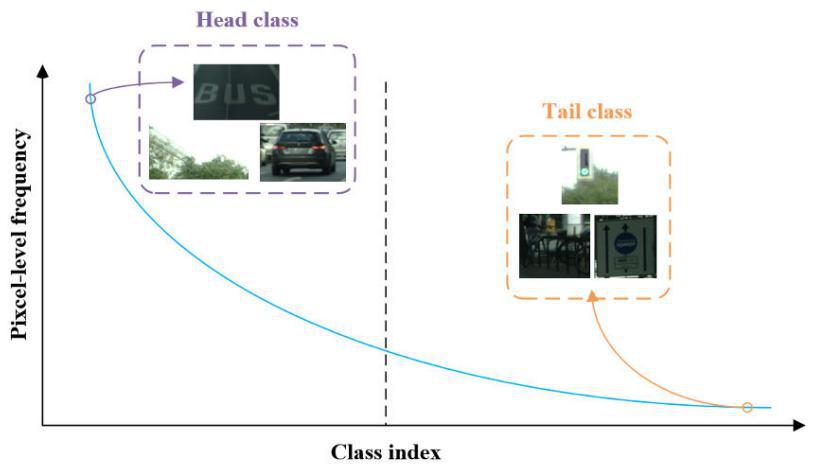}}
\caption{Long-tail problem of unbalanced class representation. \cite{b181}.}
\label{fig:long_tail_problem}
\end{figure}

Besides the problem of unlabeled data, data unbalance is also an important issue in visual perception in urban scenarios. Classes such as sky, road and buildings cover most of the visual field, thus representing great part of the information present in images. On the other hand, classes such as pole and sign – and, more importantly, pedestrian and cyclist – are underrepresented in terms of pixel frequency. This situation is termed long-tail problem, given the profile of the plot of pixel frequency per class (Fig.~\ref{fig:long_tail_problem}).

\par When we consider the processing of CNNs, as the network gets deeper, the receptive field also increases. Hence, at the prediction stage, the receptive field of a class such as traffic sign will cover a great amount of useless contextual information belonging to classes such as building or sky \cite{b41}. This can be harmful to the prediction precision, as well as represents a wasting of computation resources. 
\par Several works have already tackled data unbalance in the literature by means of frequency-weighted losses, data-driven adaptive operations and balanced sampling strategies \cite{b41}. However, these works assume supervision available for the tasks at hand. Hence, dealing with class unbalance in alternative learning strategies, such as self-supervised learning, is still an open challenge. 

\subsection{Robustness to Unseen Classes}
\label{subsection:challenges-unseen_classes}
The ability to adapt to unforeseen classes is an important aspect when considering the adaptation of production models to different contexts, such its use in different countries, where different types of vehicles, animals, traffic signs, ground signalization, and driving conditions are present. It would also be desirable to learn new classes under limited supervision, that is, under a limited number of samples. All these aspects make few-shot learning semantic segmentation an interesting future direction. In \cite{b178}, depth information is leveraged in order to improve performance in a few-shot semantic segmentation setup. Sun et al. \cite{b71} deals with unexpected obstacle detection. 

\subsection{Standardized evaluation criteria}
\label{subsection:challenges-unseen_classes}
Lack In standardization is also a major problem, since it makes it difficult to compare the vast literature on Deep Semantic Segmentation. 
\par First, in respect to the datasets used, many works compile their results only on a sub-set of classes available in the datasets used in the area (such as the Cityscapes). There are also works that evaluate their results on category groups, instead of classes. There is also a lack in standardization referring to the split used during testing; while some works provide metrics for the validation set, other use the test split. Finally, there are studies that evaluate their models on datasets privately acquired, hindering reproducibility.
\par Second, the training and testing setup varies a lot among the works. Besides that, some works leverage acceleration techniques, such as TensorRT, in order to boost performance. This prevents fair comparison of the methods – such as latency metrics. A possible workaround offered by some authors is the compilation of metrics considering normalized measures of hardware capacity \cite{b106}.
\par Finally, few works present all the metrics used for comparison in this review: mIoU, inference time, number of parameters and FLOPs. The absence of detailed description of method performance makes it difficult to make fair comparisons – notice the sparsity of tables \ref{tab:efficiency_oriented}, ~\ref{tab:temporal_aware}, and ~\ref{tab:depth_aware}.

\subsection{Efficiency-Oriented Semantic Segmentation}
\label{subsection:challenges-efficiency}
Considering the context of visual perception for autonomous navigation, the major limitation of efficient model design is its reduced accuracy. In safety-critical applications, though, precision is as important as efficiency.
\par In this context, some inherent characteristics of efficient models play against achieving high accuracy scores. First, the development of completely new architectures prevents the model to leverage the advantages of transfer learning and pre-training on large datasets for model regularization. In addition to that, efficiency-oriented models are inherently lightweight, and reduced computation and memory requirements brings as a consequence reduced feature extraction capacity. 
\par Because of the previous points, how to bridge the accuracy gap between accuracy-oriented and efficiency-oriented Deep Semantic Segmentation models, while maintaining efficiency, is a major challenge. The use of complementary data modalities and tasks, as well as the temporal reasoning on video data can be of great value in this sense.

\subsection{RGB-D Semantic Segmentation}
\label{subsection:challenges-rgb_d}
How to properly represent and fuse Depth information into Deep Semantic Segmentation remains an open issue. In this respect, \cite{b49} defines three important questions to reason about when developing multi-modal perception models: when to fuse, what to fuse and how to fuse – the three questions treat depth images as additional inputs to the network (section~\ref{subsection:rgb-d_semantic_segmentation_depth_as_input}).
\par When to fuse is refers to choosing the level in which both modalities will be fused. According to the authors, early, middle and late fusion are possible strategies. In early fusion, raw data is combined before being fed to the network; in middle fusion, intermediate feature maps are fused; finally, in late fusion setups, feature maps with stronger semantics are fused.
\par What to fuse refers to the type of data representation being fused. For instance, in early fusion setups, raw data, or even pre-processed representations – such as HHA maps –, are fused. However, given that depth maps captured in outdoor environments contain noisy information (mainly for regions with higher depth), directly applying them can lead to suboptimal results due to the influence of noise. Hence, selecting intermediate representations (feature maps) of the input modalities has been proved to work better than early fusion.
\par How to fuse deals with the mechanism selected to perform feature fusion. Element-wise addition and feature concatenation, although frequently used, can lead to limited performance, while hand-crafted modules can lead to higher complexity. In this respect, attention-based and gating mechanisms seem to be a good alternative, as they allow to softly weight feature contributions and reduce the influence of noise, while covering long-range spatial context. The main caveat of such approach is its potentially high computation cost.
\par In addition to the previous considerations, made for depth as input setups, we have discussed depth as operation, depth as prediction, depth as pre-training, and depth as augmentation as alternative approaches to embed depth information in the network architecture.
\par
\begin{figure}[htbp]
\centerline{\includegraphics[width=\columnwidth]{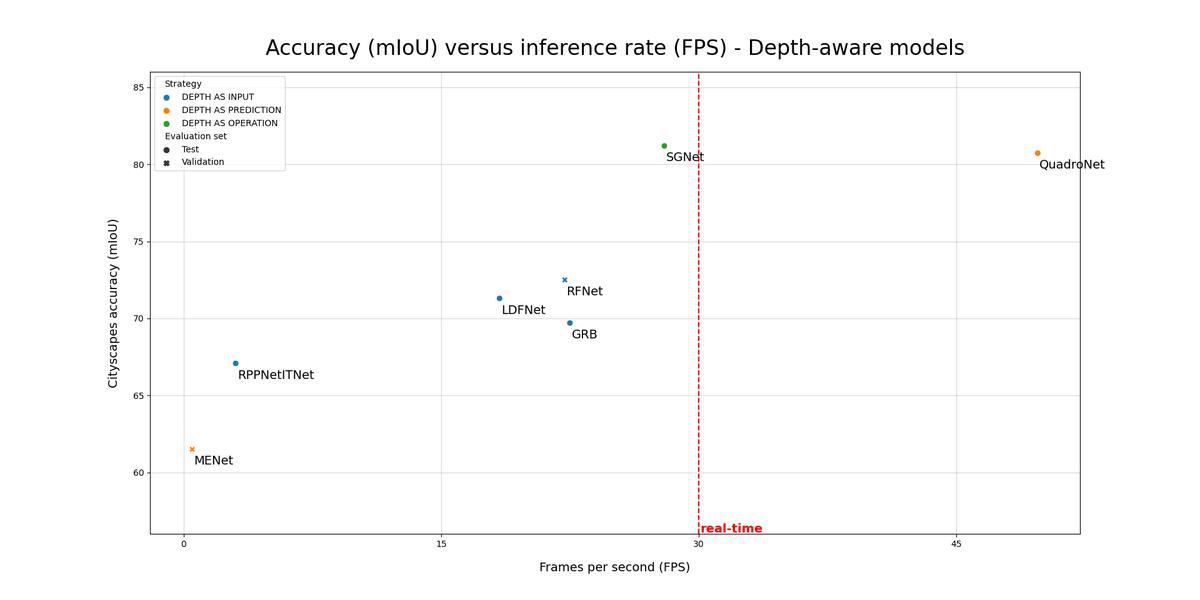}}
\caption{Accuracy (mIoU) \textit{versus} inference rate (FPS) for the depth-aware models reviewed.}
\label{fig:miou_vs_fps_depth}
\end{figure}

 Finally, only one of RGB-D Deep Semantic Segmentation methods reviewed in this work reaches real-time performance – and, for that, it uses acceleration strategies (Fig.~\ref{fig:miou_vs_fps_depth}). Thus, how to develop lightweight RGB-D Deep Semantic Segmentation models remains an open challenge.
\par In summary, how to properly ponder and choose the different aspects involved in deep multi-modal perception, so that to extract the most valuable information for the application at hand, is a major challenge.

\subsection{Temporal-aware Semantic Segmentation}
\label{subsection:challenges-temporal_aware}
\begin{figure}[htbp]
\centerline{\includegraphics[width=\columnwidth]{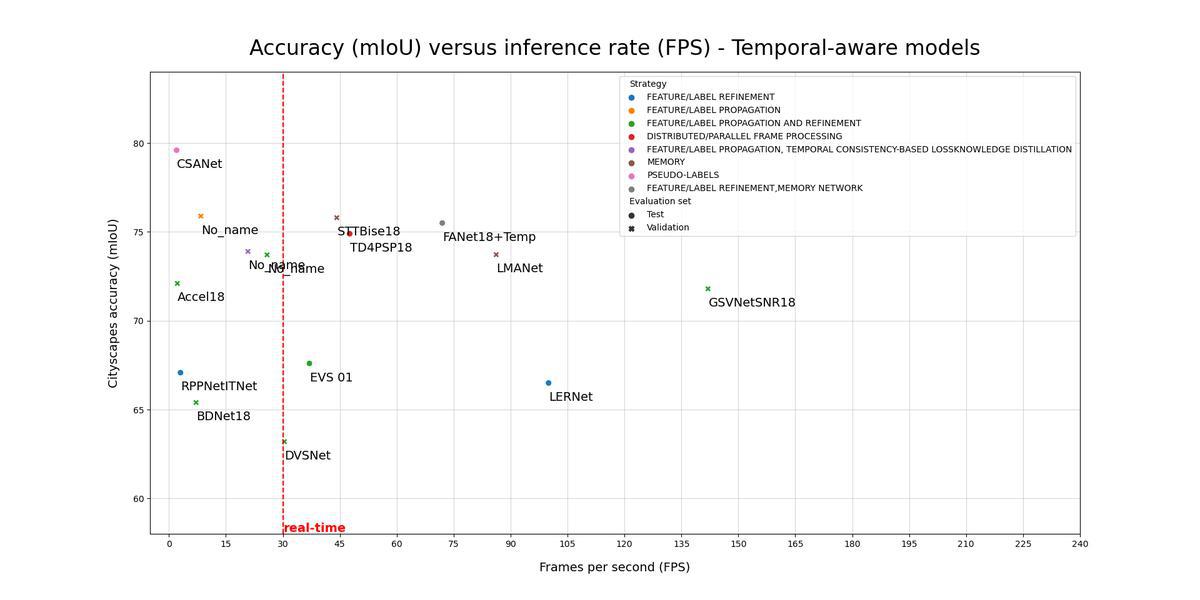}}
\caption{Accuracy (mIoU) \textit{versus} inference rate (FPS) for the temporal-aware models reviewed.}
\label{fig:miou_vs_fps_temporal}
\end{figure}

\begin{figure*}[htbp]
  \begin{subfigure}{0.5\textwidth}
    \includegraphics[width=\linewidth]{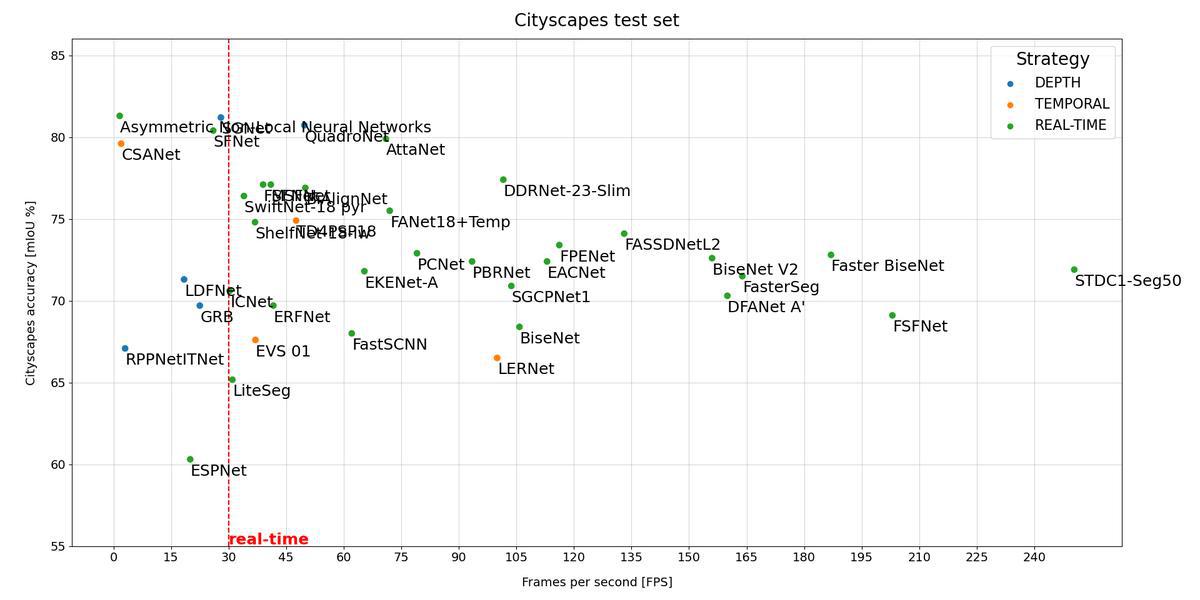}
    \caption{}
    \label{fig:miou_vs_fps_all(a)}
  \end{subfigure}%
  \hspace*{\fill}   
  \begin{subfigure}{0.5\textwidth}
    \includegraphics[width=\linewidth]{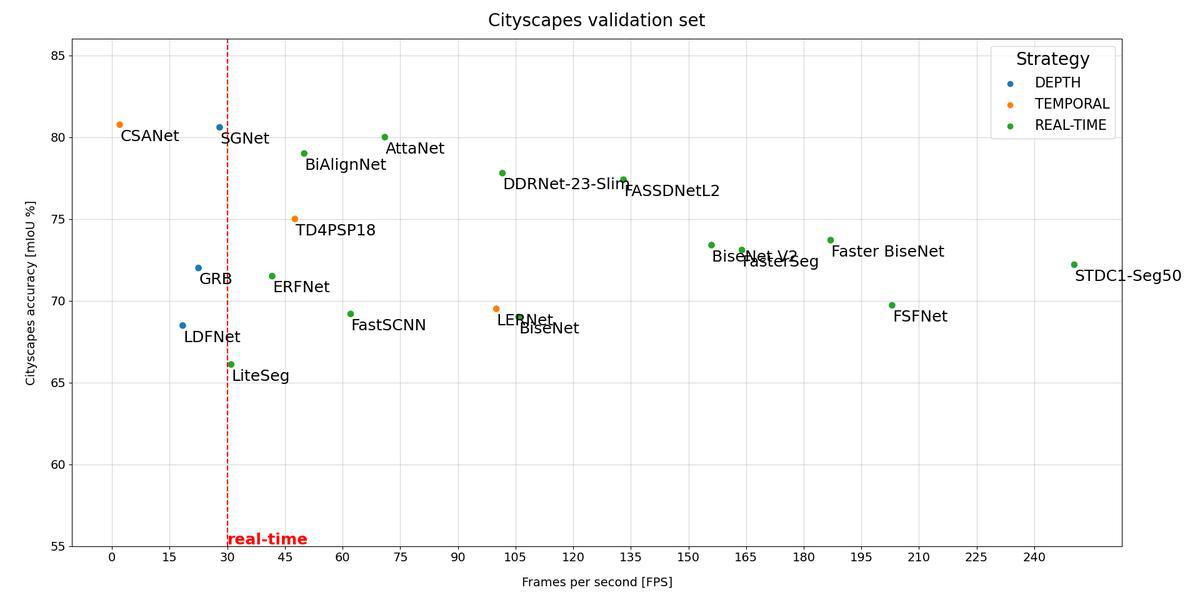}
    \caption{}
    \label{fig:miou_vs_fps_all(b)}
  \end{subfigure}%
  \hspace*{\fill}   
\caption{Accuracy (mIoU) \textit{versus} inference rate (FPS) for all models reviewed. (a) Results on Cityscapes test set. (b) Results on Cityscapes evaluation set.}
\label{fig:miou_vs_fps_all}
\end{figure*}

Video semantic segmentation setups still don’t meet real-time requirements, with many works standing among the slowest methods reviewed – Fig.~\ref{fig:miou_vs_fps_temporal} and Fig.~\ref{fig:miou_vs_fps_all}. Thus, how to leverage temporal information while keeping computations under control is still an open challenge. 
\par In addition to that, how to properly aggregate temporal information remains an open question. The most common methods for video semantic segmentation still rely on recurrent neural networks and optical flow-based feature warping/propagation; nonetheless, the sequential nature of RNNs, and the error-prone optical flow-based propagation are important limitations to take under consideration when developing new architectures. In this respect, memory networks seem to be a good fit, although incurring more memory usage.

\section{Conclusion}
\label{section:conclusion}
This review is intended for researchers who either already have experience or want to enter the field of visual perception, and more specifically semantic segmentation, for autonomous driving in urban scenes. For beginners, our work provides useful discussion and considerations on the main characteristics, advantages, disadvantages, and challenges related to Deep Learning-based Semantic Segmentation (Deep Semantic Segmentation). For experienced researchers, our paper provides a compilation of the most recent works on Deep Semantic Segmentation, as well as in-depth discussions on the use of depth in multi-modal perception setups, temporal data in video semantic segmentation, and the importance of efficiency-oriented model design.
\par Despite being a well-stablished research field, Deep Semantic Segmentation still faces important challenges, with a wide range of promising directions for further development. 
\par Firstly, depth data has proven to be a valuable source of additional information for scene parsing, since it allows the reasoning about scene geometry. Nonetheless, there are many open questions with respect to architectural choices and model design, and recent works still show limited inference time. 
\par Secondly, the exploration of temporal cues from video datasets could bring important advances in terms of robustness, stability, accuracy, and performance. Nonetheless, it has not been explored to its full potential yet. 
\par Additionally, given the current context of development of autonomous systems, semantic segmentation models must be developed considering the computation requirements of embedded devices and edge computing. Hence, the design of efficient models which still preserve competitive results in relation to accuracy-oriented methods is an important line of research.
\par Hence, as a main conclusion, although several works have dealt separately with RGB-D, temporally-aware, and efficiency-oriented semantic segmentation, the efforts on how to properly fuse aspects from these three perspectives, in order to embed Depth and temporal information in efficient architectures so that to achieve real-time inference, while meeting accuracy requirements, is still a major unexplored challenge - Fig.~\ref{fig:works_per_line}.
\par In addition to the aforementioned aspects, the problems of data scarcity and unbalance, how to make full use of unlabeled data and weak labels, and also how to leverage synthetic and related datasets for downstream tasks in real-world domains, remain as major open issues in current deep semantic segmentation.


\begin{thebibliography}{00}
\bibitem{b0} J. Long, E. Shelhamer and T. Darrell, "Fully convolutional networks for semantic segmentation," 2015 IEEE Conference on Computer Vision and Pattern Recognition (CVPR), 2015, pp. 3431-3440, doi: 10.1109/CVPR.2015.7298965.
\bibitem{b1} O. Ronneberger, P. Fischer, and T. Brox, “U-Net: Convolutional Networks for Biomedical Image Segmentation,” in Medical Image Computing and Computer-Assisted Intervention – MICCAI 2015, 2015, pp. 234–241.
\bibitem{b2} V. Badrinarayanan, A. Kendall and R. Cipolla, "SegNet: A Deep Convolutional Encoder-Decoder Architecture for Image Segmentation," in IEEE Transactions on Pattern Analysis and Machine Intelligence, vol. 39, no. 12, pp. 2481-2495, 1 Dec. 2017, doi: 10.1109/TPAMI.2016.2644615.
\bibitem{b3} H. Zhao, J. Shi, X. Qi, X. Wang and J. Jia, "Pyramid Scene Parsing Network," 2017 IEEE Conference on Computer Vision and Pattern Recognition (CVPR), 2017, pp. 6230-6239, doi: 10.1109/CVPR.2017.660.
\bibitem{b5} L. -C. Chen, G. Papandreou, I. Kokkinos, K. Murphy and A. L. Yuille, "DeepLab: Semantic Image Segmentation with Deep Convolutional Nets, Atrous Convolution, and Fully Connected CRFs," in IEEE Transactions on Pattern Analysis and Machine Intelligence, vol. 40, no. 4, pp. 834-848, 1 April 2018, doi: 10.1109/TPAMI.2017.2699184.
\bibitem{b6} L.-C. Chen, G. Papandreou, F. Schroff, and H. Adam, “Rethinking Atrous Convolution for Semantic Image Segmentation,” ArXiv, vol. abs/1706.05587, 2017.
\bibitem{b7} P. Rottmann, T. Posewsky, A. Milioto, C. Stachniss and J. Behley, "Improving Monocular Depth Estimation by Semantic Pre-training," 2021 IEEE/RSJ International Conference on Intelligent Robots and Systems (IROS), 2021, pp. 5916-5923, doi: 10.1109/IROS51168.2021.9636546.
\bibitem{b8} Q. Lv, X. Sun, C. Chen, J. Dong and H. Zhou, "Parallel Complement Network for Real-Time Semantic Segmentation of Road Scenes," in IEEE Transactions on Intelligent Transportation Systems, vol. 23, no. 5, pp. 4432-4444, May 2022, doi: 10.1109/TITS.2020.3044672.
\bibitem{b9} H. Zhao, X. Qi, X. Shen, J. Shi, and J. Jia, “ICNet for Real-Time Semantic Segmentation on High-Resolution Images,” in Computer Vision – ECCV 2018, 2018, pp. 418–434.
\bibitem{b10} L. -C. Chen, Y. Yang, J. Wang, W. Xu and A. L. Yuille, "Attention to Scale: Scale-Aware Semantic Image Segmentation," 2016 IEEE Conference on Computer Vision and Pattern Recognition (CVPR), 2016, pp. 3640-3649, doi: 10.1109/CVPR.2016.396.
\bibitem{b11} Y. Li, X. Li, C. Xiao, H. Li and W. Zhang, "EACNet: Enhanced Asymmetric Convolution for Real-Time Semantic Segmentation," in IEEE Signal Processing Letters, vol. 28, pp. 234-238, 2021, doi: 10.1109/LSP.2021.3051845.
\bibitem{b12} Y. Wu et al., "Fast and Accurate Scene Parsing via Bi-Direction Alignment Networks," 2021 IEEE International Conference on Image Processing (ICIP), 2021, pp. 2508-2512, doi: 10.1109/ICIP42928.2021.9506720.
\bibitem{b13} C. Yu, C. Gao, J. Wang, G. Yu, C. Shen, and N. Sang, “BiSeNet V2: Bilateral Network with Guided Aggregation for Real-Time Semantic Segmentation,” International Journal of Computer Vision, vol. 129, no. 11, pp. 3051–3068, Nov. 2021, doi: 10.1007/s11263-021-01515-2.
\bibitem{b14} M. Yang, K. Yu, C. Zhang, Z. Li and K. Yang, "DenseASPP for Semantic Segmentation in Street Scenes," 2018 IEEE/CVF Conference on Computer Vision and Pattern Recognition, 2018, pp. 3684-3692, doi: 10.1109/CVPR.2018.00388.
\bibitem{b15} H. Si, Z. Zhang, F. Lv, G. Yu, and F. Lu, “Real-Time Semantic Segmentation via Multiply Spatial Fusion Network,” ArXiv, vol. abs/1911.07217, 2020.
\bibitem{b16} M. Liu and H. Yin, “Efficient pyramid context encoding and feature embedding for semantic segmentation,” Image and Vision Computing, vol. 111, p. 104195, 2021, doi: https://doi.org/10.1016/j.imavis.2021.104195.
\bibitem{b17} Z. Huang, C. Lv, Y. Xing and J. Wu, "Multi-Modal Sensor Fusion-Based Deep Neural Network for End-to-End Autonomous Driving With Scene Understanding," in IEEE Sensors Journal, vol. 21, no. 10, pp. 11781-11790, 15 May15, 2021, doi: 10.1109/JSEN.2020.3003121.
\bibitem{b18} L. -Z. Chen, Z. Lin, Z. Wang, Y. -L. Yang and M. -M. Cheng, "Spatial Information Guided Convolution for Real-Time RGBD Semantic Segmentation," in IEEE Transactions on Image Processing, vol. 30, pp. 2313-2324, 2021, doi: 10.1109/TIP.2021.3049332.
\bibitem{b19} H. Li, P. Xiong, H. Fan and J. Sun, "DFANet: Deep Feature Aggregation for Real-Time Semantic Segmentation," 2019 IEEE/CVF Conference on Computer Vision and Pattern Recognition (CVPR), 2019, pp. 9514-9523, doi: 10.1109/CVPR.2019.00975.
\bibitem{b20} Y. Yuan, L. Wang and Y. Wang, "CSANet for Video Semantic Segmentation With Inter-Frame Mutual Learning," in IEEE Signal Processing Letters, vol. 28, pp. 1675-1679, 2021, doi: 10.1109/LSP.2021.3103666.
\bibitem{b21} P. Hu et al., "Real-Time Semantic Segmentation With Fast Attention," in IEEE Robotics and Automation Letters, vol. 6, no. 1, pp. 263-270, Jan. 2021, doi: 10.1109/LRA.2020.3039744.
\bibitem{b22} C. Yu, J. Wang, C. Peng, C. Gao, G. Yu, and N. Sang, “BiSeNet: Bilateral Segmentation Network for Real-Time Semantic Segmentation,” in Computer Vision – ECCV 2018, 2018, pp. 334–349.
\bibitem{b23} Q. Ning, J. Zhu, and C. Chen, “Very Fast Semantic Image Segmentation Using Hierarchical Dilation and Feature Refining,” Cognitive Computation, vol. 10, no. 1, pp. 62–72, Feb. 2018, doi: 10.1007/s12559-017-9530-0.
\bibitem{b24} Y. Hong, H. Pan, W. Sun, and Y. Jia, “Deep Dual-resolution Networks for Real-time and Accurate Semantic Segmentation of Road Scenes,” ArXiv, vol. abs/2101.06085, 2021.
\bibitem{b25} Q. Xu, Y. Ma, J. Wu and C. Long, "Faster BiSeNet: A Faster Bilateral Segmentation Network for Real-time Semantic Segmentation," 2021 International Joint Conference on Neural Networks (IJCNN), 2021, pp. 1-8, doi: 10.1109/IJCNN52387.2021.9533819.
\bibitem{b26} M. Fan et al., "Rethinking BiSeNet For Real-time Semantic Segmentation," 2021 IEEE/CVF Conference on Computer Vision and Pattern Recognition (CVPR), 2021, pp. 9711-9720, doi: 10.1109/CVPR46437.2021.00959.
\bibitem{b27} R. P. K. Poudel, S. Liwicki, and R. Cipolla, “Fast-SCNN: Fast Semantic Segmentation Network,” 2019.
\bibitem{b28} M. Paul, C. Mayer, L. Van Gool and R. Timofte, "Efficient Video Semantic Segmentation with Labels Propagation and Refinement," 2020 IEEE Winter Conference on Applications of Computer Vision (WACV), 2020, pp. 2862-2871, doi: 10.1109/WACV45572.2020.9093520.
\bibitem{b29} Y. Han, “Capturing the spatio-temporal continuity for video semantic segmentation,” IET Image Processing, vol. 13, no. 14, pp. 2813-2820(7), Dec. 2019, [Online]. Available: https://digital-library.theiet.org/content/journals/10.1049/iet-ipr.2018.6479
\bibitem{b30} Y. Li, J. Shi and D. Lin, "Low-Latency Video Semantic Segmentation," 2018 IEEE/CVF Conference on Computer Vision and Pattern Recognition, 2018, pp. 5997-6005, doi: 10.1109/CVPR.2018.00628.
\bibitem{b31} E. Romera, J. M. Álvarez, L. M. Bergasa and R. Arroyo, "ERFNet: Efficient Residual Factorized ConvNet for Real-Time Semantic Segmentation," in IEEE Transactions on Intelligent Transportation Systems, vol. 19, no. 1, pp. 263-272, Jan. 2018, doi: 10.1109/TITS.2017.2750080.
\bibitem{b32} A. Luo, F. Yang, X. Li, R. Huang, and H. Cheng, “EKENet: Efficient knowledge enhanced network for real-time scene parsing,” Pattern Recognition, vol. 111, p. 107671, 2021, doi: https://doi.org/10.1016/j.patcog.2020.107671.
\bibitem{b33} L. Rosas-Arias, G. Benitez-Garcia, J. Portillo-Portillo, G. Sánchez-Pérez and K. Yanai, "Fast and Accurate Real-Time Semantic Segmentation with Dilated Asymmetric Convolutions," 2020 25th International Conference on Pattern Recognition (ICPR), 2021, pp. 2264-2271, doi: 10.1109/ICPR48806.2021.9413176.
\bibitem{b34} X. Li, Z. Liu, P. Luo, C. C. Loy and X. Tang, "Not All Pixels Are Equal: Difficulty-Aware Semantic Segmentation via Deep Layer Cascade," 2017 IEEE Conference on Computer Vision and Pattern Recognition (CVPR), 2017, pp. 6459-6468, doi: 10.1109/CVPR.2017.684.
\bibitem{b35} Y. Liu, C. Shen, C. Yu, and J. Wang, “Efficient Semantic Video Segmentation with Per-Frame Inference,” in Computer Vision – ECCV 2020, 2020, pp. 352–368.
\bibitem{b36} P. Hu, F. Caba, O. Wang, Z. Lin, S. Sclaroff and F. Perazzi, "Temporally Distributed Networks for Fast Video Semantic Segmentation," 2020 IEEE/CVF Conference on Computer Vision and Pattern Recognition (CVPR), 2020, pp. 8815-8824, doi: 10.1109/CVPR42600.2020.00884.
\bibitem{b37} H. Lu and Z. Deng, "A Boundary-aware Distillation Network for Compressed Video Semantic Segmentation," 2020 25th International Conference on Pattern Recognition (ICPR), 2021, pp. 5354-5359, doi: 10.1109/ICPR48806.2021.9412821.
\bibitem{b38} S. Jain, X. Wang and J. E. Gonzalez, "Accel: A Corrective Fusion Network for Efficient Semantic Segmentation on Video," 2019 IEEE/CVF Conference on Computer Vision and Pattern Recognition (CVPR), 2019, pp. 8858-8867, doi: 10.1109/CVPR.2019.00907.
\bibitem{b39} S. -P. Lee, S. -C. Chen and W. -H. Peng, "GSVNET: Guided Spatially-Varying Convolution for Fast Semantic Segmentation on Video," 2021 IEEE International Conference on Multimedia and Expo (ICME), 2021, pp. 1-6, doi: 10.1109/ICME51207.2021.9428381.
\bibitem{b40} S. K. Mustikovela, M. Y. Yang, and C. Rother, “Can Ground Truth Label Propagation from Video Help Semantic Segmentation?,” in Computer Vision – ECCV 2016 Workshops, 2016, pp. 804–820.
\bibitem{b41} Y. Zhu et al., "Improving Semantic Segmentation via Video Propagation and Label Relaxation," 2019 IEEE/CVF Conference on Computer Vision and Pattern Recognition (CVPR), 2019, pp. 8848-8857, doi: 10.1109/CVPR.2019.00906.
\bibitem{b42} P. Z. Ramirez, A. Tonioni, S. Salti and L. D. Stefano, "Learning Across Tasks and Domains," 2019 IEEE/CVF International Conference on Computer Vision (ICCV), 2019, pp. 8109-8118, doi: 10.1109/ICCV.2019.00820.
\bibitem{b43} Y. Chen, W. Li, X. Chen and L. Van Gool, "Learning Semantic Segmentation From Synthetic Data: A Geometrically Guided Input-Output Adaptation Approach," 2019 IEEE/CVF Conference on Computer Vision and Pattern Recognition (CVPR), 2019, pp. 1841-1850, doi: 10.1109/CVPR.2019.00194.
\bibitem{b44} R. Chavhan, A. Jha, B. Banerjee and S. Chaudhuri, "ADA-AT/DT: An Adversarial Approach for Cross-Domain and Cross-Task Knowledge Transfer," 2021 IEEE Winter Conference on Applications of Computer Vision (WACV), 2021, pp. 3501-3510, doi: 10.1109/WACV48630.2021.00354.
\bibitem{b45} A. Cardace, L. De Luigi, P. Zama Ramirez, S. Salti and L. Di Stefano, "Plugging Self-Supervised Monocular Depth into Unsupervised Domain Adaptation for Semantic Segmentation," 2022 IEEE/CVF Winter Conference on Applications of Computer Vision (WACV), 2022, pp. 1999-2009, doi: 10.1109/WACV51458.2022.00206.
\bibitem{b46} Q. Wang, D. Dai, L. Hoyer, L. Van Gool and O. Fink, "Domain Adaptive Semantic Segmentation with Self-Supervised Depth Estimation," 2021 IEEE/CVF International Conference on Computer Vision (ICCV), 2021, pp. 8495-8505, doi: 10.1109/ICCV48922.2021.00840.
\bibitem{b47} H. Yu et al., “Methods and datasets on semantic segmentation: A review,” Neurocomputing, vol. 304, pp. 82–103, 2018, doi: https://doi.org/10.1016/j.neucom.2018.03.037.
\bibitem{b48} N. Atif, M. Bhuyan and S. Ahamed, "A Review on Semantic Segmentation from a Modern Perspective," 2019 International Conference on Electrical, Electronics and Computer Engineering (UPCON), 2019, pp. 1-6, doi: 10.1109/UPCON47278.2019.8980189.
\bibitem{b49} D. Feng et al., "Deep Multi-Modal Object Detection and Semantic Segmentation for Autonomous Driving: Datasets, Methods, and Challenges," in IEEE Transactions on Intelligent Transportation Systems, vol. 22, no. 3, pp. 1341-1360, March 2021, doi: 10.1109/TITS.2020.2972974.
\bibitem{b50} C. Wang, C. Wang, W. Li, and H. Wang, “A brief survey on RGB-D semantic segmentation using deep learning,” Displays, vol. 70, p. 102080, 2021, doi: https://doi.org/10.1016/j.displa.2021.102080.
\bibitem{b51} S. Barchid, J. Mennesson and C. Djéraba, "Review on Indoor RGB-D Semantic Segmentation with Deep Convolutional Neural Networks," 2021 International Conference on Content-Based Multimedia Indexing (CBMI), 2021, pp. 1-4, doi: 10.1109/CBMI50038.2021.9461875.
\bibitem{b52} F. Fooladgar and S. Kasaei, “A survey on indoor RGB-D semantic segmentation: from hand-crafted features to deep convolutional neural networks,” Multimedia Tools and Applications, vol. 79, no. 7, pp. 4499–4524, Feb. 2020, doi: 10.1007/s11042-019-7684-3.
\bibitem{b53} Y. Hu, Z. Chen and W. Lin, "RGB-D Semantic Segmentation: A Review," 2018 IEEE International Conference on Multimedia \& Expo Workshops (ICMEW), 2018, pp. 1-6, doi: 10.1109/ICMEW.2018.8551554.
\bibitem{b54} S. Papadopoulos, I. Mademlis and I. Pitas, "Neural vision-based semantic 3D world modeling," 2021 IEEE Winter Conference on Applications of Computer Vision Workshops (WACVW), 2021, pp. 181-190, doi: 10.1109/WACVW52041.2021.00024.
\bibitem{b55} F. Lateef and Y. Ruichek, “Survey on semantic segmentation using deep learning techniques,” Neurocomputing, vol. 338, pp. 321–348, 2019, doi: https://doi.org/10.1016/j.neucom.2019.02.003.
\bibitem{b56} W. Wang, T. Zhou, F. M. Porikli, D. J. Crandall, and L. V. Gool, “A Survey on Deep Learning Technique for Video Segmentation,” ArXiv, vol. abs/2107.01153, 2021.
\bibitem{b57} M. Siam, S. Elkerdawy, M. Jagersand and S. Yogamani, "Deep semantic segmentation for automated driving: Taxonomy, roadmap and challenges," 2017 IEEE 20th International Conference on Intelligent Transportation Systems (ITSC), 2017, pp. 1-8, doi: 10.1109/ITSC.2017.8317714.
\bibitem{b58} A. Garcia-Garcia, S. Orts, S. Oprea, V. Villena-Martinez, and J. G. Rodríguez, “A Review on Deep Learning Techniques Applied to Semantic Segmentation,” ArXiv, vol. abs/1704.06857, 2017.
\bibitem{b59} A. Garcia-Garcia, S. Orts-Escolano, S. Oprea, V. Villena-Martinez, P. Martinez-Gonzalez, and J. Garcia-Rodriguez, “A survey on deep learning techniques for image and video semantic segmentation,” Applied Soft Computing, vol. 70, pp. 41–65, 2018, doi: https://doi.org/10.1016/j.asoc.2018.05.018.
\bibitem{b60} M. Siam, M. Gamal, M. Abdel-Razek, S. Yogamani, M. Jagersand and H. Zhang, "A Comparative Study of Real-Time Semantic Segmentation for Autonomous Driving," 2018 IEEE/CVF Conference on Computer Vision and Pattern Recognition Workshops (CVPRW), 2018, pp. 700-70010, doi: 10.1109/CVPRW.2018.00101.
\bibitem{b61} Y. Mo, Y. Wu, X. Yang, F. Liu, and Y. Liao, “Review the state-of-the-art technologies of semantic segmentation based on deep learning,” Neurocomputing, vol. 493, pp. 626–646, 2022, doi: https://doi.org/10.1016/j.neucom.2022.01.005.
\bibitem{b62} M. Cordts et al., "The Cityscapes Dataset for Semantic Urban Scene Understanding," 2016 IEEE Conference on Computer Vision and Pattern Recognition (CVPR), 2016, pp. 3213-3223, doi: 10.1109/CVPR.2016.350.
\bibitem{b63} K. Simonyan and A. Zisserman, “Very Deep Convolutional Networks for Large-Scale Image Recognition,” CoRR, vol. abs/1409.1556, 2015.
\bibitem{b64} A. Chaurasia and E. Culurciello, "LinkNet: Exploiting encoder representations for efficient semantic segmentation," 2017 IEEE Visual Communications and Image Processing (VCIP), 2017, pp. 1-4, doi: 10.1109/VCIP.2017.8305148.
\bibitem{b65} X. Li et al., “Semantic Flow for Fast and Accurate Scene Parsing,” in Computer Vision – ECCV 2020, 2020, pp. 775–793.
\bibitem{b66} M. Oršić and S. Šegvić, “Efficient semantic segmentation with pyramidal fusion,” Pattern Recognition, vol. 110, p. 107611, 2021, doi: https://doi.org/10.1016/j.patcog.2020.107611.
\bibitem{b67} B. Wang, L. Li, Y. Nakashima, R. Kawasaki, H. Nagahara and Y. Yagi, "Noisy-LSTM: Improving Temporal Awareness for Video Semantic Segmentation," in IEEE Access, vol. 9, pp. 46810-46820, 2021, doi: 10.1109/ACCESS.2021.3067928.
\bibitem{b68} Y. Pei, B. Sun and S. Li, "Multifeature Selective Fusion Network for Real-Time Driving Scene Parsing," in IEEE Transactions on Instrumentation and Measurement, vol. 70, pp. 1-12, 2021, Art no. 5008412, doi: 10.1109/TIM.2021.3070611.
\bibitem{b69} D. Xu, W. Ouyang, X. Wang and N. Sebe, "PAD-Net: Multi-tasks Guided Prediction-and-Distillation Network for Simultaneous Depth Estimation and Scene Parsing," 2018 IEEE/CVF Conference on Computer Vision and Pattern Recognition, 2018, pp. 675-684, doi: 10.1109/CVPR.2018.00077.
\bibitem{b70} Q. Song, K. Mei, and R. Huang, “AttaNet: Attention-Augmented Network for Fast and Accurate Scene Parsing”, AAAI, vol. 35, no. 3, pp. 2567-2575, May 2021.
\bibitem{b71} L. Sun, K. Yang, X. Hu, W. Hu and K. Wang, "Real-Time Fusion Network for RGB-D Semantic Segmentation Incorporating Unexpected Obstacle Detection for Road-Driving Images," in IEEE Robotics and Automation Letters, vol. 5, no. 4, pp. 5558-5565, Oct. 2020, doi: 10.1109/LRA.2020.3007457.
\bibitem{b72} J. -Y. Sun, S. -W. Jung and S. -J. Ko, "Lightweight Prediction and Boundary Attention-Based Semantic Segmentation for Road Scene Understanding," in IEEE Access, vol. 8, pp. 108449-108460, 2020, doi: 10.1109/ACCESS.2020.3001679.
\bibitem{b73} L. Li, B. Qian, J. Lian, W. Zheng and Y. Zhou, "Traffic Scene Segmentation Based on RGB-D Image and Deep Learning," in IEEE Transactions on Intelligent Transportation Systems, vol. 19, no. 5, pp. 1664-1669, May 2018, doi: 10.1109/TITS.2017.2724138.
\bibitem{b74} L. Li, W. Zheng, L. Kong, Ü. Özgüner, W. Hou and J. Lian, "Real-time Traffic Scene Segmentation Based on Multi-Feature Map and Deep Learning," 2018 IEEE Intelligent Vehicles Symposium (IV), 2018, pp. 7-12, doi: 10.1109/IVS.2018.8500467.
\bibitem{b75} J. Dai et al., "Deformable Convolutional Networks," 2017 IEEE International Conference on Computer Vision (ICCV), 2017, pp. 764-773, doi: 10.1109/ICCV.2017.89.
\bibitem{b76} T. Emara, H. E. A. E. Munim and H. M. Abbas, "LiteSeg: A Novel Lightweight ConvNet for Semantic Segmentation," 2019 Digital Image Computing: Techniques and Applications (DICTA), 2019, pp. 1-7, doi: 10.1109/DICTA47822.2019.8945975.
\bibitem{b77} X. Wang, R. Girshick, A. Gupta and K. He, "Non-local Neural Networks," 2018 IEEE/CVF Conference on Computer Vision and Pattern Recognition, 2018, pp. 7794-7803, doi: 10.1109/CVPR.2018.00813.
\bibitem{b78} A. Jha, A. Kumar, S. Pande, B. Banerjee and S. Chaudhuri, "MT-UNET: A Novel U-Net Based Multi-Task Architecture For Visual Scene Understanding," 2020 IEEE International Conference on Image Processing (ICIP), 2020, pp. 2191-2195, doi: 10.1109/ICIP40778.2020.9190695.
\bibitem{b79} Y. Cao, J. Xu, S. Lin, F. Wei and H. Hu, "GCNet: Non-Local Networks Meet Squeeze-Excitation Networks and Beyond," 2019 IEEE/CVF International Conference on Computer Vision Workshop (ICCVW), 2019, pp. 1971-1980, doi: 10.1109/ICCVW.2019.00246.
\bibitem{b80} W. Shi et al., "RGB-D Semantic Segmentation and Label-Oriented Voxelgrid Fusion for Accurate 3D Semantic Mapping," in IEEE Transactions on Circuits and Systems for Video Technology, vol. 32, no. 1, pp. 183-197, Jan. 2022, doi: 10.1109/TCSVT.2021.3056726.
\bibitem{b81} Y. Li et al., "Learning Dynamic Routing for Semantic Segmentation," 2020 IEEE/CVF Conference on Computer Vision and Pattern Recognition (CVPR), 2020, pp. 8550-8559, doi: 10.1109/CVPR42600.2020.00858.
\bibitem{b82} V. John et al., "Sensor Fusion of Intensity and Depth Cues using the ChiNet for Semantic Segmentation of Road Scenes," 2018 IEEE Intelligent Vehicles Symposium (IV), 2018, pp. 585-590, doi: 10.1109/IVS.2018.8500476.
\bibitem{b83} M. Kim, B. Park and S. Chi, "Accelerator-Aware Fast Spatial Feature Network for Real-Time Semantic Segmentation," in IEEE Access, vol. 8, pp. 226524-226537, 2020, doi: 10.1109/ACCESS.2020.3045147.
\bibitem{b84} A. Krizhevsky, I. Sutskever, and G. E. Hinton, “ImageNet Classification with Deep Convolutional Neural Networks,” in Advances in Neural Information Processing Systems, 2012, vol. 25. [Online]. Available: https://proceedings.neurips.cc/paper/2012/file/c399862d3b9d6b76c8436e924a68c45b-Paper.pdf
\bibitem{b85} A. G. Howard et al., “MobileNets: Efficient Convolutional Neural Networks for Mobile Vision Applications,” ArXiv, vol. abs/1704.04861, 2017.
\bibitem{b86} M. Sandler, A. Howard, M. Zhu, A. Zhmoginov and L. -C. Chen, "MobileNetV2: Inverted Residuals and Linear Bottlenecks," 2018 IEEE/CVF Conference on Computer Vision and Pattern Recognition, 2018, pp. 4510-4520, doi: 10.1109/CVPR.2018.00474.
\bibitem{b87} A. Howard et al., "Searching for MobileNetV3," 2019 IEEE/CVF International Conference on Computer Vision (ICCV), 2019, pp. 1314-1324, doi: 10.1109/ICCV.2019.00140.
\bibitem{b88} K. He, X. Zhang, S. Ren and J. Sun, "Deep Residual Learning for Image Recognition," 2016 IEEE Conference on Computer Vision and Pattern Recognition (CVPR), 2016, pp. 770-778, doi: 10.1109/CVPR.2016.90.
\bibitem{b89} F. Chollet, "Xception: Deep Learning with Depthwise Separable Convolutions," 2017 IEEE Conference on Computer Vision and Pattern Recognition (CVPR), 2017, pp. 1800-1807, doi: 10.1109/CVPR.2017.195.
\bibitem{b90} S. Hao, Y. Zhou, Y. Guo, R. Hong, J. Cheng and M. Wang, "Real-Time Semantic Segmentation via Spatial-Detail Guided Context Propagation," in IEEE Transactions on Neural Networks and Learning Systems, doi: 10.1109/TNNLS.2022.3154443.
\bibitem{b91} X. Zhang, X. Zhou, M. Lin and J. Sun, "ShuffleNet: An Extremely Efficient Convolutional Neural Network for Mobile Devices," 2018 IEEE/CVF Conference on Computer Vision and Pattern Recognition, 2018, pp. 6848-6856, doi: 10.1109/CVPR.2018.00716.
\bibitem{b92} J. Redmon and A. Farhadi, "YOLO9000: Better, Faster, Stronger," 2017 IEEE Conference on Computer Vision and Pattern Recognition (CVPR), 2017, pp. 6517-6525, doi: 10.1109/CVPR.2017.690.
\bibitem{b93} W. Chen, X. Gong, X. Liu, Q. Zhang, Y. Li, and Z. Wang, “FasterSeg: Searching for Faster Real-time Semantic Segmentation,” ArXiv, vol. abs/1912.10917, 2020.
\bibitem{b94} H. Wang, W. Wang and J. Liu, "Temporal Memory Attention for Video Semantic Segmentation," 2021 IEEE International Conference on Image Processing (ICIP), 2021, pp. 2254-2258, doi: 10.1109/ICIP42928.2021.9506731.
\bibitem{b95} J. Zhuang, J. Yang, L. Gu and N. Dvornek, "ShelfNet for Fast Semantic Segmentation," 2019 IEEE/CVF International Conference on Computer Vision Workshop (ICCVW), 2019, pp. 847-856, doi: 10.1109/ICCVW.2019.00113.
\bibitem{b96} L. Zhou, H. Yuan and C. Ge, "ConvLSTM-based Neural Network for Video Semantic Segmentation," 2021 International Conference on Visual Communications and Image Processing (VCIP), 2021, pp. 1-5, doi: 10.1109/VCIP53242.2021.9675363.
\bibitem{b97} D. Oh, D. Ji, C. Jang, Y. Hyun, H. S. Bae and S. Hwang, "Segmenting 2K-Videos at 36.5 FPS with 24.3 GFLOPs: Accurate and Lightweight Realtime Semantic Segmentation Network," 2020 IEEE International Conference on Robotics and Automation (ICRA), 2020, pp. 3153-3160, doi: 10.1109/ICRA40945.2020.9196510.
\bibitem{b98} W. Wang, "Semi-supervised Semantic Segmentation Network based on Knowledge Distillation," 2021 IEEE 4th Advanced Information Management, Communicates, Electronic and Automation Control Conference (IMCEC), 2021, pp. 1900-1905, doi: 10.1109/IMCEC51613.2021.9482145.
\bibitem{b99} Y. -S. Xu, T. -J. Fu, H. -K. Yang and C. -Y. Lee, "Dynamic Video Segmentation Network," 2018 IEEE/CVF Conference on Computer Vision and Pattern Recognition, 2018, pp. 6556-6565, doi: 10.1109/CVPR.2018.00686.
\bibitem{b100} S. Mehta, M. Rastegari, A. Caspi, L. Shapiro, and H. Hajishirzi, “ESPNet: Efficient Spatial Pyramid of Dilated Convolutions for Semantic Segmentation,” in Computer Vision – ECCV 2018, 2018, pp. 561–580.
\bibitem{b101} M. Paul, M. Danelljan, L. V. Gool and R. Timofte, "Local Memory Attention for Fast Video Semantic Segmentation," 2021 IEEE/RSJ International Conference on Intelligent Robots and Systems (IROS), 2021, pp. 1102-1109, doi: 10.1109/IROS51168.2021.9636192.
\bibitem{b102} Z. Zhu, M. Xu, S. Bai, T. Huang and X. Bai, "Asymmetric Non-Local Neural Networks for Semantic Segmentation," 2019 IEEE/CVF International Conference on Computer Vision (ICCV), 2019, pp. 593-602, doi: 10.1109/ICCV.2019.00068.
\bibitem{b103} J. Li et al., “Video Semantic Segmentation via Sparse Temporal Transformer,” in Proceedings of the 29th ACM International Conference on Multimedia, 2021, pp. 59–68. doi: 10.1145/3474085.3475409.
\bibitem{b104} E. Xie, W. Wang, Z. Yu, A. Anandkumar, J. M. Álvarez, and P. Luo, “SegFormer: Simple and Efficient Design for Semantic Segmentation with Transformers,” 2021.
\bibitem{b105} G. Neuhold, T. Ollmann, S. R. Bulò and P. Kontschieder, "The Mapillary Vistas Dataset for Semantic Understanding of Street Scenes," 2017 IEEE International Conference on Computer Vision (ICCV), 2017, pp. 5000-5009, doi: 10.1109/ICCV.2017.534.
\bibitem{b106} J. Xiong, L. -M. Po, W. Y. Yu, Y. Zhao and K. -W. Cheung, "Distortion Map-Guided Feature Rectification for Efficient Video Semantic Segmentation," in IEEE Transactions on Multimedia, doi: 10.1109/TMM.2021.3136085.
\bibitem{b107} C. Hazirbas, L. Ma, C. Domokos, and D. Cremers, “FuseNet: Incorporating Depth into Semantic Segmentation via Fusion-Based CNN Architecture,” in Computer Vision – ACCV 2016, 2017, pp. 213–228.
\bibitem{b108} Y. Qian, L. Deng, T. Li, C. Wang and M. Yang, "Gated-Residual Block for Semantic Segmentation Using RGB-D Data," in IEEE Transactions on Intelligent Transportation Systems, vol. 23, no. 8, pp. 11836-11844, Aug. 2022, doi: 10.1109/TITS.2021.3107672.
\bibitem{b109} X. Zhang, Y. Chen, H. Zhang, S. Wang, J. Lu and J. Yang, "When Visual Disparity Generation Meets Semantic Segmentation: A Mutual Encouragement Approach," in IEEE Transactions on Intelligent Transportation Systems, vol. 22, no. 3, pp. 1853-1867, March 2021, doi: 10.1109/TITS.2020.3027556.
\bibitem{b110} J. Zhang, K. A. Skinner, R. Vasudevan and M. Johnson-Roberson, "DispSegNet: Leveraging Semantics for End-to-End Learning of Disparity Estimation From Stereo Imagery," in IEEE Robotics and Automation Letters, vol. 4, no. 2, pp. 1162-1169, April 2019, doi: 10.1109/LRA.2019.2894913.
\bibitem{b111} L. Zhou, H. Zhang, Y. Long, L. Shao and J. Yang, "Depth Embedded Recurrent Predictive Parsing Network for Video Scenes," in IEEE Transactions on Intelligent Transportation Systems, vol. 20, no. 12, pp. 4643-4654, Dec. 2019, doi: 10.1109/TITS.2019.2909053.
\bibitem{b112} H. Rashed, A. El Sallab, S. Yogamani and M. ElHelw, "Motion and Depth Augmented Semantic Segmentation for Autonomous Navigation," 2019 IEEE/CVF Conference on Computer Vision and Pattern Recognition Workshops (CVPRW), 2019, pp. 364-370, doi: 10.1109/CVPRW.2019.00049.
\bibitem{b113} S. -W. Hung, S. -Y. Lo and H. -M. Hang, "Incorporating Luminance, Depth and Color Information by a Fusion-Based Network for Semantic Segmentation," 2019 IEEE International Conference on Image Processing (ICIP), 2019, pp. 2374-2378, doi: 10.1109/ICIP.2019.8803360.
\bibitem{b114} W. Wang and U. Neumann, “Depth-Aware CNN for RGB-D Segmentation,” in Computer Vision – ECCV 2018, 2018, pp. 144–161.
\bibitem{b115} M. D. Ansari., S. Krauß., O. Wasenmüller., and D. Stricker., “ScaleNet: Scale Invariant Network for Semantic Segmentation in Urban Driving Scenes,” in Proceedings of the 13th International Joint Conference on Computer Vision, Imaging and Computer Graphics Theory and Applications - Volume 5: VISAPP, 2018, pp. 399–404. doi: 10.5220/0006723003990404.
\bibitem{b116} S. Kong and C. Fowlkes, "Recurrent Scene Parsing with Perspective Understanding in the Loop," 2018 IEEE/CVF Conference on Computer Vision and Pattern Recognition, 2018, pp. 956-965, doi: 10.1109/CVPR.2018.00106.
\bibitem{b117} A. R. Zamir, A. Sax, W. Shen, L. Guibas, J. Malik and S. Savarese, "Taskonomy: Disentangling Task Transfer Learning," 2018 IEEE/CVF Conference on Computer Vision and Pattern Recognition, 2018, pp. 3712-3722, doi: 10.1109/CVPR.2018.00391.
\bibitem{b118} S. Chennupati, G. Sistu, S. Yogamani, and S. Rawashdeh, “AuxNet: Auxiliary tasks enhanced Semantic Segmentation for Automated Driving,” arXiv e-prints, p. earXiv:1901.05808, Jan. 2019.
\bibitem{b119} M. Aladem and S. A. Rawashdeh, "A Single-Stream Segmentation and Depth Prediction CNN for Autonomous Driving," in IEEE Intelligent Systems, vol. 36, no. 4, pp. 79-85, 1 July-Aug. 2021, doi: 10.1109/MIS.2020.2993266.
\bibitem{b120} A. Jha, B. Banerjee, and S. Chaudhuri, “S3 DMT-Net: Improving Soft Sharing Based Multi-Task CNN Using Task-Specific Distillation and Cross-Task Interactions,” 2021. doi: 10.1145/3490035.3490274.
\bibitem{b121} L. Hoyer, D. Dai, Y. Chen, A. Köring, S. Saha and L. Van Gool, "Three Ways to Improve Semantic Segmentation with Self-Supervised Depth Estimation," 2021 IEEE/CVF Conference on Computer Vision and Pattern Recognition (CVPR), 2021, pp. 11125-11135, doi: 10.1109/CVPR46437.2021.01098.
\bibitem{b122} K. Goel, P. Srinivasan, S. Tariq and J. Philbin, "QuadroNet: Multi-Task Learning for Real-Time Semantic Depth Aware Instance Segmentation," 2021 IEEE Winter Conference on Applications of Computer Vision (WACV), 2021, pp. 315-324, doi: 10.1109/WACV48630.2021.00036.
\bibitem{b123} S. Chennupati, G. Sistu, S. Yogamani and S. A. Rawashdeh, "MultiNet++: Multi-Stream Feature Aggregation and Geometric Loss Strategy for Multi-Task Learning," 2019 IEEE/CVF Conference on Computer Vision and Pattern Recognition Workshops (CVPRW), 2019, pp. 1200-1210, doi: 10.1109/CVPRW.2019.00159.
\bibitem{b124} Z. Wu, X. Wu, X. Zhang, S. Wang and L. Ju, "Semantic Stereo Matching With Pyramid Cost Volumes," 2019 IEEE/CVF International Conference on Computer Vision (ICCV), 2019, pp. 7483-7492, doi: 10.1109/ICCV.2019.00758.
\bibitem{b125} A. Atapour-Abarghouei and T. P. Breckon, "Veritatem Dies Aperit - Temporally Consistent Depth Prediction Enabled by a Multi-Task Geometric and Semantic Scene Understanding Approach," 2019 IEEE/CVF Conference on Computer Vision and Pattern Recognition (CVPR), 2019, pp. 3368-3379, doi: 10.1109/CVPR.2019.00349.
\bibitem{b126} C. Zhang, Y. Tang, C. Zhao, Q. Sun, Z. Ye and J. Kurths, "Multitask GANs for Semantic Segmentation and Depth Completion With Cycle Consistency," in IEEE Transactions on Neural Networks and Learning Systems, vol. 32, no. 12, pp. 5404-5415, Dec. 2021, doi: 10.1109/TNNLS.2021.3072883.
\bibitem{b127} H. Jiang, G. Larsson, M. Maire, G. Shakhnarovich, and E. Learned-Miller, “Self-Supervised Relative Depth Learning for Urban Scene Understanding,” in Computer Vision – ECCV 2018, 2018, pp. 20–37.
\bibitem{b128} S. Papadopoulos, I. Mademlis and I. Pitas, "Semantic Image Segmentation Guided By Scene Geometry," 2021 IEEE International Conference on Autonomous Systems (ICAS), 2021, pp. 1-5, doi: 10.1109/ICAS49788.2021.9551117.
\bibitem{b129} A. Loukkal, Y. Grandvalet and Y. Li, "Disparity weighted loss for semantic segmentation of driving scenes," 2019 IEEE Intelligent Transportation Systems Conference (ITSC), 2019, pp. 3427-3432, doi: 10.1109/ITSC.2019.8917171.
\bibitem{b130} M. Siam, S. Valipour, M. Jagersand, N. Ray and S. Yogamani, "Convolutional gated recurrent networks for video semantic segmentation in automated driving," 2017 IEEE 20th International Conference on Intelligent Transportation Systems (ITSC), 2017, pp. 1-7, doi: 10.1109/ITSC.2017.8317600.
\bibitem{b131} D. Tran, L. Bourdev, R. Fergus, L. Torresani and M. Paluri, "Learning Spatiotemporal Features with 3D Convolutional Networks," 2015 IEEE International Conference on Computer Vision (ICCV), 2015, pp. 4489-4497, doi: 10.1109/ICCV.2015.510.
\bibitem{b132} Z. Qiu, T. Yao and T. Mei, "Learning Deep Spatio-Temporal Dependence for Semantic Video Segmentation," in IEEE Transactions on Multimedia, vol. 20, no. 4, pp. 939-949, April 2018, doi: 10.1109/TMM.2017.2759504.
\bibitem{b133} S. D. Jain, B. Xiong and K. Grauman, "FusionSeg: Learning to Combine Motion and Appearance for Fully Automatic Segmentation of Generic Objects in Videos," 2017 IEEE Conference on Computer Vision and Pattern Recognition (CVPR), 2017, pp. 2117-2126, doi: 10.1109/CVPR.2017.228.
\bibitem{b134} P. Tokmakov, K. Alahari and C. Schmid, "Learning Video Object Segmentation with Visual Memory," 2017 IEEE International Conference on Computer Vision (ICCV), 2017, pp. 4491-4500, doi: 10.1109/ICCV.2017.480.
\bibitem{b135} H. Li, G. Chen, G. Li and Y. Yu, "Motion Guided Attention for Video Salient Object Detection," 2019 IEEE/CVF International Conference on Computer Vision (ICCV), 2019, pp. 7273-7282, doi: 10.1109/ICCV.2019.00737.
\bibitem{b136} S. Hochreiter and J. Schmidhuber, "Long Short-Term Memory," in Neural Computation, vol. 9, no. 8, pp. 1735-1780, 15 Nov. 1997, doi: 10.1162/neco.1997.9.8.1735.
\bibitem{b137} K. Cho, B. van Merriënboer, D. Bahdanau, and Y. Bengio, “On the Properties of Neural Machine Translation: Encoder–Decoder Approaches,” in Proceedings of SSST-8, Eighth Workshop on Syntax, Semantics and Structure in Statistical Translation, Oct. 2014, pp. 103–111. doi: 10.3115/v1/W14-4012.
\bibitem{b138} X. Shi, Z. Chen, H. Wang, D.-Y. Yeung, W.-K. Wong, and chun Wang-Woo, “Convolutional LSTM Network: A Machine Learning Approach for Precipitation Nowcasting,” 2015.
\bibitem{b139} N. Ballas, L. Yao, C. J. Pal, and A. C. Courville, “Delving Deeper into Convolutional Networks for Learning Video Representations,” CoRR, vol. abs/1511.06432, 2016.
\bibitem{b140} E. E. Yurdakul and Y. Yemez, "Semantic Segmentation of RGBD Videos with Recurrent Fully Convolutional Neural Networks," 2017 IEEE International Conference on Computer Vision Workshops (ICCVW), 2017, pp. 367-374, doi: 10.1109/ICCVW.2017.51.
\bibitem{b141} V. Lup and S. Nedevschi, "Video Semantic Segmentation leveraging Dense Optical Flow," 2020 IEEE 16th International Conference on Intelligent Computer Communication and Processing (ICCP), 2020, pp. 369-376, doi: 10.1109/ICCP51029.2020.9266150.
\bibitem{b142} D. Nilsson and C. Sminchisescu, "Semantic Video Segmentation by Gated Recurrent Flow Propagation," 2018 IEEE/CVF Conference on Computer Vision and Pattern Recognition, 2018, pp. 6819-6828, doi: 10.1109/CVPR.2018.00713.
\bibitem{b143} E. Shelhamer, K. Rakelly, J. Hoffman, and T. Darrell, “Clockwork Convnets for Video Semantic Segmentation,” in Computer Vision – ECCV 2016 Workshops, 2016, pp. 852–868.
\bibitem{b144} J. Zhuang, Z. Wang and B. Wang, "Video Semantic Segmentation With Distortion-Aware Feature Correction," in IEEE Transactions on Circuits and Systems for Video Technology, vol. 31, no. 8, pp. 3128-3139, Aug. 2021, doi: 10.1109/TCSVT.2020.3037234.
\bibitem{b145} J. Wu, Z. Wen, S. Zhao, and K. Huang, “Video semantic segmentation via feature propagation with holistic attention,” Pattern Recognition, vol. 104, p. 107268, 2020, doi: https://doi.org/10.1016/j.patcog.2020.107268.
\bibitem{b146} R. Gadde, V. Jampani and P. V. Gehler, "Semantic Video CNNs Through Representation Warping," 2017 IEEE International Conference on Computer Vision (ICCV), 2017, pp. 4463-4472, doi: 10.1109/ICCV.2017.477.
\bibitem{b147} A. Dosovitskiy et al., "FlowNet: Learning Optical Flow with Convolutional Networks," 2015 IEEE International Conference on Computer Vision (ICCV), 2015, pp. 2758-2766, doi: 10.1109/ICCV.2015.316.
\bibitem{b148} E. Ilg, N. Mayer, T. Saikia, M. Keuper, A. Dosovitskiy and T. Brox, "FlowNet 2.0: Evolution of Optical Flow Estimation with Deep Networks," 2017 IEEE Conference on Computer Vision and Pattern Recognition (CVPR), 2017, pp. 1647-1655, doi: 10.1109/CVPR.2017.179.
\bibitem{b149} D. Sun, X. Yang, M. -Y. Liu and J. Kautz, "PWC-Net: CNNs for Optical Flow Using Pyramid, Warping, and Cost Volume," 2018 IEEE/CVF Conference on Computer Vision and Pattern Recognition, 2018, pp. 8934-8943, doi: 10.1109/CVPR.2018.00931.
\bibitem{b150} A. Ranjan and M. J. Black, "Optical Flow Estimation Using a Spatial Pyramid Network," 2017 IEEE Conference on Computer Vision and Pattern Recognition (CVPR), 2017, pp. 2720-2729, doi: 10.1109/CVPR.2017.291.
\bibitem{b151} A. Geiger, P. Lenz and R. Urtasun, "Are we ready for autonomous driving? The KITTI vision benchmark suite," 2012 IEEE Conference on Computer Vision and Pattern Recognition, 2012, pp. 3354-3361, doi: 10.1109/CVPR.2012.6248074.
\bibitem{b152} S. Varghese et al., "An Unsupervised Temporal Consistency (TC) Loss to Improve the Performance of Semantic Segmentation Networks," 2021 IEEE/CVF Conference on Computer Vision and Pattern Recognition Workshops (CVPRW), 2021, pp. 12-20, doi: 10.1109/CVPRW53098.2021.00010.
\bibitem{b153} J. Xie, M. Kiefel, M. -T. Sun and A. Geiger, "Semantic Instance Annotation of Street Scenes by 3D to 2D Label Transfer," 2016 IEEE Conference on Computer Vision and Pattern Recognition (CVPR), 2016, pp. 3688-3697, doi: 10.1109/CVPR.2016.401.
\bibitem{b154} A. Gurram, O. Urfalioglu, I. Halfaoui, F. Bouzaraa and A. M. López, "Monocular Depth Estimation by Learning from Heterogeneous Datasets," 2018 IEEE Intelligent Vehicles Symposium (IV), 2018, pp. 2176-2181, doi: 10.1109/IVS.2018.8500683.
\bibitem{b155} P. Liu et al., “WeClick: Weakly-Supervised Video Semantic Segmentation with Click Annotations,” in Proceedings of the 29th ACM International Conference on Multimedia, 2021, pp. 2995–3004. doi: 10.1145/3474085.3475217.
\bibitem{b156} F. S. Saleh, M. S. Aliakbarian, M. Salzmann, L. Petersson and J. M. Alvarez, "Bringing Background into the Foreground: Making All Classes Equal in Weakly-Supervised Video Semantic Segmentation," 2017 IEEE International Conference on Computer Vision (ICCV), 2017, pp. 2125-2135, doi: 10.1109/ICCV.2017.232.
\bibitem{b157} K. He, H. Fan, Y. Wu, S. Xie and R. Girshick, "Momentum Contrast for Unsupervised Visual Representation Learning," 2020 IEEE/CVF Conference on Computer Vision and Pattern Recognition (CVPR), 2020, pp. 9726-9735, doi: 10.1109/CVPR42600.2020.00975.
\bibitem{b158} T. Chen, S. Kornblith, M. Norouzi, and G. Hinton, “A Simple Framework for Contrastive Learning of Visual Representations,” 2020.
\bibitem{b159} I. Misra and L. van der Maaten, "Self-Supervised Learning of Pretext-Invariant Representations," 2020 IEEE/CVF Conference on Computer Vision and Pattern Recognition (CVPR), 2020, pp. 6706-6716, doi: 10.1109/CVPR42600.2020.00674.
\bibitem{b160} Z. Yang, H. Yu, Y. He, W. Sun, Z. -H. Mao and A. Mian, "Fully Convolutional Network-Based Self-Supervised Learning for Semantic Segmentation," in IEEE Transactions on Neural Networks and Learning Systems, doi: 10.1109/TNNLS.2022.3172423.
\bibitem{b161} C. Doersch, A. Gupta and A. A. Efros, "Unsupervised Visual Representation Learning by Context Prediction," 2015 IEEE International Conference on Computer Vision (ICCV), 2015, pp. 1422-1430, doi: 10.1109/ICCV.2015.167.
\bibitem{b162} R. Zhang, P. Isola, and A. A. Efros, “Colorful Image Colorization,” in Computer Vision – ECCV 2016, 2016, pp. 649–666.
\bibitem{b163} G. Larsson, M. Maire, and G. Shakhnarovich, “Learning Representations for Automatic Colorization,” in Computer Vision – ECCV 2016, 2016, pp. 577–593.
\bibitem{b164} R. Zhang, P. Isola and A. A. Efros, "Split-Brain Autoencoders: Unsupervised Learning by Cross-Channel Prediction," 2017 IEEE Conference on Computer Vision and Pattern Recognition (CVPR), 2017, pp. 645-654, doi: 10.1109/CVPR.2017.76.
\bibitem{b165} D. Pathak, P. Krähenbühl, J. Donahue, T. Darrell and A. A. Efros, "Context Encoders: Feature Learning by Inpainting," 2016 IEEE Conference on Computer Vision and Pattern Recognition (CVPR), 2016, pp. 2536-2544, doi: 10.1109/CVPR.2016.278.
\bibitem{b166} S. Gidaris, P. Singh, and N. Komodakis, “Unsupervised Representation Learning by Predicting Image Rotations,” ArXiv, vol. abs/1803.07728, 2018.
\bibitem{b167} S. Jenni and P. Favaro, "Self-Supervised Feature Learning by Learning to Spot Artifacts," 2018 IEEE/CVF Conference on Computer Vision and Pattern Recognition, 2018, pp. 2733-2742, doi: 10.1109/CVPR.2018.00289.
\bibitem{b168} H. -Y. Lee, J. -B. Huang, M. Singh and M. -H. Yang, "Unsupervised Representation Learning by Sorting Sequences," 2017 IEEE International Conference on Computer Vision (ICCV), 2017, pp. 667-676, doi: 10.1109/ICCV.2017.79.
\bibitem{b169} I. Misra, C. L. Zitnick, and M. Hebert, “Shuffle and Learn: Unsupervised Learning Using Temporal Order Verification,” in Computer Vision – ECCV 2016, 2016, pp. 527–544.
\bibitem{b170} D. Xu, J. Xiao, Z. Zhao, J. Shao, D. Xie and Y. Zhuang, "Self-Supervised Spatiotemporal Learning via Video Clip Order Prediction," 2019 IEEE/CVF Conference on Computer Vision and Pattern Recognition (CVPR), 2019, pp. 10326-10335, doi: 10.1109/CVPR.2019.01058.
\bibitem{b171} N. Srivastava, E. Mansimov, and R. Salakhutdinov, “Unsupervised Learning of Video Representations Using LSTMs,” in Proceedings of the 32nd International Conference on International Conference on Machine Learning - Volume 37, 2015, pp. 843–852.
\bibitem{b172} Y. Zhang and J. J. Leonard, "Bootstrapped Self-Supervised Training with Monocular Video for Semantic Segmentation and Depth Estimation," 2021 IEEE/RSJ International Conference on Intelligent Robots and Systems (IROS), 2021, pp. 2420-2427, doi: 10.1109/IROS51168.2021.9636330.
\bibitem{b173} G. Ros, L. Sellart, J. Materzynska, D. Vazquez and A. M. Lopez, "The SYNTHIA Dataset: A Large Collection of Synthetic Images for Semantic Segmentation of Urban Scenes," 2016 IEEE Conference on Computer Vision and Pattern Recognition (CVPR), 2016, pp. 3234-3243, doi: 10.1109/CVPR.2016.352.
\bibitem{b174} A. Dosovitskiy, G. Ros, F. Codevilla, A. M. López, and V. Koltun, “CARLA: An Open Urban Driving Simulator,” ArXiv, vol. abs/1711.03938, 2017.
\bibitem{b175}S. R. Richter, V. Vineet, S. Roth, and V. Koltun, “Playing for Data: Ground Truth from Computer Games,” in Computer Vision – ECCV 2016, 2016, pp. 102–118.
\bibitem{b176} M. Wang and W. Deng, “Deep visual domain adaptation: A survey,” Neurocomputing, vol. 312, pp. 135–153, 2018, doi: https://doi.org/10.1016/j.neucom.2018.05.083.
\bibitem{b177} N. Araslanov and S. Roth, "Self-supervised Augmentation Consistency for Adapting Semantic Segmentation," 2021 IEEE/CVF Conference on Computer Vision and Pattern Recognition (CVPR), 2021, pp. 15379-15389, doi: 10.1109/CVPR46437.2021.01513.
\bibitem{b178} Y. Zhang, D. Sidibé, O. Morel and F. Meriaudeau, "Incorporating Depth Information into Few-Shot Semantic Segmentation," 2020 25th International Conference on Pattern Recognition (ICPR), 2021, pp. 3582-3588, doi: 10.1109/ICPR48806.2021.9412921.
\bibitem{b179} F. Yu and V. Koltun, “Multi-Scale Context Aggregation by Dilated Convolutions,” CoRR, vol. abs/1511.07122, 2016.
\bibitem{b180} S. Woo, J. Park, J.-Y. Lee, and I. S. Kweon, “CBAM: Convolutional Block Attention Module,” in Computer Vision – ECCV 2018, 2018, pp. 3–19.
\bibitem{b181} W. Li and Z. Li, "Domain Adaptative Semantic Segmentation by alleviating Long-tail Problem," 2021 International Joint Conference on Neural Networks (IJCNN), 2021, pp. 1-8, doi: 10.1109/IJCNN52387.2021.9533948.




\end{thebibliography}
\end{document}